\documentclass[11pt, letterpaper]{article}

\usepackage[left=1in, right=1in, top=1in, bottom=1in]{geometry}
\usepackage[utf8]{inputenc}
\usepackage[T1]{fontenc}
\usepackage{bm}
\usepackage{type1cm}
\usepackage{lettrine}
\usepackage{amsmath,amssymb,amsthm}
\usepackage{moreverb}
\usepackage{mathtools}
\usepackage{amsmath}
\usepackage{amssymb}
\usepackage{algorithmic}
\usepackage{graphics}
\usepackage{graphicx}
\usepackage{subfigure}
\usepackage{caption}
\usepackage{extarrows}
\usepackage{color}
\usepackage{framed}
\usepackage{wrapfig}
\usepackage{bm}
\usepackage{mathrsfs}
\usepackage{mathabx}
\usepackage{multirow}
\usepackage{longtable}
\usepackage{hyperref}
\usepackage{paralist}
\usepackage{indentfirst}
\usepackage{relsize}
\usepackage{extarrows}
\usepackage{lineno,xcolor}
\usepackage{upgreek}
\usepackage{bm}
\usepackage{mwe}
\usepackage{xcolor}
\usepackage{booktabs}
\usepackage{authblk}
\usepackage{lettrine}
\usepackage{type1cm}
\usepackage{threeparttable}
\usepackage{pbox}
\usepackage[sort&compress,numbers]{natbib}

\definecolor{mygray}{gray}{0.8}

\graphicspath{ {./Figure/} }
\usepackage[font=footnotesize,labelfont=bf]{caption}
\newcommand{\eref}[1]{(\ref{#1})}
\providecommand{\keywords}[1]{\textbf{\textit{Keywords: }} #1}

\newcommand{\hsedit}[1]{{\color{black} #1}}

\newcommand{\hseditr}[1]{{\color{black} #1}}

\DeclareMathOperator*{\argmin}{arg\,min}

\newcommand{\redit}[1]{{\color{black} #1}}

\hypersetup{
bookmarks=true,
bookmarksopen=true,
bookmarksnumbered=true,
unicode=false,
pdftoolbar=true,
pdfmenubar=true,
pdffitwindow=false,
pdfstartview={FitH},
pdftitle={My title},
pdfauthor={Author},
pdfsubject={Subject},
pdfcreator={Creator},
pdfproducer={Producer},
pdfkeywords={keywords},
pdfnewwindow=true,
colorlinks=true,
linkcolor=blue,
citecolor=blue,
filecolor=magenta,
urlcolor=blue
}

\begin{document}

\title{\textbf{Encoding physics to learn reaction-diffusion processes}}

\author[1,2,$\dag$]{Chengping Rao}
\author[3,$\dag$]{Pu Ren}
\author[1]{\hseditr{Qi Wang}}
\author[4]{Oral Buyukozturk}
\author[1,5,*]{Hao Sun}
\author[6,$\ddag$]{Yang Liu}

\affil[1]{\small Gaoling School of Artificial Intelligence, Renmin University of China, Beijing, 100872, China}
\affil[2]{\small Department of Mechanical and Industrial Engineering, Northeastern University, Boston, MA 02115, USA}
\affil[3]{Department of Civil and Environmental Engineering, Northeastern University, Boston, MA 02115, USA}
\affil[4]{Department of Civil and Environmental Engineering, MIT, Cambridge, MA 02139, USA} 
\affil[5]{Beijing Key Laboratory of Big Data Management and Analysis Methods, Beijing, 100872, China} 
\affil[6]{\small School of Engineering Sciences, University of Chinese Academy of Sciences, Beijing, 101408, China \vspace{12pt}}
\affil[$\dag$]{Equally contributed\vspace{12pt}}
\affil[*]{Corresponding author. E-mail: haosun@ruc.edu.cn}
\affil[$\ddag$]{Corresponding author. E-mail: liuyang22@ucas.ac.cn\vspace{12pt}}

\date{}

\maketitle

\vspace{-28pt} 
\begin{abstract}
	\small
    Modeling complex \hsedit{spatiotemporal} dynamical systems, such as the reaction-diffusion processes, have largely relied on partial differential equations (PDEs). However, due to insufficient prior knowledge on some under-explored dynamical systems, such as those in chemistry, biology, geology, physics and ecology, and the lack of explicit PDE formulation used for describing the nonlinear process of the system variables, to predict the evolution of such a system remains a challenging task. Unifying measurement data and our limited prior physics knowledge via machine learning provides us with a new path to solving this problem. Existing physics-informed learning paradigms impose physics laws through soft penalty constraints, whose solution quality largely depends on a trial-and-error proper setting of hyperparameters. Since the core of such methods is still rooted in black-box neural networks, the resulting model generally lacks interpretability and suffers from critical issues of extrapolation and generalization. To this end, we propose a deep learning framework that forcibly \hseditr{encodes} given physics structure to facilitate the learning of the spatiotemporal dynamics in sparse data regimes. We show how the proposed approach can be applied to \hsedit{a variety of problems regarding the PDE system}, including forward and inverse analysis, data-driven modeling, and discovery of \hsedit{PDEs}. The resultant learning paradigm that encodes physics shows high accuracy, robustness, interpretability and generalizability demonstrated via extensive numerical experiments.
\end{abstract}


\vspace{12pt} 


\section{INTRODUCTION}
\hsedit{Spatiotemporal dynamics is ubiquitous in nature. For example, reaction-diffusion} processes exhibit interesting phenomena and are commonly seen in many disciplines such as chemistry, biology, geology, physics and ecology. \hsedit{The} autonomous formation mechanism of stripe (Turing) patterns in the skin of tropical fishes can be revealed by diffusion and reaction. Like many other systems, understanding their complex spatiotemporal dynamics, governed by the inherent partial differential equations (PDEs), is a central task. Nevertheless, the principled laws in the context of closed-form governing equations for many underexplored systems remain uncertain or partially unknown (e.g., the reaction mechanism is typically nonlinear and intractable to model). Even for some dynamical systems, e.g., other than reaction-diffusion processes, whose PDEs are already known (such as Naiver-Stokes equations), the computational cost of accurate numerical simulation is also prohibitive for scientific applications involving large-scale spatiotemporal domains. \hsedit{Yet, machine learning (ML) has opened} up new avenues for scientific modeling or discovery of the aforementioned systems \hsedit{in a data-driven manner}. 

In fact, the history of knowledge discovery from data (or observation) can be dated back to the time of Kepler, who discovered the well-known law of planetary motion from massive documented data. Recently, the revived \hsedit{ML} methods \hsedit{have} pushed the renaissance of the data-driven scientific \hsedit{computing, such as modeling of complex systems \cite{raissi2017machine,han2018solving,bar2019learning,sanchez2020learning,long2018pde,wang2020towards,pfaff2020learning,belbute2020combining, kochkov2021machine}, super-resolution of scientific data \cite{erichson2020shallow, stengel2020adversarial, fukami2021machine}, material property prediction \cite{rao2020three}, system identification and equation discovery \cite{schmidt2009distilling,brunton2016discovering,rudy2017data,udrescu2020ai}, among others. These successful applications are largely attributed to the extraordinary expressiveness of deep learning models, which enables the automatic learning of the nonlinear mapping among variables from rich labeled data \cite{lecun2015deep}. In particular, latest research showed that deep learning \cite{lusch2018deep,long2019pde,chen2021physics,cranmer2021discover} could accelerate the discovery of underlying governing PDEs given sparse/noisy data. However, the pure data-driven methods rooted on deep learning typically learn representations from and highly relies on big data (e.g., from experiment or simulation) which are often insufficient in most scientific problems. The resulting model often fails to satisfy physical constraints (e.g., conservation laws, invariants), whose generalizability cannot be guaranteed either \cite{karpatne2017theory}. To tackle this issue, physics-informed neural network (PINN) \cite{raissi2019physics, raissi2020hidden, karniadakis2021physics} has taken a remarkable leap in scientific machine learning and become a major paradigm, which leverages our prior knowledge of underlying physics to enable learning in small data regimes.}

\hsedit{PINN} has shown effectiveness in a wide range of scientific applications, including solving general PDEs \cite{raissi2019physics, rao2020physicsTAML, sheng2021pfnn}, reduced-order modeling \cite{sun2020surrogate, kim2021fast}, uncertainty quantification (UQ) \cite{yang2019adversarial, zhu2019physics}, inverse problems \cite{raissi2019physics, haghighat2021physics}, data-driven knowledge discovery \cite{chen2021physics} and others. In particular, the paradigm has been demonstrated to be effective in modeling a variety of physical systems, such as fluid dynamics \cite{raissi2020hidden, jin2021nsfnets}, subsurface transport \cite{he2020physics, he2021physics}, and engineering mechanics \cite{zhang2020physics, rao2020physicsElastic, haghighat2021physics, niaki2021physics}. However, the dominant physics-informed learning model, PINN, generally represents a continuous learning paradigm as it employs fully-connected neural networks (FCNNs) for the continuous approximation of the solution to the physical system. The resultant continuous representation of the system's prediction brings several limitations including {\color{black} poor computational efficiency} due to the nature of FCNN, inaccurate physical constraints due to the soft penalty in the loss function and lack of capability to hard-encode prior physics into the learning model. Fortunately, latest studies in discrete learning models, such as the convolutional neural networks (CNNs) \cite{yu2017deep, zhu2019physics, ren2022phycrnet, gao2021phygeonet}, graph neural networks (GNNs) \cite{gao2022physics} and transformers \cite{geneva2021transformers}, show promises in overcoming some of the above limitations. Compared with the continuous learning model, the discrete learning approaches have a distinct advantage of hard encoding the initial and boundary conditions (I/BCs), as well as the incomplete PDE \hsedit{structure}, into the learning model. This practice could avoid the ill-posedness of the optimization even without any labeled data as shown in very recent works \cite{geneva2020modeling, ren2022phycrnet, gao2021super}. Therefore, we are motivated to establish an effective, interpretable and generalizable discrete learning paradigm that can be leveraged for predicting the nonlinear physical systems, which remains a significant challenge in scientific machine learning. Recent advances show that operator learning can naturally achieve this goal, e.g., DeepONet \cite{lu2021learning} and Fourier Neural Operator (FNO) \cite{li2020fourier}. However, a rich set of labeled data should be supplied to train reliable operators for these methods. Although adding prior physics to constrain DeepONet helps alleviate the need of large data \cite{wang2021learning}, the explicit expression of PDE(s) must be given, which falls short in dealing with systems whose governing equations are partially or completely unknown.

To this end, we propose the physics-encoded model that \hseditr{encodes} the prior physics knowledge in the network architecture, in contrast to ``teaching'' models the physics through penalized loss function commonly seen in physics-informed learning. In particular, our model has four major characteristics: (1) Compared with the dominant method of PINN that utilizes FCNN as a continuous approximator to the solution, the physics-encoded model is discrete (i.e., solution is spatially mesh-based and defined on discrete time steps) {\color{black} and} hard encodes the given physics structure into the network architecture. (2) Our model employs a unique convolutional network (i.e., $\Pi$-block discussed in \textcolor{blue}{Methods}) to capture the spatial patterns of the system while the time marching is performed by the recurrent unit. This unique network has been demonstrated (with mathematical proof and numerical experiments) to promote the expressiveness of our model on nonlinear spatiotemporal dynamics. (3) Thanks to the discretization along time, our network is able to incorporate well-known numerical time integration methods (e.g., forward Euler scheme, Runge-Kutta scheme) for encoding incomplete PDEs into the network architecture. Throughout this article, we demonstrate the capabilities of the proposed network architecture by applying it to various tasks in scientific modeling \hsedit{of spatiotemporal dynamics such as the reaction-diffusion processes}.


\section*{RESULTS}

\subsection*{Physics-encoded Spatiotemporal Learning} 
\begin{figure}[t!]
    \centering
	\includegraphics[width=1.0\linewidth]{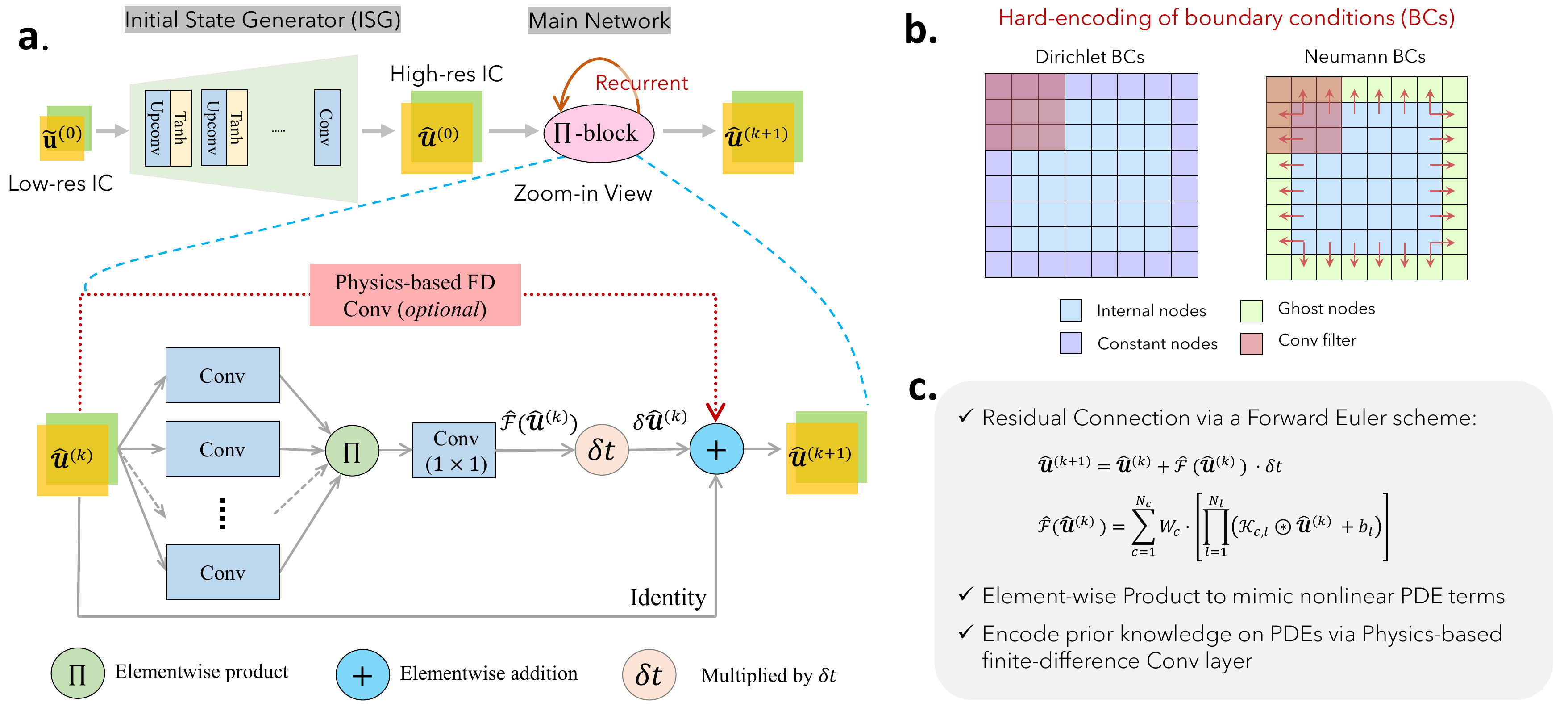} 
	\caption{Schematic architecture of PeRCNN: \textbf{a.} the network with a $\Pi$-block for the recurrent computation; \textbf{b.} embedding of BCs; and \textbf{c.} key features of the network. Here, $\tilde{\mathbf{u}}^{(0)}$ denotes the low-res noisy measurement of the initial state, while $\widehat{\boldsymbol{\mathcal{U}}}^{(k)}$ denotes the predicted fine-res solution at time $t_k$. The decoder (initial state generator) is used to downscale/upsample the low-resolution initial state.} 
	\label{Diagram}
\end{figure}

The motivation of the physics-encoded spatiotemporal learning paradigm is to establish a generalizable and robust model for predicting the physical system state based on very limited low-resolution and noisy measurement data. The established model is expected to deliver good extrapolation capability over the temporal horizon and generalization to different \hsedit{initial conditions (ICs)}. Those demands essentially require the proposed model to learn the underlying spatiotemporal dynamics from data. \hsedit{To this end, we propose a novel network, namely the Physics-encoded Recurrent Convolutional Neural Network (PeRCNN) as shown in Figure \ref{Diagram}. The network is designed to preserve the given physics structure, e.g., structure or specific terms of the governing PDEs, ICs, and boundary conditions (BCs). The prior physics knowledge is forcibly ``encoded'' which makes the network possess interpretability. More details are given in the \textcolor{blue}{Methods} section.}

\subsection*{Reaction-diffusion Systems}
\label{sec:rd_system}

Reaction-diffusion (RD) equations have found wide applications in the analysis of pattern formation \cite{halatek2018rethinking}, such as population dynamics \cite{holmes1994partial}, chemical reactions \cite{vervloet2012fischer}, cell {\color{black}proliferations \cite{maini2004traveling}, and etc}. In this article, we specifically consider three different RD systems of the Lambda--Omega ($\lambda$--$\Omega$), FitzHugh-Nagumo (FN) and Gray-Scott (GS) types \hsedit{to verify the proposed approach}. In general, the RD system can be described by the following governing equation
\begin{equation} 
    \label{eq:gs_eqn} 
    \mathbf{u}_t=\mathbf{D}\Delta \mathbf{u}+\mathbf{R(u)} 
\end{equation}
where $\mathbf{u}\in\mathbb{R}^n$ is the vector of concentration variables, $\mathbf{D}\in\mathbb{R}^{n\times n}$ is the diagonal diffusion coefficient matrix, $\Delta$ is the Laplacian operator, and $\mathbf{R(u)}$ is the reaction vector that represents the interactions among components of $\mathbf{u}$. Without loss of generality, let us assume the RD system features two components, i.e., $\mathbf{u}= [u,~v]^\texttt{T}$. Specifically, the $\lambda$--$\Omega$ RD system is governed by
\begin{equation} 
\label{eq:LO_eqn} 
\begin{aligned}
u_t&=\mu_u\Delta u + (1-u^2-v^2)u+\beta (u^2+v^2)v \\
v_t&=\mu_v\Delta v - \beta (u^2+v^2)u + (1-u^2-v^2)v
\end{aligned}
\end{equation}
while FN RD system can be described by 
\begin{equation} 
\label{eq:fn_eqn} 
\begin{aligned}
u_t&=\mu_u\Delta u+u-u^3-v+\alpha \\
v_t&=\mu_v\Delta v+(u-v)\beta
\end{aligned}
\end{equation}
where $\alpha$ and $\beta$ are the coefficients prescribing the reaction process and take different values. Similarly, the GS RD system can be described by
\begin{equation} 
\label{eq:gs_eqn} 
\begin{aligned}
u_t&=\mu_u\Delta u-uv^2+F(1-u) \\
v_t&=\mu_v\Delta v+uv^2-(F+\kappa)v
\end{aligned}
\end{equation}
where $\kappa$ and $F$ denote the kill and feed rate, respectively. For the FN and GS RD systems, we consider both two-dimensional (2D) and three-dimensional (3D) cases in this article. To generate the numerical solution as the {\color{black}ground} truth reference, we discretize the regular physical domain with Cartesian grid and utilize a high-order finite difference method to simulate the evolution of the RD system. The computational and discretization parameters for each case are provided in Extended Data Table \ref{tb:data_set}.  

\subsection*{Forward Analysis of PDE Systems}
Solving general PDEs is undoubtedly the cornerstone of scientific computing. We herein demonstrate the capability of PeRCNN for forward analysis of PDE systems (i.e., solving PDEs), in particular, the aforementioned RD systems. It is assumed that the governing PDEs, together with the necessary initial and boundary conditions (I/BCs), of the physical system are completely known. Therefore, the prescribed initial state is fed to the network for the recurrent computation without resorting to ISG. Once the forward recurrent computation is finished, the snapshot (or prediction) at each time step is collected. The finite difference is then applied on the discrete snapshots for computing the partial derivatives involved in the governing PDE. The mean squared error (MSE) of the equation residual is used as the optimization objective (or loss function) for obtaining a set of model's parameters. Multiple RD systems, including the 2D $\lambda$--$\Omega$, 2D/3D FN and 2D GS RD equations, are considered herein as the numerical examples. The ground truth reference solution is generated by the high-order finite difference method. The governing PDEs, computational domain and discretization settings for each system are provided in Extended Data Table \ref{tb:data_set}. Detailed discussions on the network settings is given in \textcolor{blue}{Supplementary Note C}. \hsedit{Details on how to properly select the spatial and temporal grid sizes, in order to ensure model's numerical stability and achieve desired solution resolution, are given in \textcolor{blue}{Supplementary Note H}.} 

\begin{figure}[htbp]
\centering
\includegraphics[width=0.95\textwidth]{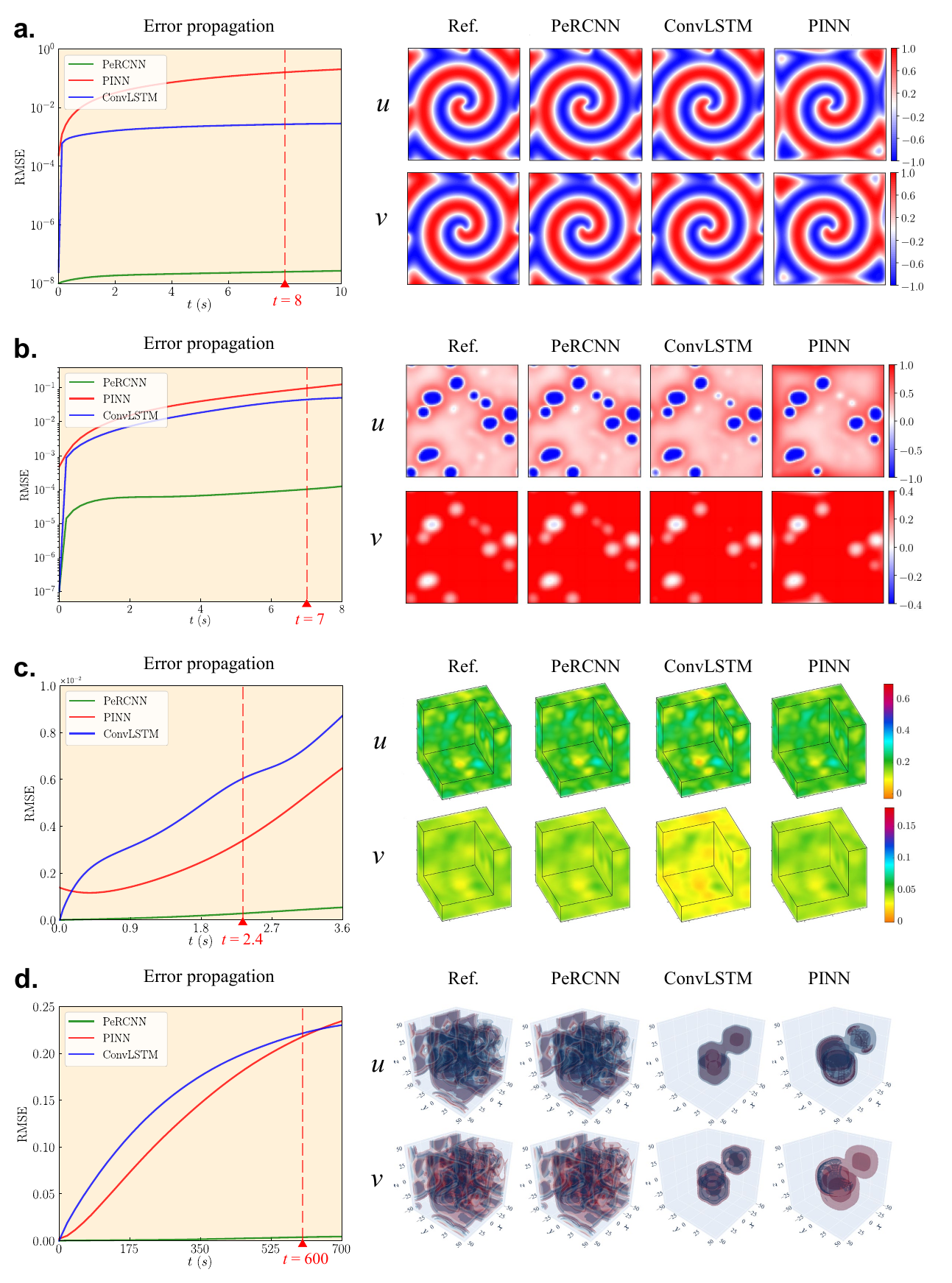}
\caption{Error propagation curve (left) and predicted snapshots (right) by PeRCNN, ConvLSTM and PINN on various RD systems. \textbf{a}, 2D $\lambda$--$\Omega$ RD equation. \textbf{b}, 2D FN RD equation. \textbf{c}, 3D FN RD equation. and $\textbf{d}$, 3D GS RD equation. Note that only a corner of the solution field is shown for the FN RD systems in \textbf{c} for better visualization of the internal solution distribution.}
\label{fig:pde_solve_result}
\end{figure}

Figure \ref{fig:pde_solve_result} shows the snapshots predicted by the PeRCNN for each system at a given time. To compare the performance of the proposed approach with existing methods, we also provide the result of two baseline models, namely, the Convolutional Long-Short Term Memory (ConvLSTM) \cite{shi2015convolutional} and Physics-informed Neural Network (PINN) \cite{raissi2019physics}. It can be seen that the solution obtained by PeRCNN agrees well with the reference for all four cases. In contrast, the ConvLSTM and PINN perform differently on 2D and 3D cases; in particular, they get a fairly good prediction for 2D cases while considerably deviate from the reference for 3D cases. To examine the accuracy of each method as a PDE solver, we compute the accumulative Rooted Mean Square Error (RMSE) of the prediction. It shows that PeRCNN achieves a significantly lower error throughout the considered time-marching interval. Although existing techniques (e.g., finite difference/volume/element methods) for solving PDEs are already mature nowadays, the result in this part demonstrates the promise of PeRCNN on modeling and simulation of complex systems.

\subsection*{Inverse Analysis of PDE Systems}
Calibrating the unknown parameters of a given model against experimental data is a commonly seen problem in scientific discovery, e.g., one might be interested in uncovering the scalar coefficients in the governing PDEs given very limited observed snapshots of the system. Since the proposed PeRCNN has the capability of encoding the PDE structure (e.g., known terms) into the network architecture for predicting spatiotemporal systems, we can apply it to identify the unknown coefficients by treating them as trainable variables. To verify the effectiveness of PeRCNN in inverse analysis of PDEs, we consider the 2D GS RD system governed by the following two coupled equations: 
$u_t=\mu_u\Delta u-c_1\cdot uv^2+c_F(1-u)$
and 
$v_t=\mu_v\Delta v+c_2\cdot uv^2-(c_F+c_\kappa)v$, 
where $\mu_u$, $\mu_v$, $c_1$, $c_2$, $c_F$ and $c_{\kappa}$ are unknown coefficients. Since the explicit form of the governing PDEs is known, we construct the physics-based FD Conv connections (i.e., diffusion and other polynomial terms) according to the RHS of the governing equation. Note that no element-wise product layer is involved in the network. In the meanwhile, each unknown coefficient is treated as an individual trainable variable in the computational graph for the forward/backward computations. 

\begin{figure}[t!]
\centering
\includegraphics[width=0.99\textwidth]{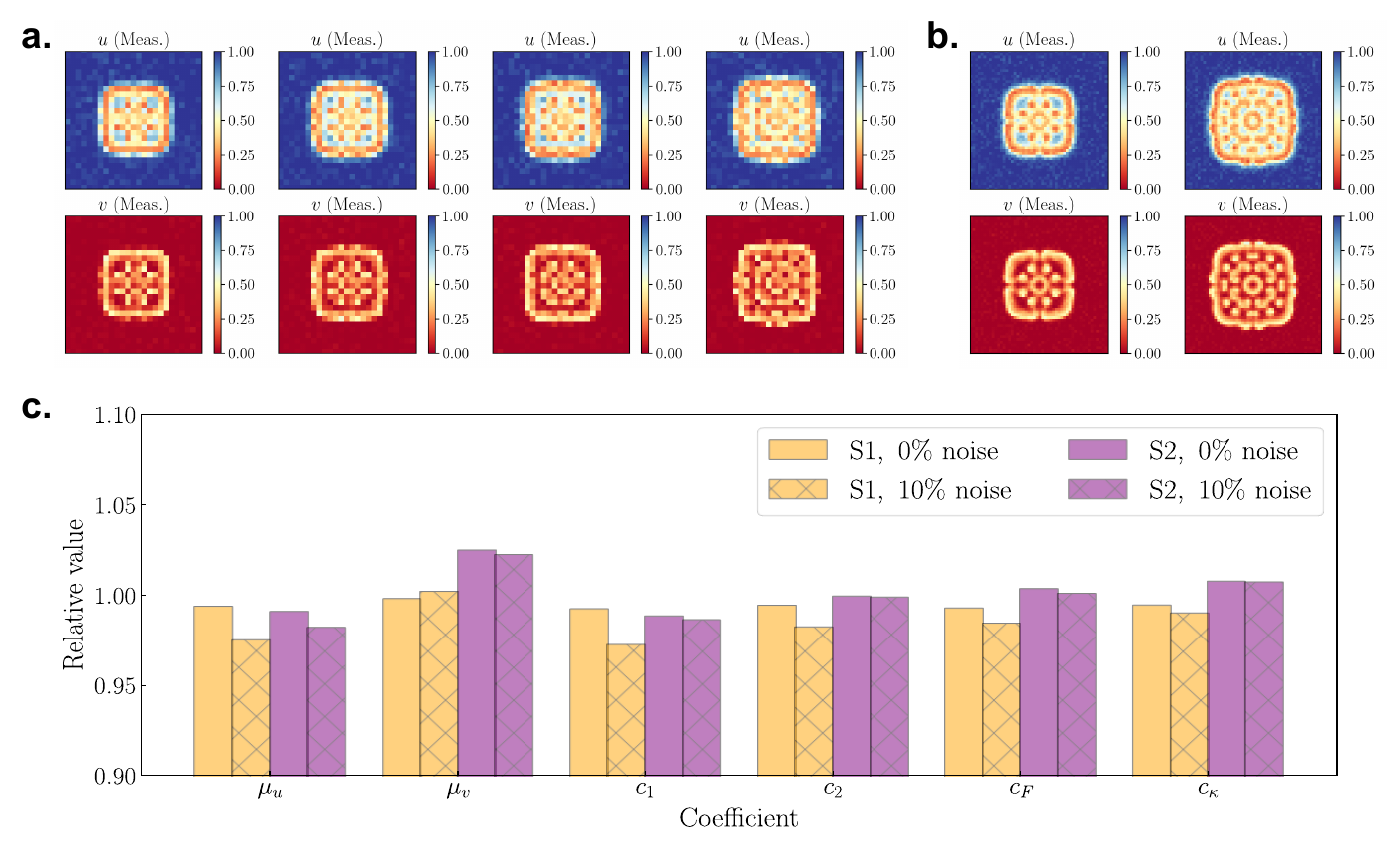}
\caption{Snapshots of the measurement data employed in the experiment and the identified coefficients. \textbf{a}, Data availability in Scenario 1 (multiple snapshots with $26\times26$ resolution). \textbf{b}, Data availability in Scenario 2 (initial and last snapshots with $51\times51$ resolution). \textbf{c}, Identified coefficient at various noise level and data availability. }
\label{fig:sys_id_result}
\end{figure}

To examine the capability of the model in scenarios of various data availability, we consider two different sets of measurement data. In the first scenario (S1), the available measurement includes multiple noisy and low-resolution snapshots of the system, which means the available data is scarce spatially while somewhat abundant in the temporal dimension. In the second scenario (S2), we assume only the first and last snapshots of the system with decent resolution are available. These two scenarios reflect the trade-off between the spatial and temporal resolution of the existing measurement. These synthetic measurements accompanied with 10\% Gaussian noise are shown in Figure \ref{fig:sys_id_result}\textbf{a}-\textbf{b}. The misfit error between the prediction and the measurement data is computed as the loss function for optimizing the unknowns. To prevent the overfitting to noise, early stopping is employed by splitting the dataset into training/validation sets. The details of computational parameters for dataset generation, network architecture, initialization of coefficients and optimization settings are presented in \textcolor{blue}{Supplementary Note D}. 

To \hsedit{test} the effects of noise, the experiment is also performed on clean data. Each case encompasses 10 runs with various random seed for coefficient initialization. The identified coefficients are presented in Figure \ref{fig:sys_id_result}\textbf{c} and Extended Data Table \ref{tb:table_coef_id}. It can be seen that, in all cases, PeRCNN is able to uncover the unknown coefficients with satisfactory accuracy. Compared with the identified coefficients in the noise-free case, the result only deteriorates slightly with 10\% noise. In the absence of the noise, the identified coefficients feature high accuracy, with the mean absolute relative error (MARE) for all the coefficients being 0.6\%. For the case with 10\% noise, the MARE is 1.61\% in spite of 10\% Gaussian noise in the measurement. \hsedit{PeRCNN also shows superiority to PINN (see the result reported in \textcolor{blue}{Supplementary Table S.4}).} \hseditr{Moreover, the potential of employing PeRCNN to identify space-varying coefficients is further demonstrated in \textcolor{blue}{Supplementary Note D.5}).} The numerical results in this section illustrates the good capability of our model on the inverse analysis of PDE systems.

\subsection*{Data-driven Modeling of Spatiotemporal Dynamics}
PDEs play an an essential role in modeling physical systems. However, there still exist a considerable portion of systems, such as those in epidemiology, climate science and biology, whose underlying governing PDEs are either completely unknown or only partially known. Thanks to the ever-growing data availability as well as recent advances in scientific machine learning, data-driven modeling nowadays becomes an effective way to establish predictive models for physical systems. Since the proposed network is characterized with excellent expressiveness for representing nonlinear dynamics (see \textcolor{blue}{Methods: Universal Polynomial Approximation Property} \textcolor{blue}{\text{for $\Pi$-block}}) and capability of encoding incomplete governing PDE, it has great potential to serve as a generalizable and robust data-driven model for predicting high-resolution nonlinear spatiotemporal dynamics. In this part, we primarily focus on data-driven modeling of spatiotemporal dynamics by the proposed physics-encoded learning paradigm given limited, noisy measurement data. 

Let us assume some low-resolution and potentially noisy snapshots of the system are measured, i.e., $\Tilde{\mathbf{u}}\in\mathbb{R}^{n_t'\times n\times H'\times W'}$ where $n_t'$ is the number of snapshots, $n$ is the number of state variable components and $H'\times W'$ is the resolution of each snapshot. We seek to establish a predictive model that gives the most likely high-resolution solution $\boldsymbol{\widehat{\mathcal{U}}}\in \mathbb{R}^{n_t\times n\times H \times W}$ where $n_t'<n_t$, $H'<H$ and $W'<W$, and possesses satisfactory extrapolation ability over the temporal horizon (e.g., for $t > t_{n_t}$). As we have seen, one salient characteristic of PeRCNN is the capability of encoding prior knowledge (e.g., the general PDE structure and/or I/BC) into the learning model. In particular, we assume the basic PDE form as shown in Eq. \eref{eq:gs_eqn} is given, where the diffusion term is known \textit{a priori} whose coefficients are however unknown. To compare PeRCNN with existing methods, we also perform experiments on several baseline models, namely, the recurrent ResNet \cite{liao2016bridging, zhang2017deep}, ConvLSTM \cite{shi2015convolutional}, PDE-Net \cite{long2018pde} and deep hidden physics model (DHPM) \cite{raissi2018deep}. Once the training is done, we infer the high-resolution prediction from the predictive data-driven models. The accumulative rooted-mean-square error (RMSE) and the physics residual (see \textcolor{blue}{Supplementary Note E.3} for definition) are utilized to evaluate the accuracy of the established data-driven models. 

We verify the performance of PeRCNN using synthetic datasets of the 2D and 3D GS RD equation systems, whose computational parameters and discretization setting are provided in Extended Data Table \ref{tb:data_set}. In the experiments, we fix the amount of data, training/validation/testing dataset splitting, the number of prediction steps, the Gaussian noise level (10\%) and the random seed for each method. The hyperparameters for each case are selected through hold-out cross validation. The synthetic measurement data (i.e., some low-resolution snapshots) is downsampled (in both spatial and temporal dimension) from the numerical solution. Once the model is finalized, extrapolation beyond temporal horizon, e.g., for $t > t_{n_t}$, would be performed to examine the extrapolation ability of each model. Note that {\color{black}comprehensive sensitivity tests} of PeRCNN in the context of some major hyperparameters (i.e., filter size, number of Conv layers and number of channels) are presented in \textcolor{blue}{Supplementary Note E.4}.

\begin{figure}[t!]
\centering
\includegraphics[width=0.99\textwidth]{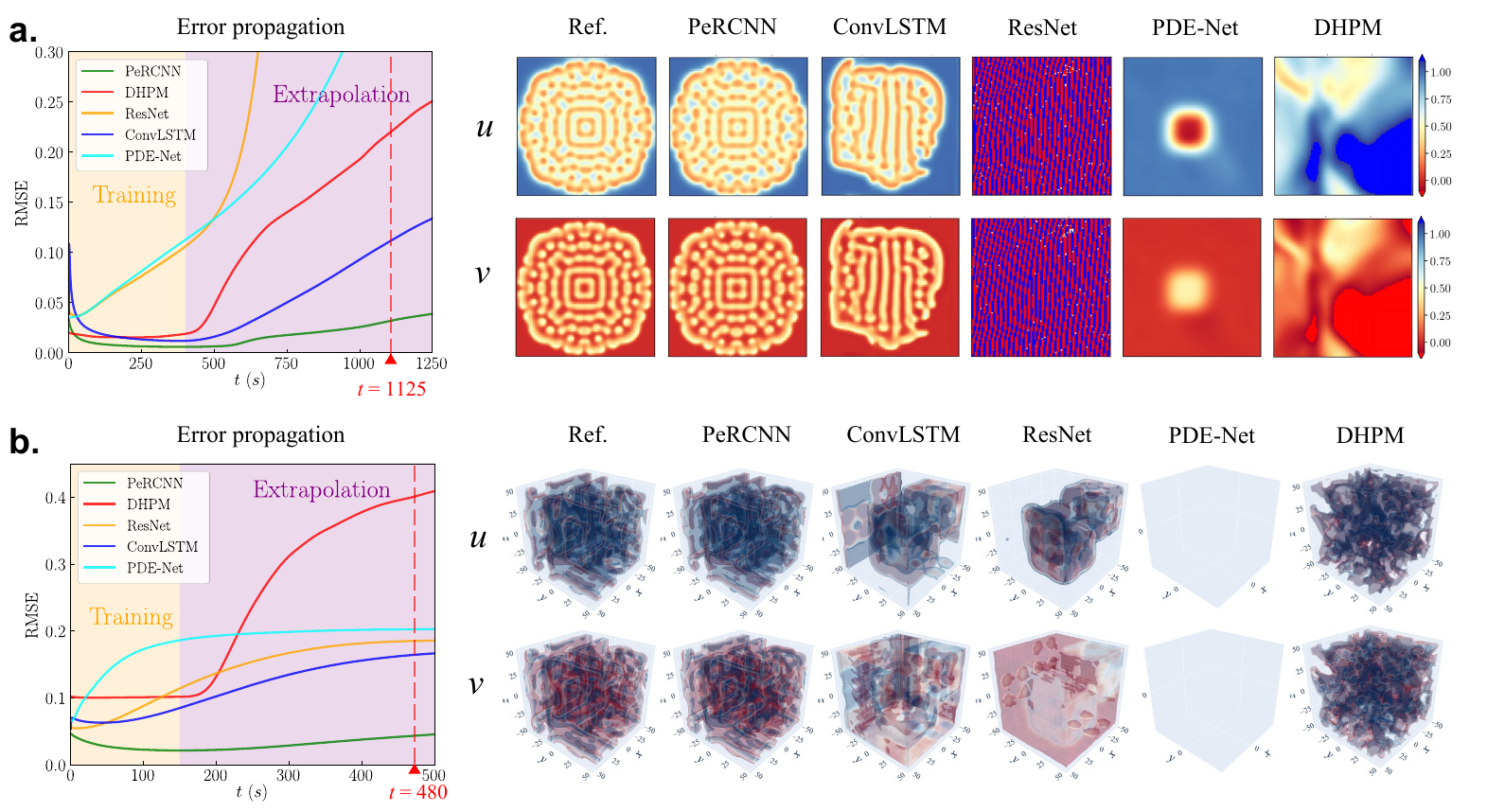}
\caption{Error propagation curve of the prediction (left) and the extrapolated snapshots (right) from each data-driven model compared with the reference solution. Note that extrapolation is performed beyond the time horizon of the measurement data while the time of each snapshot is marked by a dash red line in the error propagation curve. \textbf{a}, 2D GS RD equation. \textbf{b}, 3D GS RD equation.}
\label{fig:diagram_data_driven_result}
\end{figure}

\paragraph{\textit{2D Gray-Scott RD equation}.}
In this case, we consider a data availability scenario where the resolution of measurement data is relatively low in space but decent in time. The available measurement data in this case encompasses 41 noisy snapshots of on a $26\times 26$ grid, ranging from $t=0$ to $400$ s. Since we assume the dynamical system of interest features the ubiquitous diffusion phenomenon, the governing PDE (i.e., $\mathcal{F}$) is known to have a diffusion term ($\Delta \mathbf{u}$) whose scalar coefficients are still unknown. Therefore, we encode the diffusion term into PeRCNN by creating a physics-based FD Conv connection with the discrete Laplace operator as the Conv kernel (see \textcolor{blue}{Supplementary Note B.3}). Furthermore, the diffusion coefficient ($\Tilde{\mu}$) is first estimated by solving a linear regression problem of $u_t=\Tilde{\mu}\Delta u$ with the available data. Then a lower bound of $0$ and upper-bound of $2\Tilde{\mu}$ are applied to ensure the stability of diffusion. Each model is responsible for predicting 801 fine-resolution snapshots during the training phase, while 1700 extra snapshots are predicted for extrapolation. 

Snapshots at different time instants are presented in Figure \ref{fig:diagram_data_driven_result}\textbf{a}, which reveal the complex maze-like pattern of the GS-RD system. We report that the recurrent ResNet and PDE-Net are unable to reconstruct the fine-resolution snapshots even in the training due to the limited noisy training data, after trying all the hyperparameter combinations within a range (see \textcolor{blue}{Supplementary Note E.5.1}). Apart from that, it can be seen from the snapshots that PeRCNN is the only model working well for long-time extrapolation in spite of minor discrepancies. It is also interesting to note that PeRCNN with filter size of 1 works as good as the model with larger filters (e.g., 3 or 5). This is because the reaction term of the GS-RD system contains no spatial derivatives, making the $1\times1$ filter sufficient for representing the nonlinear reaction terms. This implies that prior knowledge on the governing PDE can also be employed while designing the data-driven model. To quantitatively measure the extrapolation capability of our model, we also plot the evolution of accumulative RMSE in Figure \ref{fig:diagram_data_driven_result}\textbf{a}. It is observed that PeRCNN outperforms the competitors at all stages in the context of error propagation, which further confirms the extrapolation ability of PeRCNN. We may notice that the accumulative RMSE starts from an initial high value. This is due to the fact that the training data is corrupted by 10\% Gaussian noise and the metric is computed from one single snapshot at the beginning. The effect of the unrelated noise gradually fades out as more time steps are considered. 

\paragraph{\textit{3D Gray-Scott RD equation}.}
In this example, we test our method on the 3D GS RD system. As the computational intensity of this higher dimensional example brings challenges to the existing methods, we aim to scrutinize the performance of our PeRCNN regarding the scalability and computational efficiency. The training data used to establish the data-driven model includes 21 noisy low-resolution snapshots ($25^3$) uniformly sampled from $t=0$ to $150$ s. The prior knowledge on the system and the estimation of the diffusion coefficients as discussed in the previous 2D GS RD example are adopted here as well. Each trained model produces 301 high-resolution ($49^3$) snapshots during the learning stage, while 700 extrapolation steps are predicted once each model is finalized. The predicted isosurfaces of two levels are plotted in Figure \ref{fig:diagram_data_driven_result}\textbf{b}. {\color{black}It should be noted that the plot of PDE-Net is blank because the prediction range falls out of the two selected isosurface levels.} Similar to the previous case, we observe that PeRCNN is the only model that gives a satisfactory long-term prediction. The flat error propagation curve of PeRCNN, as shown in Figure \ref{fig:diagram_data_driven_result}\textbf{b}, also demonstrates the remarkable generalization capability of PeRCNN. 

In \textcolor{blue}{Supplementary Note E.5}, we compare the number of trainable parameters, the training time per epoch, and the RMSE of both training and extrapolation for each model. It is found that PeRCNN is characterized with good model efficiency as it uses {\color{black}the} least amount of training parameters. For the 3D case where the training efficiency of the network is of great concern, the elapsed time for training one epoch by PeRCNN is comparable to that of the ResNet, which is widely acknowledged to be an efficient network architecture. As for the accuracy of the training and extrapolation, our model outperforms the baselines consistently across different examples. In a nutshell, PeRCNN outperforms the other three baselines with {\color{black}many fewer} trainable parameters and higher accuracy.  

\begin{figure}[t!]
\centering
\includegraphics[width=1.0\textwidth]{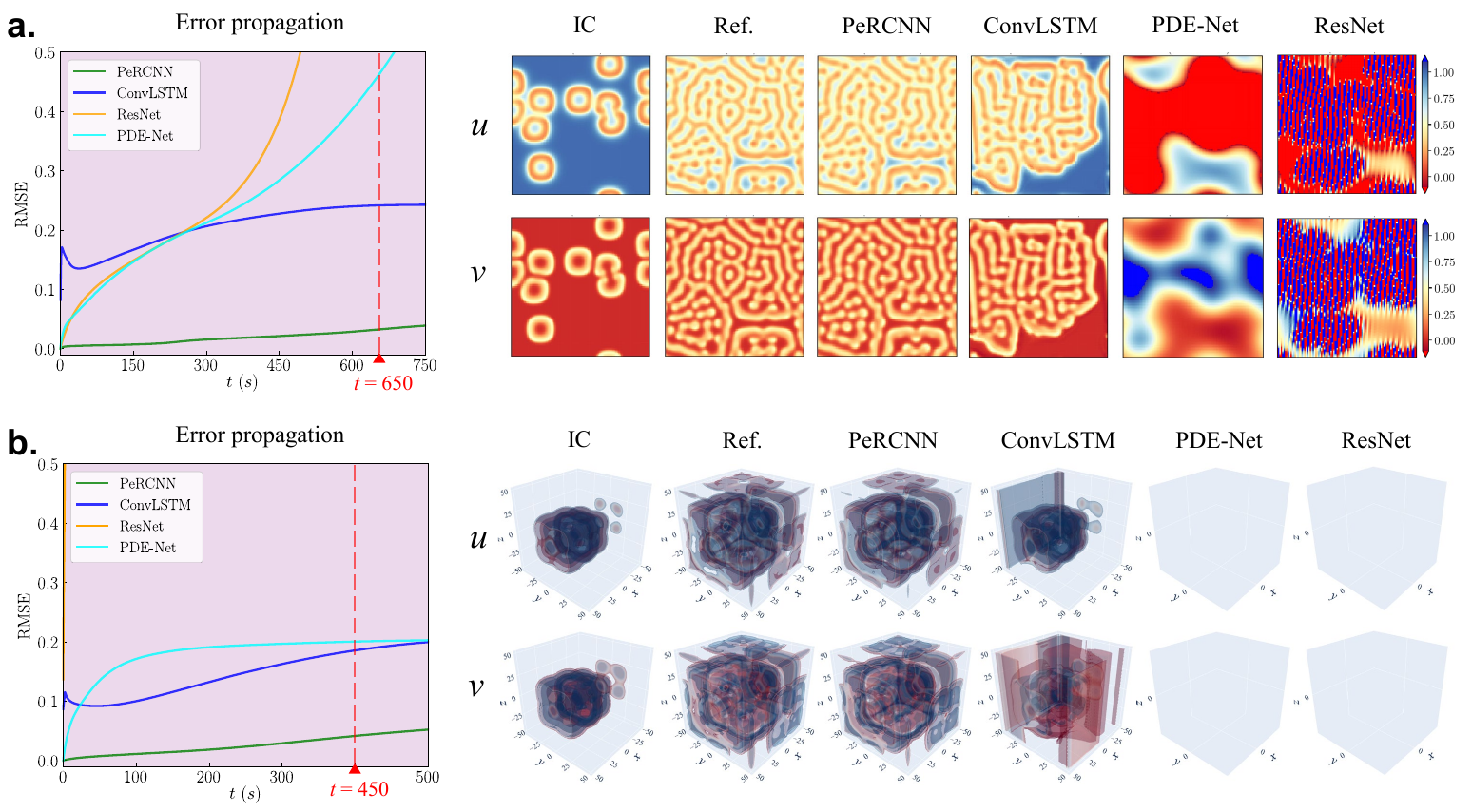}
\caption{Error propagation curve (left) and the snapshots (right) of the inference result. The results are obtained by performing inference on the data-driven model from subsection \textcolor{blue}{Data-driven Modeling} using a different initial condition.  \textbf{a}, 2D GS RD equation. \textbf{b}, 3D GS RD equation.}
\label{fig:diagram_generalization_result}
\end{figure}

\paragraph{\textit{Generalization to Different ICs}.} It is evident that the trained model has good extrapolation capability along the time horizon. Here, we further explore how the trained model generalizes to different initial conditions (ICs). To set up the experiment, we employ the above trained model to perform inference with a different IC. It should be noted that the baseline DHPM is ineligible for inference with different ICs as it is based on FCNN. The prediction result is depicted in Figure \ref{fig:diagram_generalization_result}. It is seen that PeRCNN gives consistent prediction compared with the ground truth reference solution. On the contrary, the considered baseline models (e.g., recurrent ResNet, ConvLSTM and PDE-Net) are unable to generalize to an unseen IC. They give wild prediction because of their incapability of learning the underlying physics (e.g., caused by the black-box property of the model). In addition, the error propagation of the prediction in Figure \ref{fig:diagram_generalization_result} indicates clearly the excellent generalization capability of the proposed model. In the later subsection of \textcolor{blue}{Interpretability of the Learned Model}, we show the extracted expression from the trained PeRCNN model is very close to the genuine $\mathcal{F}$, which to a large degree explains the remarkable generalization capability of our model given the fact that the trained PeRCNN model parameterizes the spatiotemporal dynamics well. 

\subsection*{Data-driven Discovery of PDEs with Scarce and Noisy Data}
\begin{figure}[t!]
\begin{center}
    \includegraphics[width=1.0\textwidth]{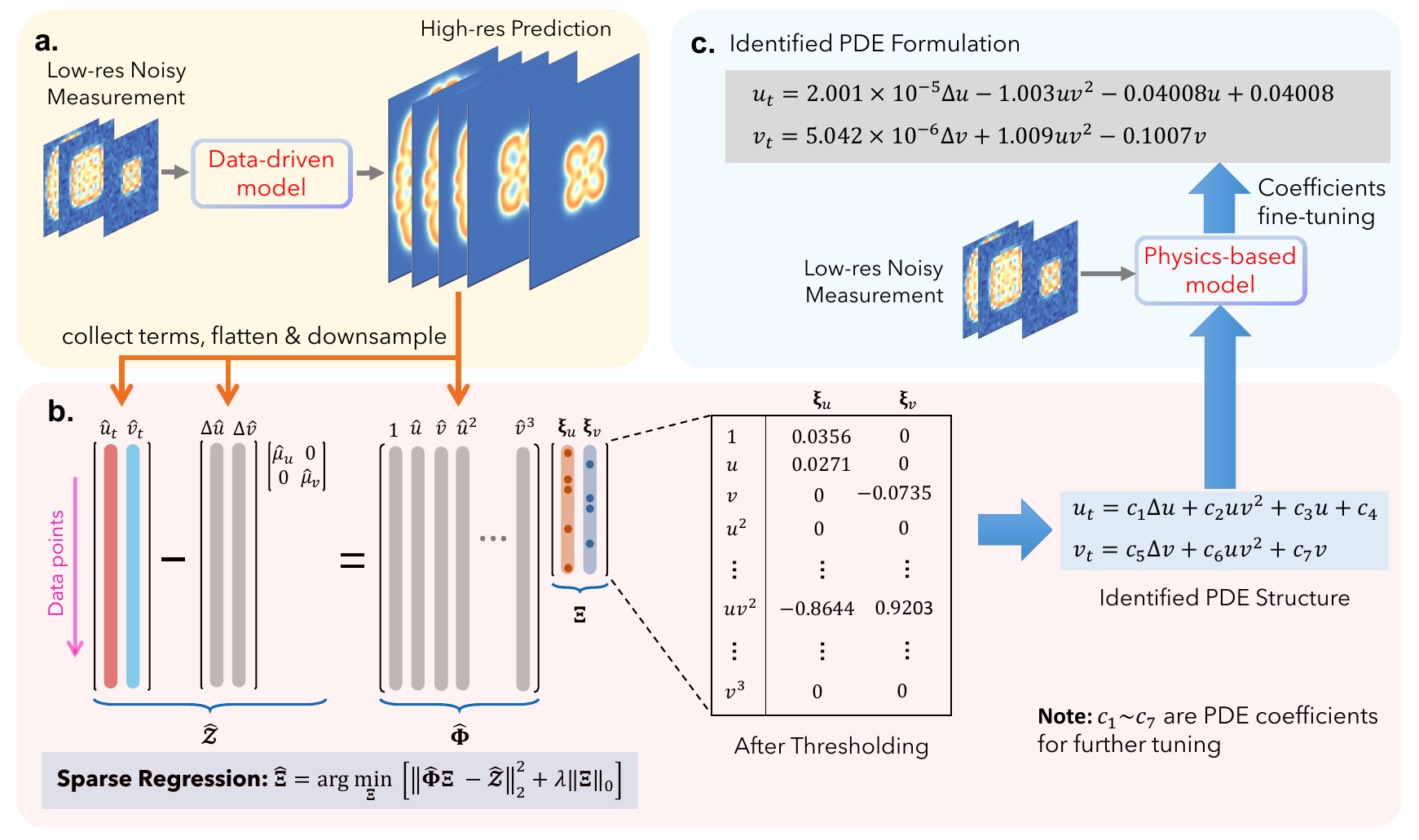}
    \vspace{-8pt}
    \caption{Flowchart of the discovery of governing PDEs. \textbf{a. data reconstruction:} data-driven model constructed from low-res and noisy measurement is used to generate high-res prediction for sparse regression; \textbf{b. sparse regression:} STRidge algorithm is used to obtain the sparse coefficient matrix $\boldsymbol{\Xi}$; \textbf{c. coefficients fine-tuning:} the PeRCNN built based on identified PDE structure is employed to fine-tune the coefficient from the sparse regression. }
    \label{fig:diagram_eqn_discover}
\end{center}
\end{figure}

In previous sections, we primarily investigate the scientific modeling tasks (e.g., forward simulation or data-driven modeling) using the proposed model, which exhibits excellent accuracy and extrapolation ability as identified from the numerical results. However, the process of knowledge discovery does not end at modeling the physical phenomena of interests. More importantly, it is the translation of the learned patterns from the data (e.g., formulated PDEs or empirical relationships) that lead scientists to understand the cause-effect relationship among physical variables, and further make inference on similar problems. Therefore, in this section, we extend the proposed physics-encoded learning model for discovering the closed-form governing PDEs \cite{rao2022discovering}. To formulate this problem, let us again consider the nonlinear system described by Eq. \eref{eq:dynamic_system_pde}. The objective of the equation discovery is to recover the closed form of the governing PDEs given the scarce and noisy measurement of the system. To this end, we integrate the sparse regression technique \cite{brunton2016discovering} with our PeRCNN model for solving this problem. The proposed framework for the PDE discovery is presented in Figure \ref{fig:diagram_eqn_discover} with the example of 2D GS-RD equation. The entire procedure consists of three steps, including data reconstruction (Figure \ref{fig:diagram_eqn_discover}\textbf{a}), sparse regression (Figure \ref{fig:diagram_eqn_discover}\textbf{b}) and coefficients fine tuning (Figure \ref{fig:diagram_eqn_discover}\textbf{c}), as discussed in \textcolor{blue}{Methods: Equation Discovery}.

To validate the effectiveness of the our method, we perform the equation discovery on two RD systems (e.g., 2D GS and $\lambda$--$\Omega$ RD systems) using synthetic datasets, which are obtained by downsampling the noise-corrupted numerical solution. Two different Gaussian noise levels (5\% and 10\%), as well as the noise-free case, are considered in the experiment. As we assume the ubiquitous diffusion phenomenon exists in the concerned system, a shortcut diffusion Conv layer is encoded into the network for data reconstruction (see Figure \ref{Diagram}\textbf{a}). Accordingly, the coefficients corresponding to $\Delta (\mathbf{u})$ are exempted from being filtered in the STRidge algorithm. Once the equation discovery is finished, we measure the performance of the proposed method using the metrics of precision, recall and relative $\ell_2$ error of the coefficient vector. Technical details of the generation of synthetic measurement data and the evaluation metrics can be found in \textcolor{blue}{Supplementary Note F.3}. 

The discovered PDEs by our method are provided in Extended Data Table \ref{tb:result_pde_discovery}. It is seen that our method is able to recover the governing PDEs completely when the measurement data is clean or mildly polluted by noise. Even though the noise level grows to 10\%, our approach still exhibits competitive performance, i.e., it uncovers the majority of terms in the PDEs. Empirical study in \textcolor{blue}{Supplementary Note F.5} shows that our method could handle even a much larger noise level, i.e., 30\% Gaussian noise. In \textcolor{blue}{Supplementary Note F.4}, we also compare our approach with some existing methods (or baselines) for governing PDE discovery, including PDE-FIND \cite{rudy2017data}, sparse regression coupled with fully-connected neural network (FCNN) or PDE-Net \cite{long2018pde}. The comparison shows that our approach outperforms (if not perform as good as) the considered baselines consistently under different noise levels and data richness. Visualizing the reconstructed high-fidelity data from each method (see \textcolor{blue}{Supplementary Figure S.17}), we observe that our method has a much smaller reconstruction error as a result of fully utilizing the prior physics knowledge and the powerful expressiveness of the model. This would give rise to more accurate derivative terms in the linear system, facilitating the discovery of the governing PDEs. In summary, the effectiveness of the proposed {\color{black}approach} is {\color{black}demonstrated} for solving the data-driven equation discovery problem, especially when the measurement data is characterized with poor resolution and noise. 

\subsection*{Interpretability of the Learned Model}\label{Interpretability}
Compared with the traditional deep neural networks, which are usually considered to be ``black-box'', the proposed network architecture is designed to possess good interpretability. As each channel of the input for $\Pi$-block (i.e., $\widehat{\boldsymbol{\mathcal{U}}}^{(k)}$) corresponds to a solution component (i.e., $u$ and $v$), the multiplicative form of $\Pi$-block (see Eq. \eref{eq:update_rule}) makes it possible to extract an explicit form of learned $\mathcal{F}$ from the learned weights and biases via symbolic computations. This section is dedicated to the discussion on how the learned model can be interpreted as an analytical expression, which is useful for people to understand the underlying cause-effect relationships among physical variables. 

To demonstrate how to interpret the learned model, we first use the learned model from 3D GS RD case in \textcolor{blue}{Data-driven Modeling} as an example. In this case, the parallel Conv layers in $\Pi$-block have the filter size of 1, which implies that each output channel is the linear combination of $u$, $v$ and a constant. With the element-wise product operation among three Conv layers, a third degree polynomial will be produced to account for the reaction term of the system. With the help of the $\texttt{SymPy}$ \cite{sympy} -- a symbolic computation Python package, we can extract the learned reaction term:
\begin{equation}
    \label{eq:factor_out_terms}
    \resizebox{0.99\linewidth}{!}{$
    \mathbf{R(u)}=
    \begin{bmatrix}
    -0.0074u^3 - 0.0051u^2v - 0.2uv^2 - 0.0386v^3 - 0.0018u^2 - 0.11uv- 0.055v^2 - 0.016u  - 0.022v + 0.025 \\
    0.0005u^3 - 0.013u^2v + 0.54uv^2 - 0.087v^3 - 0.0076u^2 + 0.023uv + 0.046v^2 + 0.017u - 0.036v - 0.0097
    \end{bmatrix} 
    $}
\end{equation}
In the meanwhile, the identified diffusion term can be also extracted from the trainable variables in diffusion connections, which reads
$\mathbf{D(u)}=
\begin{bmatrix}
0.18\Delta u,~
0.08\Delta v
\end{bmatrix}^\text{T} $.
Comparing the extracted term with the ground truth PDEs, we observe some distracting terms due to the 10\% noise in the training data and the redundancy of the network. Further pruning on the raw expression can be done to make it more parsimonious. 

The above example is a special case where the Conv layer in $\Pi$-block has filter size of 1, which indicates no spatial derivatives are involved in reaction terms. However, we can extend the network architecture design to make it applicable to general cases. To interpret terms involving partial derivatives (e.g., $u\Delta u$, $uu_x$), we could completely freeze or impose moment matrix constraints on part of the convolutional filters \cite{long2018pde}. Here, an experiment is conducted on the 2D Burgers' equation, which has wide applications in applied mathematics such as fluid/traffic flow modeling, given by $\mathbf{u}_t+ \mathbf{u} \cdot \nabla \mathbf{u} = \nu \Delta \mathbf{u}$, where $\mathbf{u}=[u,~v]^\texttt{T}$ denotes the fluid velocities and $\nu$ is the viscosity coefficient. The network employed in the experiment has two Conv layers with two channels. The first Conv layer are associated with derivative operators $\partial (\cdot)/\partial x$ and $\partial (\cdot)/\partial y$ respectively, by fixing the filters with corresponding FD stencils. The synthetic dataset is generated on $101\times 101$ Cartesian grid using high-order FD method with $\nu=0.005$. The noise-free synthetic measurement data used for constructing the model includes 11 low-resolution ($51\times 51$) snapshots uniformly selected from the time period of $t\in[0,~0.1]$. 

After the {\color{black}model} is trained, we interpret the expression from the PeRCNN model, which reads:
\begin{equation}
    \label{eq:burgers_terms}
    \resizebox{0.9\linewidth}{!}{$
    \mathbf{u}_t=
    \begin{bmatrix}
        \begin{aligned}
    &0.0051 \Delta u - 0.95u_x(1.07u - 0.0065v - 0.17) + 0.98u_y(0.0045u - 1.01v + 0.17) + 0.053 \\
    &0.0051 \Delta v -0.82v_x(1.22u + 0.0078v - 0.18) - 0.91v_y(0.0063u + 1.08v - 0.17) + 0.058
        \end{aligned} 
    \end{bmatrix} 
    $}
\end{equation} 
It can be observed that the equivalent expression of the learned model matches well the genuine governing PDEs, except some minor terms whose coefficients are close to zero. In addition, the extracted expression helps explain the extraordinary extrapolation and generalization capabilities of our model. Although the selection of differential operators to be embedded is crucial for identifying the genuine form of the $\mathcal{F}$, the above two examples demonstrate the interpretability of PeRCNN over common ``black-box'' models.



\section*{DISCUSSION}
This paper introduces a novel deep learning architecture, namely PeRCNN, for modeling and discovery of nonlinear spatiotemporal dynamical systems based on sparse and noisy data. One major advantage of PeRCNN is that the prior physics knowledge can be encoded into the network, which guarantees the resulting network strictly obeys given physics (e.g., initial and boundary conditions, general PDE structure, and known terms in PDEs). This brings distinct benefits for improving the convergence of training and accuracy of the model. Through extensive numerical experiments, we show the efficacy of PeRCNN for forward and inverse analysis of reaction-diffusion type PDEs. The comparison with several baseline models demonstrates that the proposed physics-encoded learning paradigm uniquely possesses remarkable extrapolation ability, generalizability and robustness against data noise/scarcity. \hsedit{Although we demonstrate the effectiveness of the PeRCNN on various reaction-diffusion systems, the model is in theory applicable to other types of spatiotemporal PDEs (e.g., the 2D Burgers' equation with the convection term shown in \textcolor{blue}{Supplementary Note F.4}, \hseditr{and the Kolmogorov turbulent flows at Reynolds number = 1,000 discussed in \textcolor{blue}{Supplementary Note J}}).}

Equally important, PeRCNN shows good interpretability due to the multiplicative form of the $\Pi$-block. An analytical expression that governs the underlying physics can be further extracted from the learned model via symbolic computation. In particular, we successfully marry PeRCNN to the sparse regression algorithm to solve the crucial PDE discovery issues. The coupled scheme enables us to iteratively optimize the network parameters, and fine-tune the discovered PDE structures and coefficients, essentially leading to the final parsimonious closed-form PDEs. The resulting framework will serve as an effective, interpretable and flexible approach to accurately and reliably discover the underlying physical laws from imperfect and coarse-meshed measurements. 

Although PeRCNN shows promise in data-driven modeling of complex systems, it is restricted by the computational bottleneck due to the high dimensionality of the discretized system, especially when it comes to systems in a large 3D spatial domain with long-term evolution. However, this issue is expected to be addressed via temporal batch and multi-GPU training. In addition, the current {\color{black}model} is rooted in standard convolution operations, which limits its applicability to irregular meshes of arbitrary computational geometries. This {\color{black}issue} might be resolved by introducing graph convolution into the network architecture. Lastly, since the PeRCNN network is designed based on the assumption that the underlying governing PDEs have a polynomial form (very commonly seen in standard PDEs for modeling of physics such as diffusion, reaction, convection, and rotation), it might be less capable or too redundant (if many channels are used to achieve a high polynomial degree) of modeling unique spatiotemporal dynamics whose governing PDEs are parsimonious but involves other advanced symbolic operators such as division, sin, cos, exp, tan, sinh, log, etc. \hseditr{Although PeRCNN shows success in data-driven modeling of a PDE system with non-polynomial term in \textcolor{blue}{Supplementary Note I}, how} to design a network that properly incorporates a limited number of mathematical operators as symbolic activation functions to improve the representation ability \hseditr{still remains} an open question. We aim to systematically address these issues in our future study.

\section*{METHODS}
We herein introduce the method of the proposed PeRCNN model. More details can be found in \textcolor{blue}{Supplementary Note B}.

\subsection*{Network Architecture}
Let us first consider a spatiotemporal dynamical system described by a set of nonlinear, coupled PDEs as
\begin{equation} 
    \label{eq:dynamic_system_pde} 
    \mathbf{u}_t=\mathcal{F}\left(\mathbf{x}, t, \mathbf{u}, \mathbf{u}^2, \nabla_\mathbf{x}  \mathbf{u}, \mathbf{u}\cdot\nabla_\mathbf{x} \mathbf{u}, \nabla^2 \mathbf{u}, \cdots\right)
\end{equation}
where $\mathbf{u}(\mathbf{x},t)\in\mathbb{R}^n$ denotes the state variable with $n$ components defined over the {\color{black}spatiotemporal domain} $\{(\mathbf{x}, t)\} \in \Omega\times\mathcal{T}$. Here, $\Omega$ and $\mathcal{T}$ represent the spatial and temporal domain, respectively; $\nabla_{\mathbf{x}}$ is the Nabla operator with respect to the spatial coordinate $\mathbf{x}$; and $\mathcal{F}(\cdot)$ is a nonlinear function describing the right hand side (RHS) of PDEs. The solution to this problem is subject to the initial condition (IC)  $\mathcal{I}(\mathbf{u};t=0,\mathbf{x} \in \Omega)=0$ and the boundary condition (BC) $\mathcal{B}(\mathbf{u},\nabla_{\mathbf{x}}\mathbf{u}, \cdots;\mathbf{x} \in \partial \Omega)=0$. Since we mainly focus on regular physical domains in this paper, state variable $\mathbf{u}$ are defined on a discretized Cartesian grid. 

Borrowing the concepts of numerical discretization, we build the physics-encoded spatiotemporal learning model on the basis of a forward Euler scheme. That said, the state variable $\mathbf{u}$ would be updated by a recurrent network given by
$ \widehat{\mathbf{u}}^{(k+1)}=\widehat{\mathbf{u}}^{(k)} + \widehat{\mathcal{F}}(\widehat{\mathbf{u}}^{(k)};\boldsymbol{\theta})\delta t$,
where $\delta t$ is the time spacing, $\widehat{\mathbf{u}}^{(k)}$ is the prediction at time $t_k$ and $\widehat{\mathcal{F}}$ is the approximated $\mathcal{F}$ parameterized by $\boldsymbol{\theta}$ that ensembles a series of operations for computing the RHS of Eq. \eref{eq:dynamic_system_pde}. Similar {\color{black}ideas} of applying numerical discretization (e.g., backward Euler or Runge-Kutta) to designing deep learning architectures can be found in some recent literature \cite{he2016deep, chen2015learning, long2018pde, lu2018beyond, ruthotto2019deep, larsson2016fractalnet}.

Following the above intuition, here we introduce the proposed network, namely the Physics-encoded Recurrent Convolutional Neural Network (PeRCNN). The architecture of this network \hsedit{(as shown in Figure \ref{Diagram})} consists of two major components: a fully convolutional (Conv) network as initial state generator (ISG) and an novel Conv block called $\Pi$-block (product) used for recurrent computation. ISG is introduced to produce the high-resolution initial state $\widehat{\boldsymbol{\mathcal{U}}}^{(0)}$ in the case that only low-resolution initial state (or measurement) $\Tilde{\mathbf{u}}^{(0)}$ is available as initial condition. Note that $\Tilde{\mathbf{u}}$ is used to denote the low-resolution snapshots (or measurement) while the {\color{black}superscript} ``0'' indicates the first one. Similarly, $\widehat{\boldsymbol{\mathcal{U}}}$ is used to represent the high-resolution prediction from the model. Within the $\Pi$-block, the core of PeRCNN, the state variable $\widehat{\boldsymbol{\mathcal{U}}}^{(k)}$ from the previous time step goes through multiple parallel Conv layers. The feature maps produced by these layers are then {\color{black}fused} through an elementwise product layer. The $1\times 1$ Conv layer is subsequently used to linearly combine multiple channels into the desired output (i.e., approximated $\mathcal{F}$). Mathematically, the $\Pi$-block seeks to approximate the function $\mathcal{F}$ via polynomial combination of solution $\widehat{\boldsymbol{\mathcal{U}}}^{(k)}$ and its spatial derivatives, given by
\begin{equation} 
    \label{eq:update_rule} 
   \widehat{\mathcal{F}}\left (\widehat{\boldsymbol{\mathcal{U}}}^{(k)}\right)=\sum_{c=1}^{N_c} W_{c} \cdot \left [ \prod_{l=1}^{N_l} \left ( \mathcal{K}_{c,l} \circledast \widehat{\boldsymbol{\mathcal{U}}}^{(k)} + b_{l}\right ) \right ]
\end{equation}
where $N_c$ and $N_l$ are the numbers of channels and parallel Conv layers respectively; $\circledast$ denotes the Conv operation;  $\mathcal{K}_{c,l}$ denotes the weight of Conv filter of $l$-th layer and $c$-th channel, while $b_{l}$ represents the bias of $l$-th layer; $W_c$ is the weight corresponding to $c$-th channel in $1\times 1$ Conv layer while the bias is omitted for simplicity. This multiplicative representation promotes the network expressiveness for nonlinear functions $\mathcal{F}$, compared with the additive representation commonly seen in related work \cite{long2018pde, guen2020disentang}. For detailed discussion on the design of $\Pi$-block, please refer to \textcolor{blue}{Supplementary Note B}. 

Due to the discretized scheme of the learning model, it is possible to encode prior physics knowledge of the system into network architecture, which contributes to a well-posed optimization problem. Given some existing terms in the PDE, we could encode these terms into the network by creating a shortcut connection, namely the physics-based {\color{black}finite difference (FD)} Conv connection, from $\widehat{\boldsymbol{\mathcal{U}}}^{(k)}$ to  $\widehat{\boldsymbol{\mathcal{U}}}^{(k+1)}$ as shown in Figure \ref{Diagram}. The convolutional kernel in this physics-based Conv layer would be fixed with the corresponding FD stencil to account for the known terms. A major {\color{black}advantage} of this encoding mechanism over the soft penalty in physics-informed learning {\color{black}models} is the capability to leverage the incomplete PDE in the learning. In the numerical examples, we demonstrate that such a highway connection could accelerate the training speed and improve the model inference accuracy significantly. In a nutshell, the physics-based Conv connection is built to account for the known physics, while the $\Pi$-block is designed to learn the complementary unknown dynamics. 

In addition to the incomplete PDE, the boundary conditions (e.g., Dirichlet or Neumann type) can also be encoded into the learning model. Inspired by the idea from the FD method, we apply the physics-based padding to the model's prediction at each time step, as shown by Figure \ref{Diagram}\textbf{b}. Specifically, for the Dirichlet BCs, we pad the prediction with prescribed values. Likewise, the padding value of \hsedit{Neumann or Robin} BCs will be computed based on the boundary values and the gradient information. \hsedit{A comprehensive discussion on the padding mechanism for various BCs (e.g., Dirichlet, Neumann, Robin and Periodic) can be found in \textcolor{blue}{Supplementary Note B.3}. In particular, we show the effectiveness of the proposed padding method in \textcolor{blue}{Supplementary Note G}, where the Neumann BCs are considered for example.}

\subsection*{Motivation of the Network Architecture Design}
In the subsection of \textcolor{blue}{Physics-encoded Spatiotemporal Learning}, we have introduced the proposed network architecture for learning spatiotemporal dynamical systems. Here a further discussion on the design philosophy is presented to showcase the primary motivations. A distinct characteristic of this architecture is the usage of $\Pi$-block as {\color{black}a} universal polynomial approximator to nonlinear {\color{black}functions}, instead of utilizing a sequence of linear layers intertwined with nonlinear activation layers commonly seen in traditional deep {\color{black}networks}. The motivations for introducing the elementwise product operation in $\Pi$-block are three-fold: 

\vspace{-6pt}
\begin{itemize}
\item Though the nonlinear activation function is crucial to the universal approximation property of the deep neural network (DNN), it is also a source of poor interpretability. For example, the conventional DNN would form a prolonged nested function that is usually intractable to {\color{black}humans}. We consider it unfavorable to use these nonlinear functions to build a recurrent block that aims to generalize the unknown physics. 
\vspace{-3pt}
\item The elementwise product operation makes a better approximation to $\mathcal{F}$ in the form of multivariate polynomial (e.g., $\mathbf{u}\cdot\nabla u+u^2v$), which covers a wide range of well-known dynamical systems, such as Navier-Stokes, reaction-diffusion (RD), Lorenz, Schr$\ddot{\text{o}}$dinger equations, to name only a few. Since the spatial derivatives can be computed by Conv filters \cite{cai2012image}, a $\Pi$-block with $n$ parallel Conv layers of appropriate filter size is able to represent a polynomial up to the $n^\text{th}$ order. 
\vspace{-3pt}
\item Compared with the regression models (e.g., linear or symbolic regression) relying on predefined basis functions or prior knowledge (e.g., the highest order) on $\mathcal{F}$ \cite{brunton2016discovering, long2018pde}, the $\Pi$-block is flexible at generalizing the nonlinear function $\mathcal{F}$. For example, a $\Pi$-block with 2 parallel layers of appropriate filter size ensembles a family of polynomials up to the 2$^\text{nd}$ order (e.g., $u$, $\Delta u$, $uv$, $\mathbf{u}\cdot\nabla u$), with no need to explicitly define the basis.
\end{itemize}
\vspace{3pt}

Since the network architecture roots on numerical discretization, nice mathematical properties (see the next subsection) exist to guarantee the universal polynomial approximation property of $\Pi$-block. In addition, the flexibility to deal with a variety of problems in scientific modeling is another {\color{black}advantage} possessed by the proposed network architecture. The $\Pi$-block acts as {\color{black}a} universal polynomial approximator to unknown nonlinear function while the physics-based Conv layer accounts for the prior knowledge on the governing equation. Such a way of encoding the prior physics knowledge into the network architecture could effectively narrow down the space of feasible model parameters, hence leading to the reduced training effort required (e.g. memory, FLOPs, etc.). Furthermore, in many prediction tasks involving nonlinear system, a mixture of partial physics knowledge and scarce amount of measurement data of the system is available, which is when the proposed PeRCNN has a huge advantage over the traditional deep network. As shown in the subsection of \textcolor{blue}{\text{Data-driven Modeling}}, we demonstrated the capability of proposed PeRCNN at handling the modeling tasks given limited physics knowledge and some low-resolution snapshots of the system. An extreme case is when the analytical form of a physical system is completely known except some scalar coefficients. The physics-based Conv layers associated with trainable variables can be created to exactly express the physical system up to discretization error. In such a case, the PeRCNN can be assumed to recover to FD method with some trainable variables. In the subsection of \textcolor{blue}{\text{Inverse Analysis of PDE Systems}}, we showed how the proposed PeRCNN can be applied to identify the system coefficients from the noisy and scarce data in such a scenario. 

Noteworthy, we mainly consider the nonlinear function $\mathcal{F}$ in the form of polynomial \hseditr{which is very commonly seen in PDEs}. Other terms such as trigonometric and exponential functions are not considered in this work for simplicity. However, incorporating them would require no more effort than adding a particular symbolic activation (e.g., sin, cos, exp, etc.) layer following the Conv operation. What's more, these functions can be approximated by polynomials based on Taylor series as argued in \cite{brunton2016discovering}.

\subsection*{Universal Polynomial Approximation Property for $\Pi$-block}

In the proposed PeRCNN, $\Pi$-block acts as an universal polynomial approximator to unknown nonlinear functions while the physics-based FD Conv layer (i.e., with FD stencil as the Conv filter) accounts for the prior knowledge on the governing equation. Notably, the $\Pi$-block achieves its nonlinearity through the element-wise product operation (see Eq. \eref{eq:update_rule}) which renders the network better expressiveness compared with the additive form representation $\widehat{\mathcal{F}}(\mathbf{u})=\sum_{1\le i\le N} f_{i} \cdot  \left ( \mathcal{K}_{i} \circledast \mathbf{u} \right )$ seen in related work \cite{long2018pde, guen2020disentang} where $N$ is the number of Conv layer, $\mathcal{K}_{i}$ is the Conv kernel of $i$th layer and $f_{i}$ is the weight of $i$th layer's output. To support this claim, we propose \hseditr{the following theorem and} Lemmas to prove that any dynamical system described by Eq. \eref{eq:dynamic_system_pde} whose $\mathcal{F}$ is \hseditr{continuous (e.g., preferably in the form of polynomial)} can be approximated by the proposed network. Without loss of generality, we consider the state variable $\mathbf{u}$ with one components $u$. \textit{Lemma 1} and \textit{Lemma 2} guarantee the accuracy of the approximation of $\widehat{\mathcal{F}}$ (see Eq. \eref{eq:update_rule}) and the forward computation 
$    \widehat{u}^{(k+1)}=\widehat{u}^{(k)} + \widehat{\mathcal{F}}(\widehat{u}^{(k)};\boldsymbol{\theta})\delta t
$ respectively. 

\textit{\textbf{Lemma 1:} \hseditr{Trainable} convolutional filter $\mathcal{K}$ can approximate any differential operator with prescribed order of accuracy. }

\textit{\textbf{Proof:}} Consider a bivariate differential operator $\mathcal{L}(\cdot)$, we have (reproduced from \cite{long2019pde}):
\begin{equation}
    \label{eq:lemma1}
    \begin{aligned}
    \mathcal{L}(u)&=\sum_{k_1,k_2=-\frac{p-1}{2}}^{\frac{p-1}{2}} \mathcal{K}[k_1,k_2]\sum_{i,j=0}^{p-1}\frac{\partial^{i+j}u}{\partial^{i}x\partial^{j}y}\bigg|_{(x,y)} 
    \frac{k_1^i k_2^j}{i!j!} \delta x^i \delta y^j + \mathcal{O}(|\delta x|^{p-1}+|\delta y|^{p-1}) \\
    &=
    \sum_{k_1,k_2=-\frac{p-1}{2}}^{\frac{p-1}{2}} \mathcal{K}[k_1,k_2]u(x+k_1\delta x, y+k_2\delta y) + \mathcal{O}(|\delta x|^{p-1}+|\delta y|^{p-1}) \\
    &=\mathcal{K} \circledast u + \mathcal{O}(|\delta x|^{p-1}+|\delta y|^{p-1})
    \end{aligned}
\end{equation} 
where $p$ is the size of the filter indexed by $k_1$ and $k_2$.
Letting the filter's entry $\mathcal{K}[k_1,k_2]$ be the corresponding Taylor series coefficient, we can see the error of approximation is bounded by $\mathcal{O}(|\delta x|^{p-1}+|\delta y|^{p-1})$. 

\textit{\textbf{Lemma 2:} Local truncation error of the forward computation (i.e., $    \widehat{u}^{(k+1)}=\widehat{u}^{(k)} + \widehat{\mathcal{F}}(\widehat{u}^{(k)};\boldsymbol{\theta})\delta t
$) diminishes as $\delta t$ decreases. }

\textit{\textbf{Proof}}: With the Taylor expansion of $u^{(k+1)}=u^{(k)} + \mathcal{F}(u^{(k)})\delta t+\mathcal{O}(\delta t^2)$, we can see the truncation error of the forward computation converges to zero as $\delta t$ decreases. 

{\color{black} \textit{\textbf{Theorem 1:} Suppose $\mathcal{F}:\mathbb{R}^{s} \rightarrow \mathbb{R}$ is a continuous real-valued function of multidimensional variables $\boldsymbol{\eta}\in\mathbb{R}^{s}$, where $\boldsymbol{\eta}$ denotes the set of system state $\mathbf{u}$ and its derivative terms, consisting of $s$ elements in total. For any small positive number $\epsilon$, there exist positive integers $M$ and $N$, real numbers $w_j$, $\gamma_{ij}$ and $b_j$ ($i=1,2,..., N$ and $j=1,2,..., M$), and variable set $\mathbf{E}\in\mathbb{R}^{N\times M}$, such that:}

\begin{equation}\label{eq:theorem1}
    \left| \mathcal{F}(\boldsymbol{\eta}) - \sum_{j=1}^M w_j \cdot \left[ \prod_{i=1}^N \Big( \gamma_{ij}E_{ij} + b_i \Big) \right]  \right| < \epsilon 
\end{equation} 

\textit{\textbf{Proof}}: Let us firstly denote the set of system state $\mathbf{u}$ and its derivative terms, consisting of $s$ elements in total, as $\boldsymbol{\eta} = [\mathbf{u}, \mathcal{L}_1(\mathbf{u}), \mathcal{L}_{2}(\mathbf{u}), ..., \mathcal{L}_{s-1}(\mathbf{u})]^\texttt{T}\in\mathbb{R}^{s}$. For example, $\boldsymbol{\eta} = [u, v, u_x, v_y, ...]^\texttt{T}$. The RHS of PDEs in Eq. \eref{eq:dynamic_system_pde} can then be represented by $\mathcal{F}(\boldsymbol{\eta})$. Based on the multivariate Taylor's theorem, for any small positive number $\epsilon$, there is a real-valued polynomial function $\mathcal{T}$ such that
\begin{equation}\label{eq:taylor}
    \left| \mathcal{F}(\boldsymbol{\eta}) - \mathcal{T}(\boldsymbol{\eta}) \right| < \epsilon
\end{equation} 
Here, $\mathcal{T}(\boldsymbol{\eta})$ can be expressed as:
\begin{equation}\label{eq:taylor1}
    \mathcal{T}(\boldsymbol{\eta}) = \sum_{n_1=0}^{n}\sum_{n_2=0}^{n}\cdots\sum_{n_s=0}^{n} \mathcal{N}(\boldsymbol{\eta})
\end{equation} 
where 
\begin{equation}\label{eq:taylor2}
    \mathcal{N}(\boldsymbol{\eta}) = c_{n_1} c_{n_2}\cdots c_{n_s}\big(\eta_1 - \bar{b}_1\big)^{n_1}\big(\eta_2 - \bar{b}_2\big)^{n_2}\cdots\big(\eta_s - \bar{b}_s\big)^{n_s}
\end{equation} 
Here, $c$'s denote the real-valued coefficients, $\bar{b}$'s the biases, and $n$ the maximum polynomial order. For simplicity, we omit the subscripts $\{n_1, ..., n_s\}$ in $\mathcal{N}(\boldsymbol{\eta})$.

\textit{\textbf{Lemma 3:} For real numbers $\eta$, $b$ and $c$, and integer $n'$, there exist real-valued vector $\boldsymbol{\alpha}\in\mathbb{R}^{n+1}$, real number $\tilde{b}$ and integer $n'\leq n$ such that $c(\eta-b)^{n'} = \prod_{i=0}^n\big(\alpha_i \eta - \tilde{b}\big)$ if $\|\boldsymbol{\alpha}\|_0=n'$.}

Based on Lemma 3, there are real numbers $\alpha$'s and $\tilde{b}$'s such that $\mathcal{N}(\boldsymbol{\eta})$ can be re-written as:
\begin{equation}\label{eq:rewrite}
    \mathcal{N}(\boldsymbol{\eta}) = \prod_{i=0}^n \left[\big(\alpha_{i1}\eta_1 - \tilde{b}_{1}\big)\big(\alpha_{i2}\eta_2 - \tilde{b}_{2}\big)\cdots\big(\alpha_{is}\eta_s - \tilde{b}_{s}\big) \right] ~~~\text{s.t.}~~~\|\boldsymbol{\alpha}_{k}\|_0 = n_k
\end{equation} 
where $\boldsymbol{\alpha}_{k}=[\alpha_{0k}, \alpha_{1k}, ..., \alpha_{nk}]^\texttt{T}\in\mathbb{R}^{n+1}$ is the $k$-th vector of $\alpha$'s ($k = 0, 1, ..., s$); $\|\cdot\|_0$ denotes the $\ell_0$ norm of a vector. By defining proper weights ($\boldsymbol{\beta}\in\mathbb{R}^{(n+1)s}$) and biases ($\hat{\mathbf{b}}\in\mathbb{R}^{(n+1)s}$), we can further express $\mathcal{N}(\boldsymbol{\eta})$ by:
\begin{equation}\label{eq:rewrite1}
    \mathcal{N}(\boldsymbol{\eta}) = \prod_{i=1}^{(n+1)s} \left[\beta_i\hat{\eta}_i + \hat{b}_i \right]
\end{equation} 
where $\hat{\boldsymbol{\eta}}:=\boldsymbol{\ell}\otimes\boldsymbol{\eta}\in\mathbb{R}^{(n+1)s}$ is the Kronecker transformation of $\boldsymbol{\eta}$; $\boldsymbol{\ell}\in\mathbb{R}^{n+1}$ is a column vector with all elements equal to 1; $\otimes$ denotes the Kronecker product. Note that $\boldsymbol{\beta}$ is sparse. Substituting Eq. \eref{eq:rewrite1} into Eq. \eref{eq:taylor1}, we obtain the equivalent formulation for $\mathcal{T}(\boldsymbol{\eta})$ as follows:
\begin{equation}\label{eq:taylor3}
    \mathcal{T}(\boldsymbol{\eta}) = \sum_{j=1}^M w_j \cdot \left[ \prod_{i=1}^N \Big( \gamma_{ij}E_{ij} + b_i \Big) \right]
\end{equation} 
where $\mathbf{E}:=\tilde{\boldsymbol{\ell}}\otimes\hat{\boldsymbol{\eta}}\in\mathbb{R}^{N\times M}$ is the Kronecker transformation of $\hat{\mathbf{E}}$; $\tilde{\boldsymbol{\ell}}\in\mathbb{R}^{M}$ is a row vector with all elements equal to 1; $\mathbf{w}\in\mathbb{R}^M$ and $\boldsymbol{\gamma}\in\mathbb{R}^{N\times M}$ denote some properly defined real-valued coefficients;  $\mathbf{b}\in\mathbb{R}^N$ is the bias vector. Note that the formulation of Eq. \eref{eq:taylor3} can be guaranteed when $M\geq (n+1)^s$ and $N\geq (n+1)s$. Substituting Eq. \eref{eq:taylor3} into Eq. \eref{eq:taylor} can thus prove Theorem 1. 

It is noted that the term $\gamma_{ij}E_{ij}$ in Eq. \eref{eq:taylor3} can be approximated by a series convolutional filters as shown in Lemma 1, inspiring the design of the universal polynomial approximator shown in Eq. \eref{eq:update_rule}. Although Theorem 1 still holds by selecting different values of $M$, $N$, $\mathbf{w}$ and $\mathbf{b}$, the approximation capability might be affected (with varying approximation errors). In particular, a small value of $n$ (thus $M$ and $N$) that represents less polynomial terms used for approximation will likely lead to a large truncation error. Nevertheless, since the proposed universal polynomial approximator is fully learnable, an equivalent model can be achieved by adapting the channel number (see \textcolor{blue}{Supplementary Note I}). If the underlying form of $\mathcal{F}$ is a polynomial type, Theorem 1 holds with much less parameters and terms required for satisfactory accuracy.

}

\subsection*{Loss Functions}

We employ different loss functions depending on the problem on hand. In the case of forward analysis of nonlinear systems, we assume the full knowledge on the system (e.g., governing equation, I/BCs) are available. The most straightforward approach to construct a predictive model would be utilizing the numerical discretization, i.e., customizing all the physics-based connections to realize the finite difference time updating. That is to say, the network architecture parameterizes an {\color{black}explicit} solver of PDE system which hence has a requirement on $\delta t$ for numerical stability. To avoid this issue, we construct the predictive model in an implicit manner. To be more concrete, we employ $\Pi$-block in the network as the approximator to $\mathcal{F}$ and compute the governing equation's residual from the spatiotemporal prediction using high-order finite difference stencils. The mean squared error (MSE) of the physics residual (following Eq. \eref{eq:dynamic_system_pde}) is employed as the loss function, which reads
\begin{equation} 
    \label{eq:loss_func_solve_pde} 
    \begin{aligned}
    \mathcal{J}(\mathbf{W}, \mathbf{b})=\textrm{MSE}\left ( \boldsymbol{\widehat{\mathcal{U}}}_t-\mathcal{F}( \boldsymbol{\widehat{\mathcal{U}}}) \right )
    \end{aligned}
\end{equation}
where $\boldsymbol{\widehat{\mathcal{U}}}\in \mathbb{R}^{n_t\times n\times H \times W}$ is the high-resolution prediction from the model, $\boldsymbol{\widehat{\mathcal{U}}}_t$ is the time derivative of $\boldsymbol{\widehat{\mathcal{U}}}$ computed through numerical discretization while $\mathcal{F}( \boldsymbol{\widehat{\mathcal{U}}})$ is the RHS of Eq. \eref{eq:dynamic_system_pde}), $(\mathbf{W}, \mathbf{b})$ denotes the trainable parameters of the network. With the gradient descent method for optimization, we can obtain a suitable set of $\Pi$-block parameters. This implicit way of establishing predictive model is more stable numerically regarding the selection of $\delta t$. We also need to note that the loss of initial/boundary conditions is not included in the loss function as they are already encoded through customized padding (see the subsection of \textcolor{blue}{Physics-encoded Spatiotemporal Learning}). 

In the problem of data-driven modeling, the goal is to reconstruct the most likely full-field solution $\boldsymbol{\widehat{\mathcal{U}}}$ given some low-resolution snapshots $\Tilde{\mathbf{u}}\in\mathbb{R}^{n_t'\times n\times  H'\times W'}$ where $n_t'<n_t$, $H'<H$ and $W'<W$. Therefore, the loss function to train the network is defined as
\begin{equation} 
    \label{sys_id_loss} 
    \displaystyle
    \mathcal{J}(\mathbf{W}, \mathbf{b})=\textrm{MSE}\left(\boldsymbol{\widehat{\mathcal{U}}}(\tilde{\mathbf{x}})-\tilde{\mathbf{u}}\right)+\lambda\cdot \textrm{MSE}\left(\boldsymbol{\widehat{\mathcal{U}}}^{(0)}-\mathcal{P}(\tilde{\mathbf{u}}^{(0)})\right)
\end{equation}
where $\boldsymbol{\widehat{\mathcal{U}}}(\tilde{\mathbf{x}})$ denotes the mapping of HR prediction $\boldsymbol{\widehat{\mathcal{U}}}\in \mathbb{R}^{n_t\times n\times H \times W}$ on the coarse grid $\mathbf{\Tilde{x}}$; $\tilde{\mathbf{u}}$ denotes the low-resolution measurement; $\mathcal{P}(\cdot)$ is a spatial interpolation function (e.g., bicubic or bilinear); $\lambda$ is the regularizer weighting. The regularization term denotes the IC discrepancy between the interpolated HR initial state $\mathcal{P}(\tilde{\mathbf{u}}^{(0)})$ and the predicted HR initial state $\widehat{\mathcal{U}}^{(0)}$ from ISG, which is found effective in preventing network overfitting. Compared with the existing work on physics-informed learning \cite{raissi2019physics, raissi2019deep, rao2020physicsElastic, gao2021phygeonet}, one major distinction of the loss function employed here is the absence of the physics loss. This {\color{black}is} because the prior physics knowledge is already encoded into the network architecture as shown in the subsection of \textcolor{blue}{Physics-encoded Spatiotemporal Learning}. This facilitates the learning process of the spatiotemporal system significantly. Eq. \eref{sys_id_loss} is also utilized as the loss function in the problem of system coefficient identification where the noisy and scarce measurement is available. However, a different network design is employed as elaborated in \textcolor{blue}{Supplementary Note Figure S.10}. {\color{black}In this case, multiple physics-based Conv layers are created to represent existing terms in the PDE (e.g., $\Delta u$, $uv^2$, $u$) while each layer (or term) is associated with a trainable variable to represent the corresponding coefficient.} By minimizing the the loss function with IC discrepancy regularizer, the unknown scalar coefficients in the system could be obtained.

\subsection*{\hsedit{Equation Discovery}}
\hsedit{The proposed PeRCNN-based PDE discovery model consists of three steps, including data reconstruction, sparse regression and coefficients fine tuning, discussed as follows.}

\paragraph{\textit{Data Reconstruction}.} Since the available measurement data collected in the real world is usually sparse and accompanied with noise, it is of common practice to pre-process the raw data so as to reconstruct the high-fidelity data (e.g., de-noised or high-resolution). In our proposed framework, the data reconstruction step (see Figure \ref{fig:diagram_eqn_discover}\textbf{a}) is first performed with the help of PeRCNN as a high-resolution data-driven predictive model. This step follows the same routine described in previous subsection of \textcolor{blue}{\text{Data-driven Modeling}}. Specifically, we establish a data-driven model from some low-resolution snapshots and then infer the high-resolution prediction (or solution) from the trained model. The reconstructed high-resolution data is then employed in the subsequent sparse regression to ensure the accuracy of the constructed library. The derivative terms in the library are estimated via finite-difference-based filtering on the reconstructed high-fidelity data.

\paragraph{\textit{Sparse Regression.}} With the reconstructed high-fidelity (i.e., HR and de-noised) solution, we are able to reliably estimate the library and thus accurately perform sparse regression for the explicit form \redit{or analytical structure} of PDEs. \hsedit{Note that} sparse regression is an extensively used technique for data-driven PDE discovery. It is rooted on a critical observation that the RHS of Eq. \eref{eq:dynamic_system_pde} for the majority of natural systems consists of only a few terms. To demonstrate how the sparse regression works, let us consider the measurement data with one single component, i.e., $u\in\mathbb{R}^{n_s\times n_t}$, which is defined on $n_s$ spatial locations and at $n_t$ time steps. After flattening the state variable into a column vector $\mathbf{U}\in\mathbb{R}^{n_s\cdot n_t\times 1}$, we are able to establish a library matrix $\mathbf{\Theta(U)}\in\mathbb{R}^{n_s\cdot n_t \times s}$ such that each of $s$ column vectors denotes a candidate function in $\mathcal{F}$ (e.g., linear, nonlinear, trigonometric, etc.). Accordingly, each row of $\mathbf{\Theta(U)}$ denotes a spatiotemporal location. If the column space of the library matrix is sufficiently rich, the governing PDE of the system can then be written as a linear system, namely,
\begin{equation} 
    \label{eq:kernel_space} 
    \mathbf{U}_t=\mathbf{\Theta(U)\Xi}
\end{equation}
where $\mathbf{U}_t$ is the vector of time derivative of $\mathbf{U}$; $\mathbf{\Theta(\cdot)}$ maps the original state variable space to a higher dimensional nonlinear space, e.g., $\mathbf{\Theta(U)}=[1,\mathbf{U},\mathbf{U}^2,\dots,\mathbf{U}_x,\mathbf{U}_y,\dots]$; $\mathbf{\Xi}\in\mathbb{R}^{s\times 1}$ is the sparse coefficient vector that represents the governing PDE. Sparse regression seeks to find a suitable $\mathbf{\Xi}$ such that the sparsity of the vector and the regression error are balanced. Specifically in our proposed framework, the Sequential Threshold Ridge regression (STRidge) algorithm \cite{rudy2017data} is adopted among other effective sparsity-promoting methods such as the Iterative Hard Thresholding (IHT) method \cite{haupt2006signal, blumensath2009iterative}, due to its superior performance compared with other sparsity-promoting algorithms, such as LASSO \cite{tibshirani1996regression} and Sequentially Thresholded Least Squares (STLS) \cite{brunton2016discovering}. For a given tolerance that filters the entries of $\mathbf{\Xi}$ with small value, we can obtain a sparse representation of $\mathcal{F}$ with the help of the STRidge algorithm, whose technical details are provided in \textcolor{blue}{Supplementary Note F.2}. Iterative search with STRidge can be performed to find the optimal tolerance according to the selection criteria given by:
\begin{equation} 
    \label{eq:STRidge} 
    \mathbf{\Xi}^*=\argmin_{\mathbf{\Xi}}\left \{ ||\mathbf{U}_t-\mathbf{\Theta(U)\Xi}||_2 + \gamma||\mathbf{\Xi}||_0 \right \}
\end{equation}
where $||\mathbf{\Xi}||_0$ is used to measure the sparsity of coefficient vector while regression error $||\mathbf{U}_t-\mathbf{\Theta(U)\Xi}||_2$ is used to measure model accuracy. Since the optimization objective has two components, we apply Pareto front analysis to select an appropriate weighting coefficient $\gamma$ (see \textcolor{blue}{Supplementary Note F.5.3}). As the accurate computation of the library matrix is critical to obtaining an accurate coefficient vector via sparse regression, \hsedit{the reconstructed} high-fidelity solution is subsequently used for computing the partial derivative terms involved in the library. 

As shown in Figure \ref{fig:diagram_eqn_discover}\textbf{b}, we collect the candidate set from the network for data reconstruction by performing symbolic computations on the $\Pi$-block and establish the library matrix $\mathbf{\Theta(U)}$. Note this is different from the traditional sparse regression in which the candidate set is predefined. A comparative study of these two ways of establishing the library matrix is performed in \textcolor{blue}{Supplementary Note F.5.1}. With the established linear system, sparse regression is performed afterwards to find a suitable coefficient vector $\mathbf{\Xi}$ that balances model complexity and accuracy. This is realized by solving the optimization problem described by Eq. \eref{eq:STRidge} with the help by the STRidge algorithm.  

\paragraph{\textit{Fine-tuning of Coefficients}.} Due to the high dimensionality of the reconstructed data, the sparse regression is performed on the subsampled linear system (e.g., randomly sampled 10\% rows, $8.2\times10^6$ rows in the 2D GS RD case) to avoid the very large number of rows which retains computational {\color{black}efficiency} without the loss of accuracy. To fully exploit the available measurement and further improve the accuracy of the discovered equations, we introduce the coefficient fine-tuning step to produce the final explicit PDE formula. The rest of training procedure is the same as that discussed in \textcolor{blue}{\text{Inverse Analysis of PDE Systems}}: all the original measurements are used to train a PDE structure preserved network (see Figure \ref{fig:diagram_eqn_discover}\textbf{c}) while the coefficient of each term is treated as a trainable variable. In \textcolor{blue}{Supplementary Note F.5.4}, we show that such a fine-tuning can considerably improve the accuracy of the discovered PDEs.

\section*{Data availability} 
All the used datasets in this study are available on GitHub at \url{https://github.com/isds-neu/PeRCNN} upon final publication.

\section*{Code availability} 
All the source codes to reproduce the results in this study are available on GitHub at \url{https://github.com/isds-neu/PeRCNN} upon final publication.

\bibliographystyle{unsrt}
\bibliography{references}

\vspace{20pt}
\noindent\textbf{Acknowledgement:}
\hsedit{The work is supported by the National Natural Science Foundation of China (No. 92270118 and No. 62276269), the Beijing Natural Science Foundation (No. 1232009), the National Key R\&D Program of China (No. 2021ZD0110400), and the Beijing Outstanding Young Scientist Program (No. BJJWZYJH012019100020098). We also acknowledge the support by the Huawei MindSpore platform. Y.L. and H.S. would like to acknowledge the support from the Fundamental Research Funds for the Central Universities. C.R. acknowledges the sponsorship of visiting research by H.S. at Renmin University of China.} \\

\noindent \textbf{Author contributions:} C.R., H.S. and Y.L. contributed to the ideation and design of the research; \hseditr{C.R., P.R. and Q.W.} performed the research; C.R., P.R., O.B., H.S. and Y.L. wrote the paper. \\

\noindent \textbf{Corresponding authors:} Hao Sun (\url{haosun@ruc.edu.cn}) and Yang Liu (\url{liuyang22@ucas.ac.cn}). \\

\noindent\textbf{Competing interests:}
The authors declare no competing interests.\\

\noindent\textbf{Supplementary information:}
The supplementary information is attached.

\clearpage
\setcounter{figure}{0}
\renewcommand{\figurename}{Extended Data Figure}

\setcounter{table}{0}
\renewcommand{\tablename}{Extended Data Table }

\begin{table}[h!]
\caption{Computational parameters for datasets generation. $\delta x$ denotes the spacing of the grid while $\delta t$ denotes the time spacing. \hsedit{A detailed discussion on how to properly select $\delta x$ and $\delta t$ is given in \textcolor{blue}{Supplementary Note H}.} }
\vspace{-15pt}
\label{tb:data_set}
\begin{center}
\begin{small}
\begin{tabular}{lccccc}
\toprule 
Dataset & Equation & Parameters &  Dimensions & $\delta t$ & $\delta x$ \\
\midrule
2D $\lambda$-$\Omega$& Eq. \eref{eq:LO_eqn} &  $\mu_u$=0.1, $\mu_v$=0.1 $\beta$=1.0  & $101^2\times 801$ & 0.0125 & 0.2 \\
2D FN & Eq. \eref{eq:fn_eqn} & $\mu_u=1.0$, $\mu_v=10.0$, $\alpha=0.01$, $\beta=0.25$  & $101^2\times 6001$ & 0.002 & 1.0 \\
3D FN & Eq. \eref{eq:fn_eqn} & $\mu_u=1.0$, $\mu_v=10.0$, $\alpha=0.01$, $\beta=0.25$  & $51^3\times 1001$ & 0.004 & 1.0 \\
2D GS & Eq. \eref{eq:gs_eqn} & $\mu_u=2.0e-5$, $\mu_v=5.0e-6$, $F=0.04$, $\kappa=0.06$  & $101^2\times 2501$ & 0.5& $0.01$ \\
3D GS & Eq. \eref{eq:gs_eqn} &  $\mu_u=0.2$, $\mu_v=0.1$, $F=0.025$, $\kappa=0.055$  & $49^3\times 1501$  & 0.5 & 25/12\\
\bottomrule
\end{tabular}
\end{small}
\end{center}
\end{table}

\clearpage

\begin{table*}[h!]
\centering
\caption{Summary of the coefficient identification results for 2D Gray-Scott reaction-diffusion system. The training dataset (or measurement) includes 26 snapshots with resolution of $26\times 26$ for S1 while 2 snapshots with resolution of $51\times 51$ for S2. }
\resizebox{\textwidth}{!}{%
\begin{tabular}{lcccccccc}

\toprule
Scenario & Noise (\%) & $\mu_u (10^{-5})$ & $\mu_v (10^{-6})$ & $c_1$ & $c_2$ & $c_F (10^{-2})$ & $c_\kappa (10^{-2})$ & MARE (\%)   \\
\midrule
Ground truth & -& $2.0$ & $5.0$ & 1.0 & 1.0 & $4.0$ & $6.0$ & - \\
\cmidrule{1-9}
S1 & 0  & $1.987$ & $4.989$ & $0.9920$ & $0.9941$ & $3.970$ & $5.965$ & $0.60$ \\
   & 10 & $1.950$ & $5.010$ & $0.9724$ & $0.9823$ & $3.938$ & $5.941$ & $1.61$ \\
\midrule
S2 & 0  & $1.981$ & $5.124$ & $0.9886$ & $0.9993$ & $4.014$ & $6.046$ & $0.96$ \\
   & 10 & $1.964$ & $5.111$ & $0.9864$ & $0.9987$ & $4.003$ & $6.044$ & $1.05$ \\
\bottomrule
\end{tabular}}
\label{tb:table_coef_id} 
\end{table*}

\clearpage

\begin{table}[h!]
    \centering
    \caption{Discovered PDEs from the measurement data under various noise levels compared with the ground truth. }
    \label{tb:result_pde_discovery}
    \begin{small}
    \begin{tabular}{lll}
    \toprule 
    Example & Noise & Discovered PDE \\
    \midrule
    \multirow{8}{*}{2D $\lambda$--$\Omega$ RD} & 0\%  &  
        $\begin{aligned}
        u_t&=0.096\Delta u + 1.013u - 1.019u^3 + 1.001u^2v -1.021uv^2 + 0.9977v^3 \\
        v_t&=0.096\Delta v + 1.006v - 0.998u^3 -1.0139u^2v -1.002uv^2 - 1.012v^3
        \end{aligned}$\\
        \cmidrule(lr){2-3}
        & 5\%  &  
        $\begin{aligned}
        u_t=& 0.096 \Delta u + 1.038u - 1.048u^3 + 1.004u^2v - 1.050uv^2 + 0.998v^3 \\
        v_t=& 0.100 \Delta v + 1.014v - 0.998u^3 - 1.025u^2v - 0.999uv^2 - 1.015v^3
        \end{aligned}$\\
        \cmidrule(lr){2-3}
        & 10\% &  
        $\begin{aligned}
        u_t&=0.101\Delta u + 1.079u - 1.090u^3 + 1.008u^2v -1.090uv^2 + 0.9982v^3\\
        v_t&=0.105\Delta v + 1.033v - 0.965u^3 - 1.046u^2v - 0.967uv^2 - 1.029v^3 + {\underline{0.029u}}
        \end{aligned}$\\
        \cmidrule(lr){2-3}
        & Truth &      
        $\begin{aligned}
        u_t=&0.1\Delta u + (1-u^2-v^2)u+ (u^2+v^2)v \\
        v_t=&0.1\Delta v - (u^2+v^2)u + (1-u^2-v^2)v
        \end{aligned}$\\
    \midrule
    \multirow{8}{*}{2D GS RD} & 0\%  &  
        $\begin{aligned}
        u_t=&1.999\times10^{-5} \Delta u - 0.992uv^2 - 0.04003u + 0.03999 \\
        v_t=&5.008\times10^{-6} \Delta v + 1.021uv^2 - 0.1001v
        \end{aligned}$\\
        \cmidrule(lr){2-3}
        & 5\%  &  
        $\begin{aligned}
        u_t=&2.001\times10^{-5} \Delta u - 1.003uv^2 - 0.04008u + 0.04008 \\
        v_t=&5.042\times10^{-6} \Delta v + 1.009uv^2 - 0.1007v
        \end{aligned}$\\
        \cmidrule(lr){2-3}
        & 10\% &   
        $\begin{aligned}
        u_t=&1.846\times10^{-5} \Delta u - 0.904uv^2 - {\underline{0.0863u^3}} + 0.04019 \\
        v_t=&5.438\times10^{-6} \Delta v + 1.051uv^2 - 0.1174v
        \end{aligned}$\\
        \cmidrule(lr){2-3}
        & Truth &  
        $\begin{aligned}
        u_t=&2.0\times10^{-5} \Delta u - 1.0uv^2 + 0.04(1-u) \\
        v_t=&5.0\times10^{-6} \Delta v + 1.0uv^2 - 0.1v
        \end{aligned}$\\
    \bottomrule
    \end{tabular}
    \end{small}
\end{table}

\end{document}


\title{\textbf{Supplementary Information} for: \\ Encoding physics to learn reaction-diffusion processes}

\author[1,2,$\dag$]{Chengping Rao}
\author[3,$\dag$]{Pu Ren}
\author[1]{Qi Wang}
\author[4]{Oral Buyukozturk}
\author[1,5,*]{Hao Sun}
\author[6,$\ddag$]{Yang Liu}

\affil[1]{\small Gaoling School of Artificial Intelligence, Renmin University of China, Beijing, 100872, China}
\affil[2]{\small Department of Mechanical and Industrial Engineering, Northeastern University, Boston, MA 02115, USA}
\affil[3]{Department of Civil and Environmental Engineering, Northeastern University, Boston, MA 02115, USA}
\affil[4]{Department of Civil and Environmental Engineering, MIT, Cambridge, MA 02139, USA} 
\affil[5]{Beijing Key Laboratory of Big Data Management and Analysis Methods, Beijing, 100872, China} 
\affil[6]{\small School of Engineering Sciences, University of Chinese Academy of Sciences, Beijing, 101408, China \vspace{12pt}}
\affil[$\dag$]{Equally contributed\vspace{12pt}}
\affil[*]{Corresponding author. E-mail: haosun@ruc.edu.cn}
\affil[$\ddag$]{Corresponding author. E-mail: liuyang22@ucas.ac.cn\vspace{12pt}}

\date{}

\maketitle

\tableofcontents

\vspace{24pt}

\noindent This supplementary document provides a detailed description of the proposed method, algorithm, examples, and discussion of technical challenges for for modeling and discovery of reaction-diffusion systems.

\section{Background}

\subsection{Methodology for modeling scientific problems}

\begin{figure}[t!]
	\centering
	\includegraphics[width=0.6\textwidth]{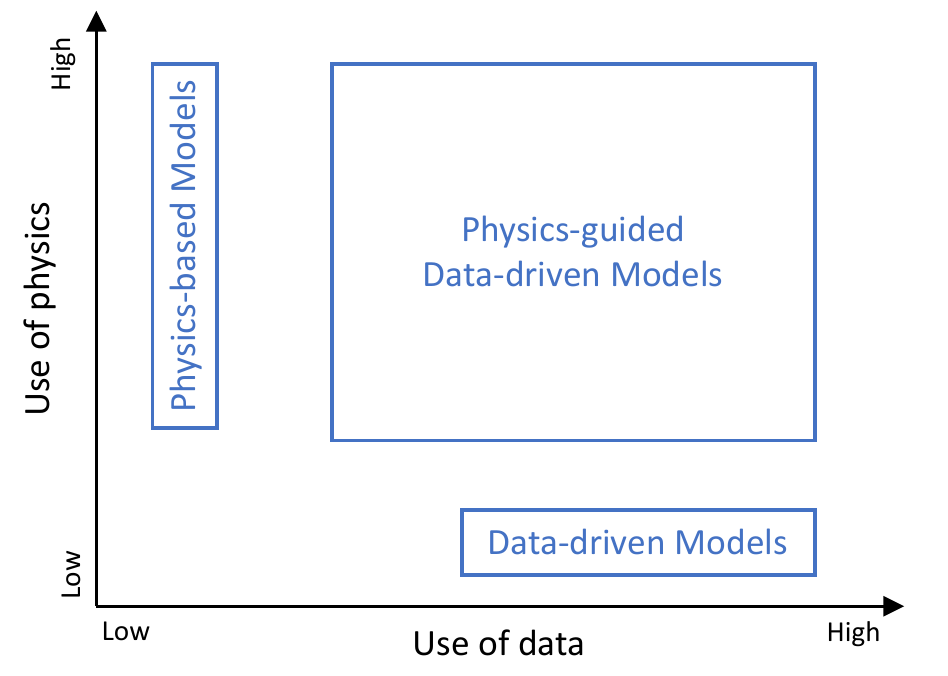}
	\caption{Categories of the computational methods for physics modeling. $x$-axis denotes the amount of measurement data used and $y$-axis denotes the amount of physics (or domain) knowledge used. }
	\label{c1:fig:Phy_guided_model}
\end{figure}

Since the dawn of modern science, people have been seeking various mathematical models to explain the physical phenomena in nature, e.g., orbits of planets and turbulence on rivers. After centuries of development, the existing methods for modeling scientific problems can be primarily categorized into three classes, the physics-based model, the data-driven model and the hybrid physics-guided data-driven model, as shown by Fig. \ref{c1:fig:Phy_guided_model}. The criteria for identifying each category is based on the usage of data and physics knowledge. The physics-based models, as the bedrock of computational physics, have been extensively studied in the past decades. One common characteristic shared by physics-based models is that they are usually derived by rigorously following the some \textit{physical principles}, such as the conservation laws, geometric brownian motion assumptions and knowledge-based deductions. Some representative physics-based models include the finite element method, finite difference method and finite volume method. The ever-growing data, thanks to the rapid advancement of sensing technologies, contributes to the development of data-driven models. In general, the establishment of data-driven model requires little physics knowledge but \textit{a wealth amount of measurement data} (or observations) for generalizing the hidden (or unknown) physics. Furthermore, the advancements of deep learning (DL) models in recent years also provide solid foundation for developing the data-driven model. People utilize experimental or numerical data to establish data-driven predictive models \cite{WANG2018337, wang2019meta, HAN2019112603, rao2020three, vlassis2020geometric}. However, the above physics-based and data-driven models reflect two extreme scenarios of scientific modeling, in which either the complete physics of a problem or a large amount of data is available. More in the real world are the cases that partial physics knowledge and limited amount of data are available, in which the physics-guided data-driven model \cite{raissi2018deep, raissi2019physics, raissi2020hidden, rao2020physicsElastic, geneva2020modeling, wang2020towards, gao2021phygeonet} comes to help. 

The physics-guided data-driven model refers to the computational method that integrates the available measurement data with the partial physics knowledge for predicting scientific problems. As the name implies, physics-guided data-driven model primarily focuses on ``teaching'' the data-driven models about fundamental physics or domain knowledge so that the resultant model can produce physically consistent result. One primary benefit bought by the guiding physics is to reduce the need of big data. Furthermore, guided by the prior physical knowledge, the resultant data-driven model could become more interpretable to human and less vulnerable to imperfect training data (e.g., data is missing or contains outliers). Despite the concept of physics-guided data-driven model has gained momentum just in the last few years, it has been demonstrated to be applicable and effective in diverse disciplines, such as turbulence modeling \cite{mohan2018deep, xiao2019reduced, bode2021using, mohan2020embedding, wang2020towards}, climate science \cite{faghmous2014big, karpatne2017physics, kashinath2021physics}, material discovery \cite{raccuglia2016machine, liu2021physics}, cardiovascular flows \cite{sun2020physics, kissas2020machine, sahli2020physics}, geophysics \cite{karimpouli2020physics, song2021solving, rasht2021physics}, and so on. A good review on this topic could be found in \cite{karpatne2017theory, willard2020integrating, karniadakis2021physics}.

\subsection{Related work}

\paragraph{ResNet:} 
Deep residual network (ResNet) has been one of the most successful DL architectures, which addresses the notorious problem of gradient vanishing/exploding for very deep networks \cite{he2016deep}. Studies reveal its close connection with the forward Euler time-stepping scheme \cite{chen2015learning, chang2017multi, chen2018neural, ruthotto2019deep}, which provides a new perspective for the architecture design of DL models \cite{lu2018beyond}. That is to consider a DL task in the view of dynamical system and leverage our rich knowledge of numerical methods for network design, e.g., the backward Euler scheme in PolyNet \cite{zhang2017polynet} and the Runge-Kutta scheme in FractalNet \cite{larsson2016fractalnet}. Recurrent ResNets \cite{liao2016bridging, zhang2017deep}, whose parameters are shared across time, are also developed to solve sequential prediction problems. Although ResNet has shown success in lots of applications, residual blocks composed of traditional convolutional or fully connected layers face the issue of poor interpretability, hence hindering its applications to spatiotemporal dynamical systems where governing PDEs are potentially available.

\paragraph{Physics-informed DL:} 
Recently, there have been attempts to leverage physics principles to inform deep neural network (DNN) training. \citep{raissi2018deep} proposed a deep hidden physics model (DHPM) to enable data-driven modeling of spatiotemporal dynamics, where two DNNs are designed to predict the dynamics meanwhile approximating the underlying PDEs. Another representative method is called physics-informed neural network (PINN) for solving forward and inverse PDE problems \cite{raissi2019physics, raissi2019deep, raissi2020hidden}. The network is informed by physics through a weakly imposed penalty loss consisting of residuals of PDEs and initial/boundary conditions (I/BCs). One major limitation of PINN is that its accuracy relies largely on the these soft physical constraints \cite{wang2020understanding, rao2020physicsElastic} which may not be satisfied well during training given a poor selection of hyperparameters. Furthermore, the use of fully connected layers poses intrinsic limitations to low-dimensional parameterizations. Efforts have been placed to overcome these issues by employing discrete learning schemes via convolutional filters, such as HybridNet \cite{long2018hybridnet}, dense convolutional encoder-decoder network \cite{zhu2019physics}, auto-regressive encoder-decoder model \cite{geneva2020modeling}, TF-Net \cite{wang2020towards} and PhyGeoNet \cite{gao2021phygeonet}. These methods generally show better computational efficiency and accuracy. However, the core learning component of these networks is still a black box and the resulting models lack the capability to ``hard-encode'' our prior physical knowledge.

\paragraph{PDE-Net:} 
\citep{long2018pde,long2019pde} proposed two versions of PDE-Net for predicting dynamics of complex systems and uncovering the hidden PDE structures based on measurement data, where the differential operators are approximated by convolutional filters while the nonlinear PDE functional is learned from neural networks (e.g., symbolic neural networks). The interpretability lies in the constraints on (1) the learnable convolutional filters which pose a well-defined multivariate regression problem and (2) the predefined candidate terms potentially appears in the underlying PDEs. 

One major distinction of our work with PDE-Net is the usage of multiplicative form of networks. The $\delta t$-block in PDE-Net \citep{long2018pde} approximates the response function of $\mathcal{F}$ as
\begin{equation} 
    \label{eq:pde-net} 
    \widehat{\mathcal{F}}(u)=\sum_{0\le i+j\le N} f_{ij}(x,y) \cdot \left ( \mathcal{K}_{ij} \circledast u \right ) 
\end{equation}
where $u$ denotes the state variable defined on a Cartesian grid (a.k.a. snapshot of the system), $\circledast$ denotes the convolutional operation, $N$ is the highest derivative order known $a~prior$ and partial differential operators are approximated by Conv filters $\mathcal{K}_{ij}$, i.e., $\mathcal{K}_{ij} \circledast u \approx \frac{\partial^{i+j}u}{\partial x^i \partial y^j}$. We can assume the $\delta t$-block consists of one Conv layer with multiple channels, each of which associated with a filter $\mathcal{K}_{ij}$, followed by a $1\times1$ Conv layer to linearly combine the output of each Conv layer with the coordinate-dependent weight $f_{ij}(x,y)$. Our network remedies the issue that PDE-Net lacks the capability to represent power functions of $u$ or its derivatives, e.g., $u\Delta u$ and $u^2v$, by introducing the elementwise product operation on the output of Conv layers. In contrast to Eq. \eref{eq:pde-net} by the $\delta t$-block in PDE-Net, the $\Pi$-block can be expressed as 
\begin{equation} 
    \label{eq:pi-block} 
    \widehat{\mathcal{F}}(u)=\sum_{c=1}^{N_c} W_c \cdot \left [ \prod_{l=1}^{N_l} \left ( \mathcal{K}_{c,l} \circledast u + b_l \right ) \right ]
\end{equation}
where $N_l$ denotes the number of parallel Conv layers; $N_c$ denotes the number of channels of each Conv layer and $\mathcal{K}_{c,l}$ is the weight of Conv filter associated with $l^\text{th}$ layer and $c^\text{th}$ channel while $b_l$ is the bias of $l^\text{th}$ layer; $W_c$ is the weight corresponding to $c^\text{th}$ channel in $1\times 1$ Conv layer while the bias is omitted for simplicity. The weights and biases of the Conv filters $\mathcal{K}_{c,l}$ are made trainable to render the network more flexibility. Note that the dependent variables $(x,y)$ in the weight $W_c$ is omitted since we mainly study isotropic dynamical systems in this paper.


\paragraph{Product unit neural network (PUNN):} 
Product unit \cite{durbin1989product} is a type of computational unit for feedforward NN proposed in the 90's. Different from the weighted sum unit in conventional feedforward neural network (NN), the product unit calculates a weighted product, where each input is raised to a power determined by the trainable weight. The intuition of this unit is that it can learn to represent any polynomial functions regarding the input. Although the product unit is proven efficient for some tasks involving higher order representations, it suffers from the increased number of local minima, deep ravines and valleys \cite{ghosh1992efficient, ismail2000global}. Our work explores this idea of product unit on the convolutional neural network (CNN) which is rarely studied so far.

\section{Methodology}

\subsection{Residual learning for dynamical systems}\label{sec:residual_pde}

Let us consider a spatiotemporal dynamical system governed by a set of nonlinear, coupled PDEs as
\begin{equation} 
    \label{eq:dynamic_system_pde} 
    \mathbf{u}_t=\mathcal{F}\left(\mathbf{x}, t, \mathbf{u}, \mathbf{u}^2, \nabla_\mathbf{x}  \mathbf{u}, \mathbf{u}\cdot\nabla_\mathbf{x} \mathbf{u}, \nabla^2 \mathbf{u}, \cdots\right)
\end{equation}
where the state variable/solution $\mathbf{u}(\mathbf{x},t)\in\mathbb{R}^n$ (e.g., $\mathbf{u}=[u,~v]^\texttt{T}$ for $n=2$) is defined over the spatiotemporal temporal domain $\{(\mathbf{x}, t)\} \in \Omega\times\mathcal{T}$ where $\Omega$ denotes the spatial domain while $\mathcal{T}$ denotes the temporal domain; $\nabla_{\mathbf{x}}$ is the Nabla operator with respect to spatial coordinate $\mathbf{x}$; and $\mathcal{F}(\cdot)$ is a nonlinear function that parameterizes the dynamics of the system. The solution to this problem is subject to the initial condition (IC)  $\mathcal{I}(\mathbf{u};t=0,\mathbf{x} \in \Omega)=0$ and boundary condition (BC) $\mathcal{B}(\mathbf{u},\nabla_{\mathbf{x}}\mathbf{u}, \cdots;\mathbf{x} \in \partial \Omega)=0$, where $\partial \Omega$ denotes the boundary of the system domain. In this paper, we will mainly focus on regular (e.g., rectangular or cubic) physical domains. That said, the state variable $\mathbf{u}$ can be discretized on a Cartesian grid at time steps $\{t_1,..., t_k, ..., t_{n_t}\}$, where $n_t$ denotes the total number of time steps. The design principle of the proposed physics-encoded recurrent-convolutional neural network (PeRCNN) roots on the fact that the state variable $\mathbf{u}$ can be updated iteratively through numerical differentiation, e.g., the forward Euler scheme or Runge-Kutta scheme. Specifically, we propose a recurrent network based on the forward Euler scheme as follows


\begin{equation} 
    \label{eq:residual_rcnn} 
    \begin{aligned}
    \widehat{\mathbf{u}}^{(k+1)}=\widehat{\mathbf{u}}^{(k)} + \widehat{\mathcal{F}}(\widehat{\mathbf{u}}^{(k)};\boldsymbol{\theta})\delta t
    \end{aligned}
\end{equation}
where $\delta t$ is the time spacing, $\widehat{\mathbf{u}}^{(k)}$ is the solution at time $t_k$ and $\widehat{\mathcal{F}}$ is an approximate function parameterized by $\boldsymbol{\theta}$ that ensembles a series of operations for computing the right hand side (RHS) of Eq. \eref{eq:dynamic_system_pde}. The form of Eq. \eref{eq:residual_rcnn} makes it possible to design a recurrent network for spatiotemporal prediction of the solution. In fact, ResNet \cite{he2016deep}, one of the most successful DL architecture, has been shown in many studies \cite{chen2015learning, chang2017multi, chen2018neural, ruthotto2019deep} to resemble the forward Euler time-stepping scheme. Inspired by a rich family of numerical differentiation techniques, people also invented various network architectures, e.g., FractalNet \cite{larsson2016fractalnet} based on Runge-Kutta scheme, PolyNet \cite{zhang2017polynet} based on backward Euler scheme, etc. 

The goal of this paper is to develop a flexible recurrent network architecture that could encode the prior physics knowledge into the architecture design for solving some representative tasks in scientific modeling, including (1) to solve the partial differential equation (PDE) of a dynamical system when the initial and boundary value problem (IBVP) is well defined; (2) to establish a high-fidelity (i.e., high-resolution and denoised) predictive data-driven model for a system given scarce and noisy measurement data; (3) to discover the explicit form or some scalar coefficients of the governing PDE of a dynamical system from measurement data.

\subsection{Design philosophy of the network architecture}
\label{sec:architecture}


In this part, we elaborate the design philosophy of the network architecture of physics-encoded recurrent-convolutional neural network (PeRCNN), which is shown in \textcolor{blue}{Main Text Fig. 1\textbf{a}}. As the name implies, PeRCNN is a recurrent and convolutional network. It consists of two major components: a fully convolutional (Conv) decoder as initial state generator (ISG) and a novel recurrent block, namely $\Pi$-block (product), for updating the state variables recursively as depicted by \textcolor{blue}{Main Text Fig. 1\textbf{a}}. Since we assume the collected measurement data is in low resolution, ISG is introduced to produce a full-resolution (or high-resolution) initial state $\widehat{\boldsymbol{\mathcal{U}}}^{(0)}\in \mathbb{R}^{n\times H \times W}$ (where $n$ is the number of components and $H \times W$ is the resolution of the grid) so that the recurrent computation can start. In the rest of this paper, we use $\Tilde{\mathbf{u}}\in\mathbb{R}^{n_t'\times n\times  H'\times W'}$ to denote the available measurement data, i.e., $n_t'$ LR snapshots of the system. In contrast, $\boldsymbol{\widehat{\mathcal{U}}}\in \mathbb{R}^{n_t\times n\times H \times W}$ is used to represent the full-resolution prediction by PeCRNN where $n_t$ is the number of prediction steps and $n_t'<n_t$, $H'<H$ and $W'<W$. 

In the recurrent $\Pi$-block (see zoom-in view of \textcolor{blue}{Main Text Fig. 1\textbf{a}}), which is the core of PeRCNN, the state variable $\widehat{\boldsymbol{\mathcal{U}}}^{(k)}$ from the previous time step first goes through multiple parallel Conv layers. The feature maps produced by these layers are then fused via the elementwise product operation. As a result of the elementwise product operation, the output of each Conv layer should has the same size. A Conv layer with the filter size of 1 \cite{lin2013network} is appended following the elementwise product operation to aggregate multi-channel output into the output of desired number of channels. We seek to train the network such that the output of the $1\times1$ Conv layer approximates the nonlinear function $\mathcal{F}(\cdot)$. Multiplying the approximated $\widehat{\mathcal{F}}\left (\boldsymbol{\widehat{\mathcal{U}}}^{(k)} \right)$ by the time spacing $\delta t$ yields the residual of the state variable at time $t_k$, i.e., $\delta \boldsymbol{\widehat{\mathcal{U}}}^{(k)}$, which would be used to update the state variable. In a nutshell, $\Pi$-block can be written equivalently as 

\begin{equation} 
    \label{eq:update_rule} 
    \begin{aligned}
    \widehat{\mathcal{F}}\left (\boldsymbol{\widehat{\mathcal{U}}}^{(k)}\right)&=\sum_{c=1}^{N_c} W_{c} \cdot \left [ \prod_{l=1}^{N_l} \left (\mathcal{K}_{c,l} \circledast \boldsymbol{\widehat{\mathcal{U}}}^{(k)} + b_l \right ) \right ] \\
    \boldsymbol{\widehat{\mathcal{U}}}^{(k+1)} &= \boldsymbol{\widehat{\mathcal{U}}}^{(k)} + \widehat{\mathcal{F}}\left (\boldsymbol{\widehat{\mathcal{U}}}^{(k)} \right )\cdot \delta t
    \end{aligned}
\end{equation}
where $N_c$ and $N_l$ are the number of channels and parallel Conv layers respectively; $\circledast$ denotes the Conv operation; $\mathcal{K}_{c,l}$ denotes the weight of Conv filter of $l^\text{th}$ layer and $c^\text{th}$ channel while $b_{l}$ denotes the bias of $l^\text{th}$ layer; $W_c$ is the weight of $c^\text{th}$ channel in $1\times 1$ Conv layer with the bias omitted for the sake of simplicity. It should be noted that a highway physics-based Conv layer could be created when some specific terms are known \textit{a priori} in the PDE, which would be detailed in Section \ref{sec:encoding_mechanism}. 


One salient characteristic of this network architecture is to employ the parallel Conv layers whose output is multiplied together elementwisely. This architecture bears a resemblance to the famous Inception network \cite{szegedy2015going} which has a similar block of parallel Conv layers. However, feature maps produced by the parallel Conv layers in the Inception network are fused through the traditional way of concatenation instead of the elementwise product operation. With such a way of achieving nonlinearity, the network is free from using any activation functions, e.g., Sigmoid or ReLU function. The benefits brought by this setting to achieve the nonlinearity are three-fold:
\vspace{-6pt}
\begin{itemize}
\item Though the nonlinear activation function is crucial to the expressiveness of the DL model, it is also a source of the poor interpretability. Since we seek to build a recurrent block that generalizes the unknown physics, these activation functions that appear in the traditional network are not adopted in our network design. 
\vspace{-3pt}
\item The nonlinear function $\mathcal{F}$ in the form of polynomial\footnote[1]{The polynomial herein encompasses spatial derivatives, e.g., $\mathbf{u}\cdot\nabla u$ and $u^2v$ are 2$^\text{nd}$ and 3$^\text{rd}$ degree respectively.} covers a wide range of well-known dynamical systems, such as Navier-Stokes, reaction-diffusion (RD), Schr$\ddot{\text{o}}$dinger equations, to name only a few. As it is proven the spatial derivatives can be computed by Conv filters \cite{cai2012image}, a $\Pi$-block with $n$ parallel Conv layers of appropriate filter size has the expressiveness to represent a polynomial up to the $n^\text{th}$ degree. In Section \textcolor{blue}{Methods} of the \textcolor{blue}{Main Text}, we prove that $\Pi$-block is a universal polynomial approximator. 
\vspace{-3pt}
\item Compared with the regression models that largely rely on predefined basis functions or prior knowledge (e.g., the highest order) of $\mathcal{F}$ \cite{brunton2016discovering, long2018pde}, our $\Pi$-block is flexible at representing a family of nonlinear function $\mathcal{F}$. For example, a $\Pi$-block with 2 parallel layers of appropriate filter size ensembles a family of polynomials up to the 2$^\text{nd}$ degree (e.g., $u$, $\Delta u$, $uv$, $\mathbf{u}\cdot\nabla u$), with no need to explicitly define the basis.
\end{itemize}
\vspace{-6pt}

Noteworthy, we above consider the nonlinear function $\mathcal{F}$ in the form of polynomial primarily. Terms of other types, such as trigonometric and exponential functions, are not considered in this work for simplicity. However, incorporating them would require no more effort than adding a particular symbolic activation (e.g., sin, cos, exp, etc.) layer following the Conv operation. More importantly, these functions can be approximated by polynomials based on Taylor series as argued in \cite{brunton2016discovering}. {\color{black}To verify the effectiveness of $\Pi$-block on more complex function, we perform tests on a two-dimensional reaction-diffusion equation whose reaction term is a non-polynomial component. Readers can visit Section \ref{sec:non_poly_term} for more details.}



\subsection{Encode prior physics into the network architecture}
\label{sec:encoding_mechanism}

\begin{figure}[t!]
    \centering
	\includegraphics[width=0.8\linewidth]{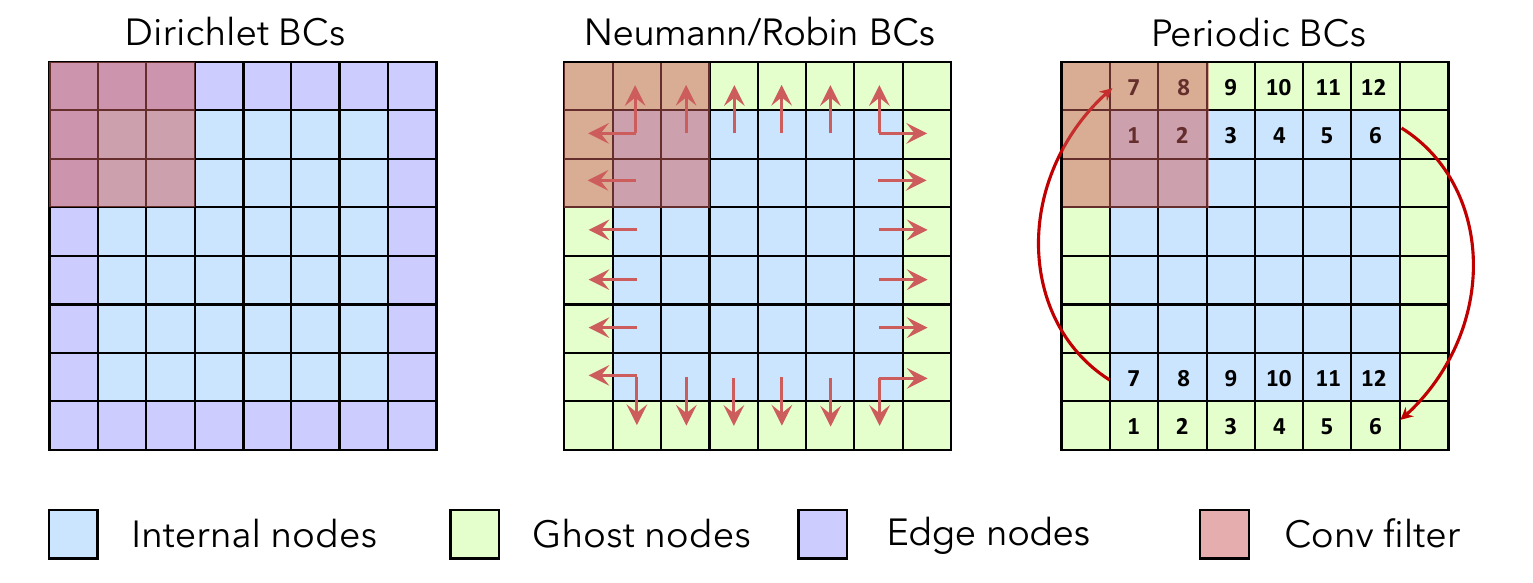} 
	\caption{\hsedit{Diagram of boundary condition (BC) padding.} } 
	\label{fig:periodic_padding}
\end{figure}

The encoding mechanism is employed in the network to strictly impose the prior physics knowledge of the system, which contributes to a well-posed optimization problem. In this work, two types of physics can be considered for encoding, namely, the prior knowledge on I/BCs and known terms in the governing PDE. The ICs (or initial states) can be naturally imposed when PeRCNN starts the recurrent computation from $\tilde{\mathbf{u}}^{(0)}$. For the BCs (Dirichlet or Neumann type), we borrow the idea from the finite difference (FD) method and apply the physics-based padding to the model's prediction at each time step (i.e., $\widehat{\boldsymbol{\mathcal{U}}}^{(k)}$). Specifically, we pad the prediction with prescribed values defined by the Dirichlet BCs. The padded value of the Neumann BCs is computed based on the boundary value and their gradient information. In this work, we mainly consider the periodic Dirichlet BCs for dataset generation and network training, which have been widely adopted in related works \cite{long2018pde, raissi2018deep, raissi2019physics}. We introduce the periodic padding, as shown in Fig. \ref{fig:periodic_padding}, that utilizes the boundary value from the other side of the grid, for both the prediction and the feature maps. The strength of periodic padding over traditional zero-padding is that it avoids the loss of the nodal information on the physical boundaries. \hsedit{While we consider the periodic boundary condition throughout this paper, our approach can be extended to various types of BCs through the convolutional padding that mimics the finite difference method. Detailed formulations of various boundary conditions in our approach are given in Table \ref{tb:BC_padding}. A numerical example with Neumann BCs is also provided in Section \ref{sec:neumann_bound} to demonstrate the effectiveness of our approach in handling various BCs. }

\begin{sidewaystable}[htbp]\color{black}
\centering
\caption{\hsedit{Summary of the convolutional padding for various BCs.}}
\small
\makebox[\textwidth][c]{
\begin{tabular}{llllll}
\toprule
Name & Continuous & Discrete & Ghost & Padding & Padding \\
 & Form & Form & Nodes & Location (Nodes) & Value  \\
\midrule
Dirichlet & $u(\mathbf{x}) = \bar{u}(\mathbf{x}), \mathbf{x}\in\Gamma_d$ & $u_{pj} = \bar{u}_{j}$ & No & Edge & $u_{pj} = \bar{u}_{j}$ \\
\midrule
Neumann & $\displaystyle \frac{\partial u(\mathbf{x})}{\partial \mathbf{n}} = f(\mathbf{x}), \mathbf{x}\in\Gamma_n$ & $\displaystyle \frac{u_{(p+1)j}-u_{(p-1)j}}{2\delta_x} = f_{j}$ & Yes & Ghost & $u_{(p+1)j} = u_{(p-1)j} + 2{\delta_x}f_{j}$ \\
\midrule
Robin & $\displaystyle \alpha u(\mathbf{x}) + \beta\frac{\partial u(\mathbf{x})}{\partial \mathbf{n}} = g(\mathbf{x}), \mathbf{x}\in\Gamma_r$ & $\displaystyle \alpha u_{pj} + \beta\frac{u_{(p+1)j}-u_{(p-1)j}}{2\delta_x} = g_{j}$ & Yes & Ghost & $\displaystyle u_{(p+1)j} = \frac{2\delta x}{\beta}\big(g_{j} -\alpha u_{pj} \big) + u_{(p-1)j}$ \\
\midrule
Periodic & $u(\mathbf{x}_1) = u(\mathbf{x}_2), \mathbf{x}_1\in\Gamma_{1}, \mathbf{x}_2\in\Gamma_{2}$ & $u_{pj} = u_{1j}$ & Yes & Edge/Ghost & $u_{pj}=u_{1j}$ and $u_{(p+1)j} = u_{2j}$ \\
\bottomrule
\end{tabular}}
\label{tb:BC_padding} 
\begin{tablenotes}
\item \hsedit{\small Note: We take the right boundary as illustration (e.g., the outer unit normal vector $\mathbf{n} = [1, 0]^\texttt{T}$). The problem domain is discretized into $p\times q$ grids with a resolution of $\delta_x$. The discretized coordinate can be denoted by $\mathbf{x}_{ij} = (x_i, y_j)$ where $i=1,2,...,p$ and $j=1,2,...,q$. For the Robin BC, $\alpha$ and $\beta$ are known coefficients. For the periodic BC, $\Gamma_1$ and $\Gamma_2$ denote the left and right boundary, respectively, in this illustration.}
\end{tablenotes}
\end{sidewaystable}

Our network also features the mechanism to encode known terms in the governing PDE via a highway Conv layer (see \textcolor{blue}{Main Text Fig. 1\textbf{a}}) with predefined FD-based filters. For instance, in the numerical example of Section \ref{sec:data_driven_model} where we consider a reaction-diffusion system and assume the diffusion term $\Delta\mathbf{u}$ is known, a Conv layer with discrete Laplacian operator (see Eq. \eref{eq:diff_spatial_filter}) as its filter is created to approximate $\Delta\mathbf{u}$. Note that the associated coefficient for this term is unknown and placed as part of the trainable variable set. Note that by using the residual connection in the recurrent $\Pi$-block, we also implicitly encode the existing term of $\mathbf{u}_t$.
\begin{equation}
    \label{eq:diff_spatial_filter}
    \mathcal{K}_\Delta = \frac{1}{12(\delta x)^2} {
    \begin{bmatrix} 
        0 & 0 & -1 & 0 & 0 \\
        0 & 0 & 16 & 0 & 0 \\
        -1 & 16 & -60 & 16 & -1 \\
        0 & 0 & 16 & 0 & 0 \\  
        0 & 0 & -1 & 0 & 0 
    \end{bmatrix}}
\end{equation}

Such a way of encoding prior physics into the network is very useful in scientific modeling, since for many underexplored dynamical systems we usually have partial knowledge from the past study that can be utilized. By encoding those part of knowledge into the network architecture, the model's prediction would obey the partial physics rigorously. This would significantly narrow down the space of feasible model parameters, and more importantly, reduce the required training effort (e.g. memory, FLOPs, etc.). As the reaction-diffusion systems we studied in this paper are always accompanied with the diffusion phenomenon, the diffusion term (i.e., $\Delta\mathbf{u}$) would be encoded into the network architecture. 

\subsection{Network initialization and pretraining of ISG}

The initialization of trainable parameters for deep neural network is also crucial to the network training. It could significantly affect the accuracy and convergence of the final results \cite{glorot2010understanding}. In traditional network architectures (e.g., DNN or CNN), biases are usually initialized to be zero while the weights are randomly sampled from a Gaussian or uniform distribution. Xavier \cite{glorot2010understanding} and He's \cite{he2015delving} weight initialization are two widely used strategies in deep learning. However, since our network architecture replaces the nonlinear activation function with an elementwise product operation among convolutional feature maps, traditional initialization strategy may result in floating number overflow. From the numerical experiments, we found that a variant of Xavier initialization works well on our network architecture, shown as

\begin{equation} 
    \label{eq:initialization} 
    \displaystyle
    \mathbf{W} \sim \text{Uniform}\left ( -c\cdot\sqrt{\frac{6}{n_i+n_o}},~ c\cdot\sqrt{\frac{6}{n_i+n_o}} \right )
\end{equation}
where $n_i$ and $n_o$ denote the number of input and output neurons respectively, and $c\in (0,1]$ is a small number introduced on the basis of Xavier initialization to scale the range of uniform distribution. The bias is initialized to be zero. 

In addition to the network's parameter, the initial state for the recurrent computation is also important to the accuracy of the trained model. In traditional spatiotemporal sequence modeling \cite{shi2015convolutional}, people initialize all the states to zero or from a Gaussian distribution, which implies no prior knowledge is available about the initial state. Our network seeks to learn the dynamics of a system from a small amount of measurement data, which is expected to be low-resolution (LR) and corrupted by noise. To achieve train high-resolution (HR) predictive model, we employ the ISG to generate the HR initial state from LR measurement for recurrent computation. ISG is essentially a Conv decoder that upsamples the LR measurement. These parameters in ISG, altogether with that in $\Pi$-block, form the set of trainable variable. Note that we pretrain the ISG with the interpolated (e.g., cubic interpolation) HR initial state to ensure a good initialization of the ISG's parameters. In the case where full-resolution\footnote{That said, the available measurement data is in the desired spatial resolution.} measurement data is available, we can directly use one snapshot to initialize the initial state. 


\section{Solving PDEs with PeRCNN}

In this part, we present how to solve the reaction-diffusion equation when the initial and boundary value problem (IBVP) is well defined using the proposed network architecture. To compare our model with existing methods, Convolutional Long-short Term Memory (ConvLSTM) \cite{shi2015convolutional} and physics-informed neural network (PINN) \cite{raissi2019physics} would also be employed as baselines to solve the same problem. 


\subsection{Problem description}

Reaction-diffusion equations are members in the family of PDEs that have found wide applications in computational chemistry and biochemistry. A general reaction-diffusion equation has the form of $
\mathbf{u}_t=\mathbf{D}\Delta \mathbf{u}+\mathbf{R(u)} 
$
where $\mathbf{u}\in\mathbb{R}^n$ is the vector of concentration variables, $\mathbf{D}\in\mathbb{R}^{n\times n}$ is the diagonal diffusion coefficient matrix, $\Delta$ is the Laplacian operator and $\mathbf{R(u)}$ is the reaction vector that represents the interactions between each components of $\mathbf{u}$. 

Without loss of generality, let us assume the reaction-diffusion system features two components, i.e., $\mathbf{u}= [u,~v]^\texttt{T}$, where the state variable/solution is defined over the spatiotemporal temporal domain $\{(\mathbf{x}, t)\} \in \Omega\times\mathcal{T}$. The uniqueness of the solution to this problem is guaranteed by the initial and boundary conditions (I/BCs) of the system. Throughout the paper, we focus on regular physical domains. In this framework for solving the PDE, PeRCNN is employed as a parameterized approximator to the solution on the entire spatiotemporal domain. Once the prediction $\boldsymbol{\widehat{\mathcal{U}}}$ is obtained from the trained model, we compute the residual of the governing PDE using FD method, which is subsequently employed to construct the loss function. The gradient descent method is employed for the minimization of the loss function to obtain a set of model's parameters. It should be noted that the I/BCs are hard encoded into the network architecture using the encoding mechanism introduced in Section \ref{sec:encoding_mechanism}. Therefore, the loss function only consist of the mean squared error (MSE) of the residual of the governing PDE (also known as physics loss), which is
\begin{equation} 
    \label{eq:loss_func_solve_pde} 
    \begin{aligned}
    \mathcal{J}(\boldsymbol{\theta})=\textrm{MSE}\left ( \boldsymbol{\widehat{\mathcal{U}}}_t-\mathcal{F}( \boldsymbol{\widehat{\mathcal{U}}}) \right )
    \end{aligned}
\end{equation}
$\boldsymbol{\widehat{\mathcal{U}}}_t$ is the time derivative of $\boldsymbol{\widehat{\mathcal{U}}}$ computed through numerical discretization while $\mathcal{F}( \boldsymbol{\widehat{\mathcal{U}}})$ is the RHS of Eq. \eref{eq:dynamic_system_pde}). We should be aware that the prediction $\boldsymbol{\widehat{\mathcal{U}}}$ is a function of the network's parameters $\boldsymbol{\theta}$, which are omitted for conciseness.

\subsection{2D problems}\label{sec:solve_pde_2d}

Two types of reaction-diffusion equations are considered in this part, including the Lambda--Omega ($\lambda$--$\Omega$) model as described by Eq. \eref{eq:LO_eqn}
\begin{equation} 
\label{eq:LO_eqn} 
\begin{aligned}
u_t&=\mu_u\Delta u + (1-u^2-v^2)u+\beta (u^2+v^2)v \\
v_t&=\mu_v\Delta v - \beta (u^2+v^2)u + (1-u^2-v^2)v
\end{aligned}
\end{equation}
and FitzHugh-Nagumo (FN) model by Eq. \eref{eq:fn_eqn}.
\begin{equation} 
\label{eq:fn_eqn} 
\begin{aligned}
u_t&=\mu_u\Delta u+u-u^3-v+\alpha \\
v_t&=\mu_v\Delta v+(u-v)\beta
\end{aligned}
\end{equation}
where $\mu_u$ and $\mu_v$ are diffusion coefficients; $\alpha$ and $\beta$ are the reaction coefficients of the system.

\begin{figure}[t!]
\centering
\subfigure[$1.25$s]{
\begin{minipage}[t]{0.24\linewidth}
\centering
\includegraphics[width=\linewidth]{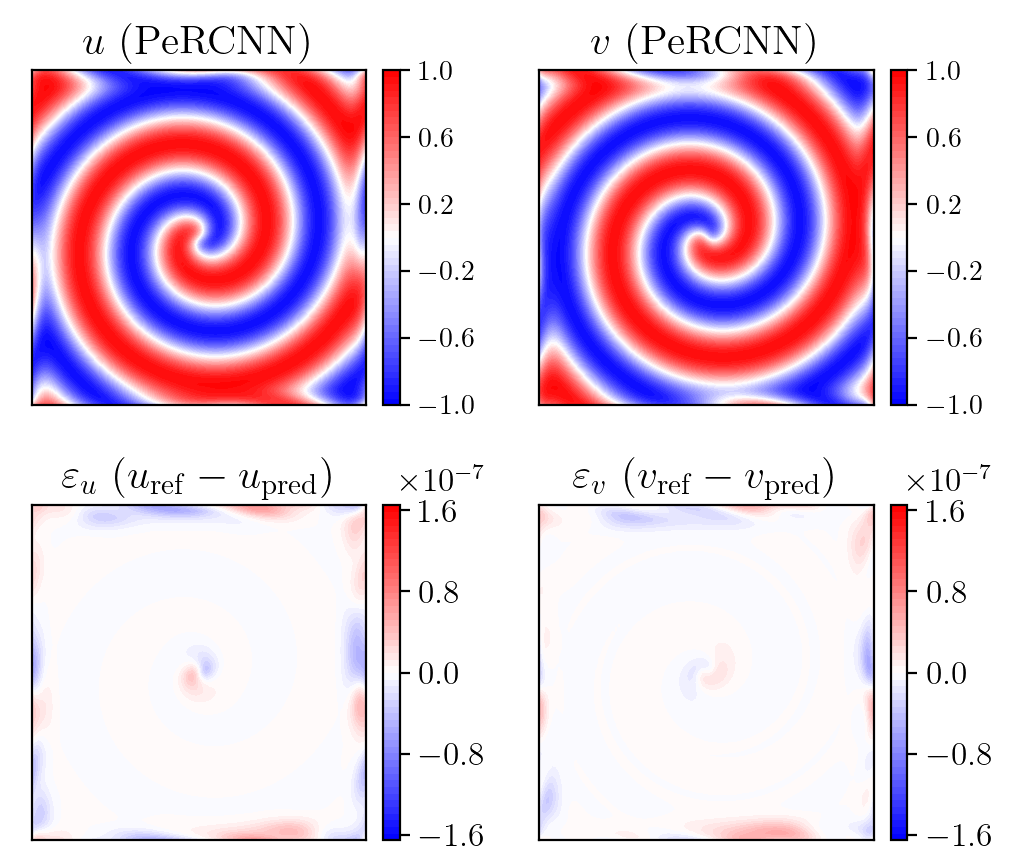}
\end{minipage}%
}%
\subfigure[$2.5$s]{
\begin{minipage}[t]{0.24\linewidth}
\centering
\includegraphics[width=\linewidth]{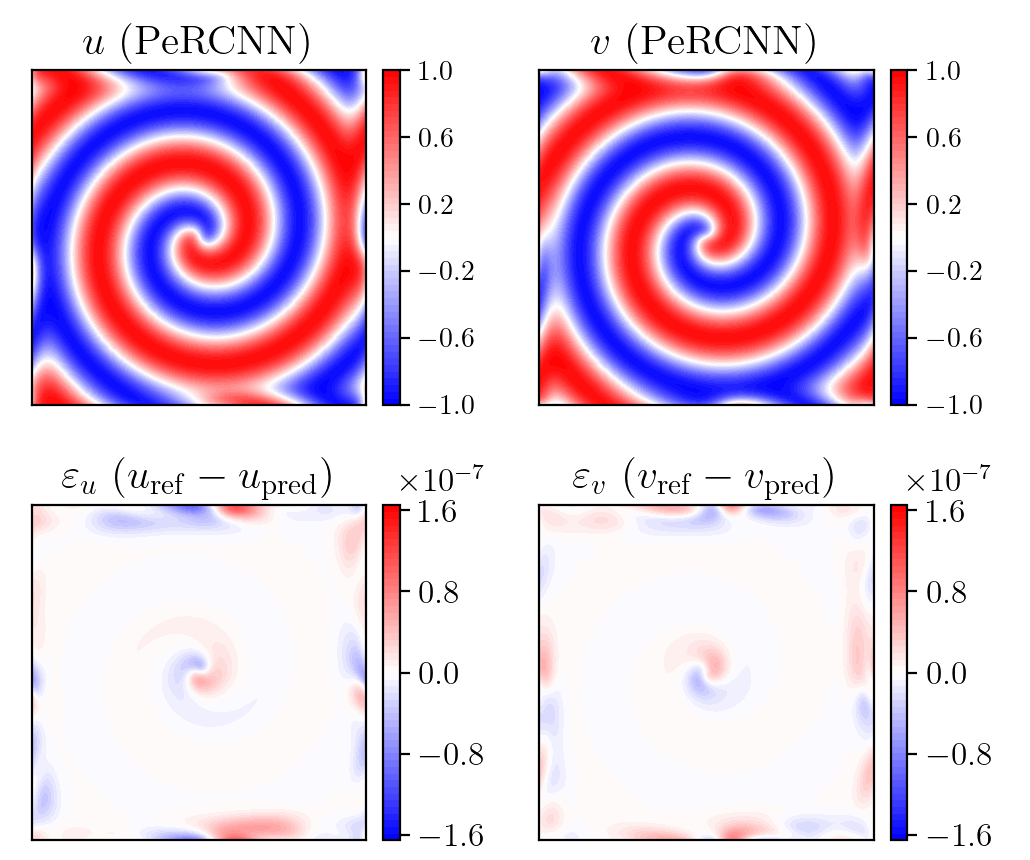}
\end{minipage}%
}%
\subfigure[$3.75$s]{
\begin{minipage}[t]{0.24\linewidth}
\centering
\includegraphics[width=\linewidth]{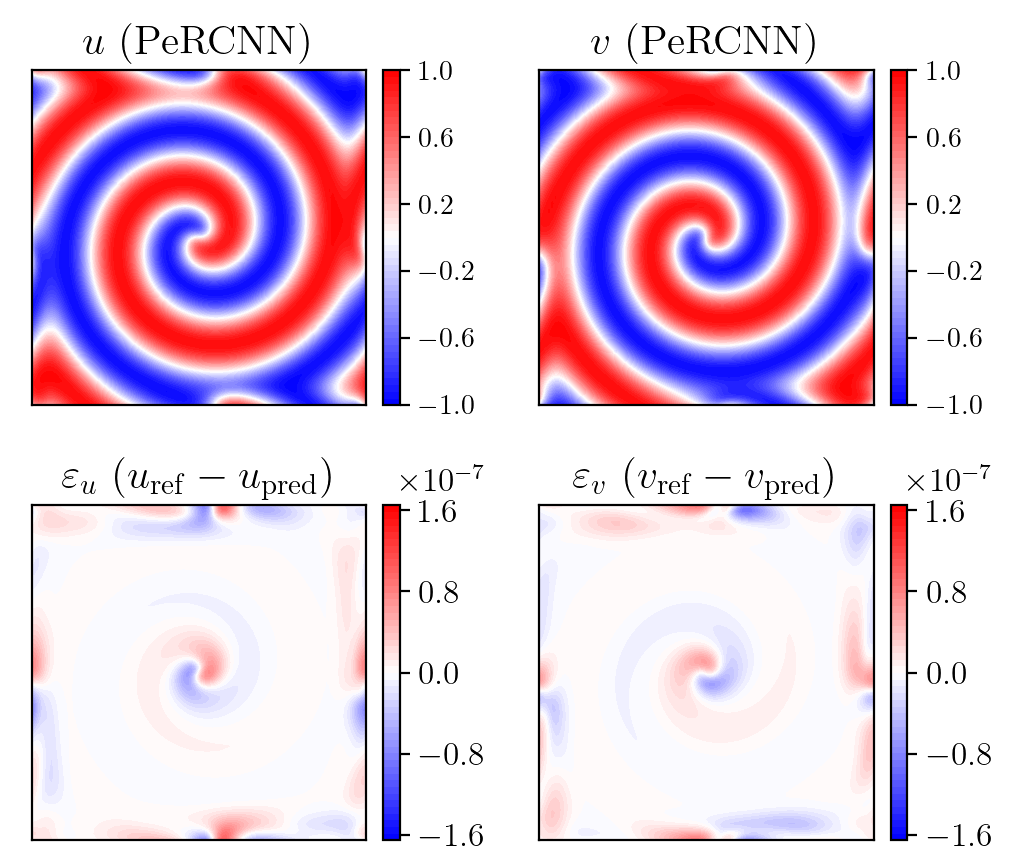}
\end{minipage}%
}%
\subfigure[$5.0$s]{
\begin{minipage}[t]{0.24\linewidth}
\centering
\includegraphics[width=\linewidth]{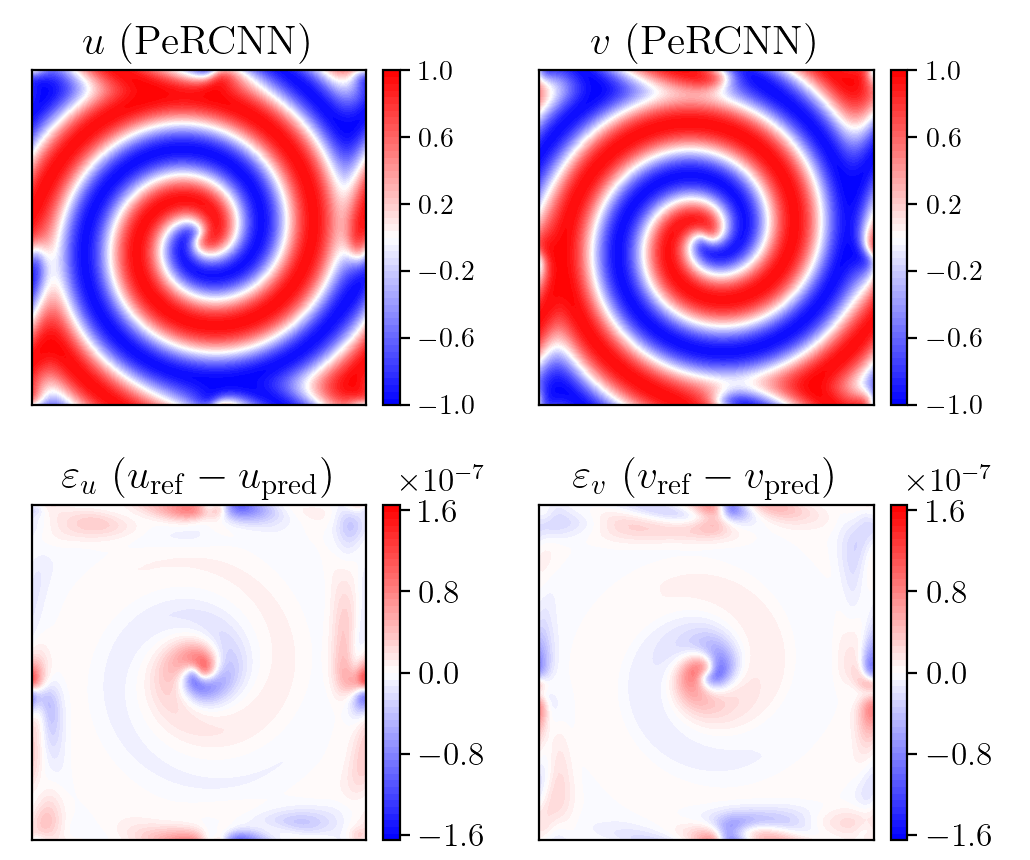}
\end{minipage}%
}%
\hfill \\
\vskip -0.1in
\subfigure[$6.25$s]{
\begin{minipage}[t]{0.24\linewidth}
\centering
\includegraphics[width=\linewidth]{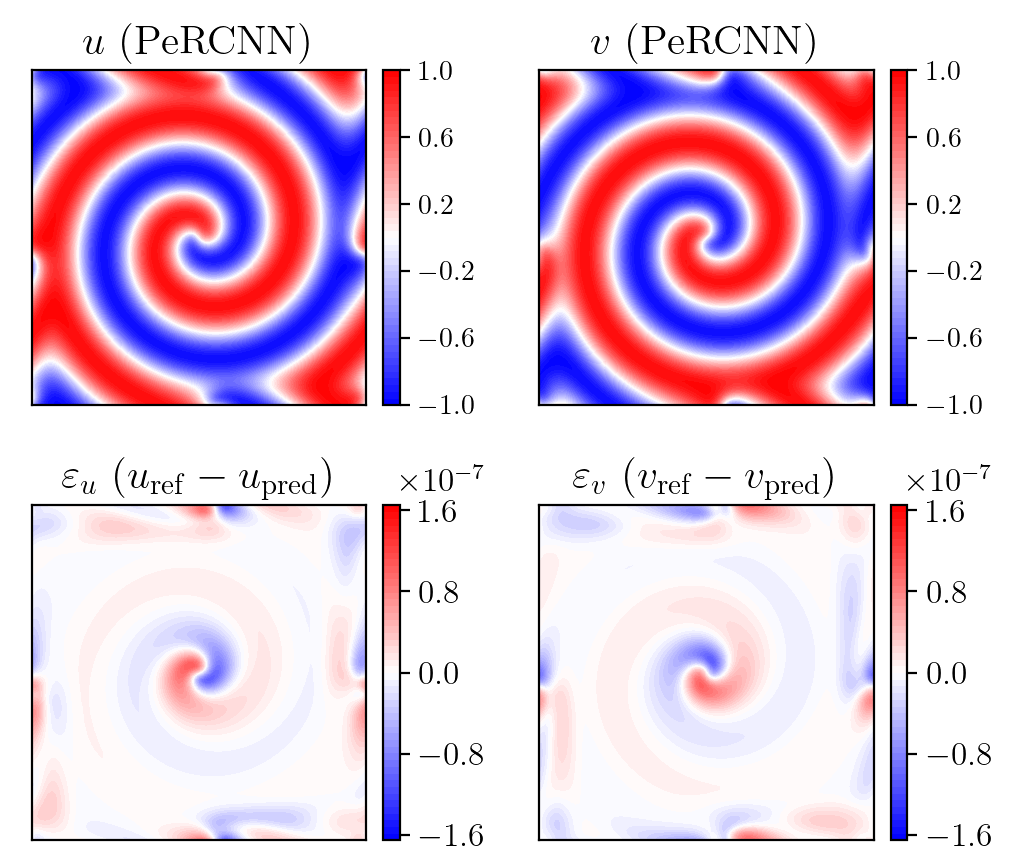}
\end{minipage}
}%
\subfigure[$7.5$s]{
\begin{minipage}[t]{0.24\linewidth}
\centering
\includegraphics[width=\linewidth]{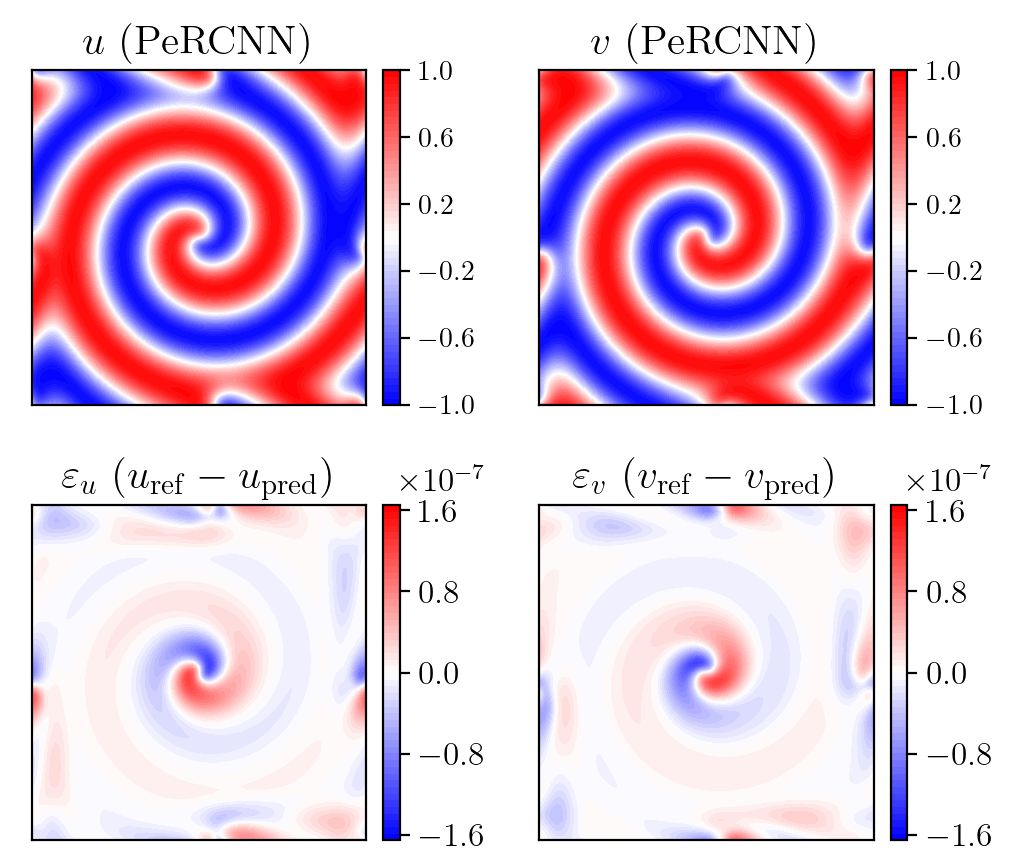}
\end{minipage}
}%
\subfigure[$8.75$s]{
\begin{minipage}[t]{0.24\linewidth}
\centering
\includegraphics[width=\linewidth]{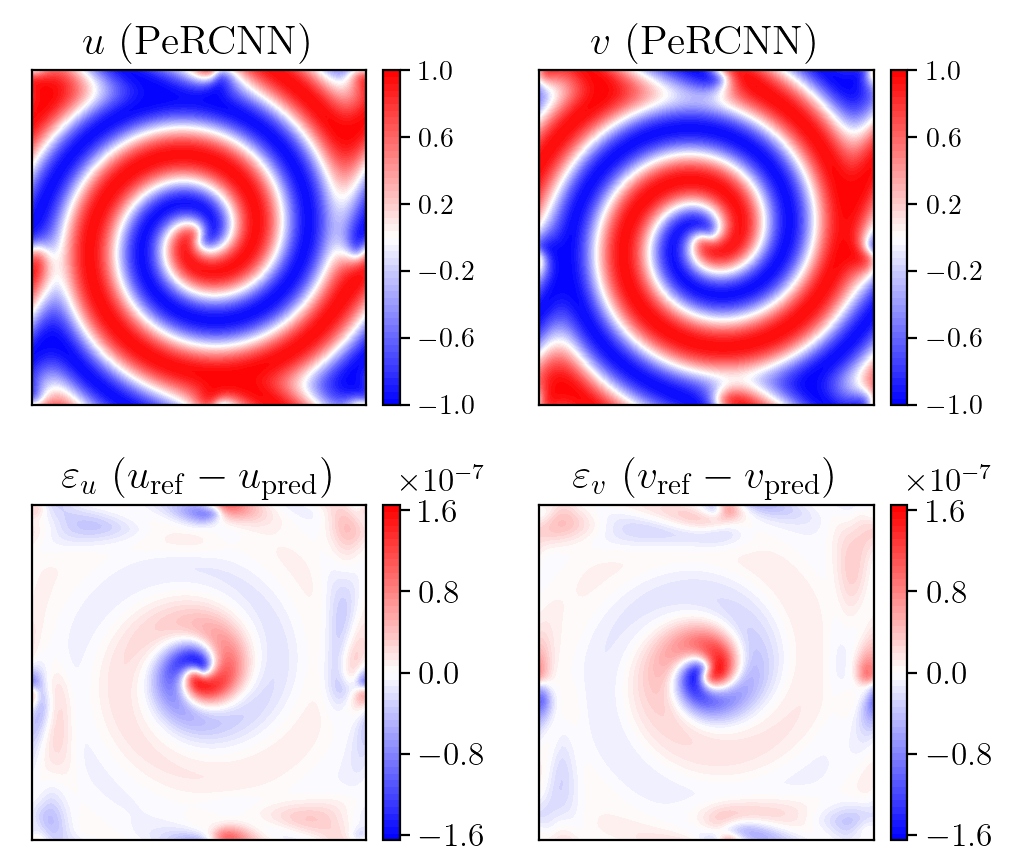}
\end{minipage}
}%
\subfigure[$10.0$s]{
\begin{minipage}[t]{0.24\linewidth}
\centering
\includegraphics[width=\linewidth]{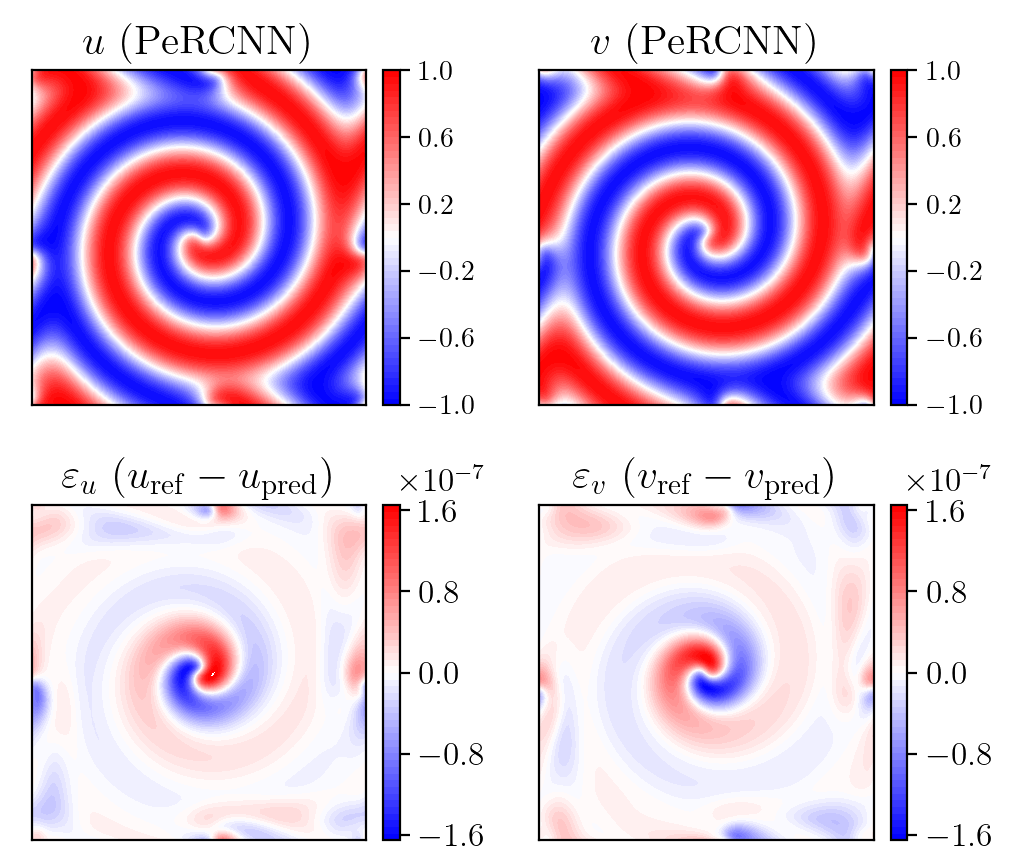}
\end{minipage}
}%
\hfill \\
\vskip -0.1in
\caption{PeRCNN prediction ($\mathbf{u}$) of the $\lambda$--$\Omega$ reaction diffusion system and corresponding error ($\varepsilon_\mathbf{u} =\mathbf{u}_\text{pred}-\mathbf{u}_\text{ref}$). }
\label{fig:pde_solve_lam_omg}
\end{figure}

\begin{figure}[h!]
\centering
\subfigure[$1.5$s]{
\begin{minipage}[t]{0.24\linewidth}
\centering
\includegraphics[width=\linewidth]{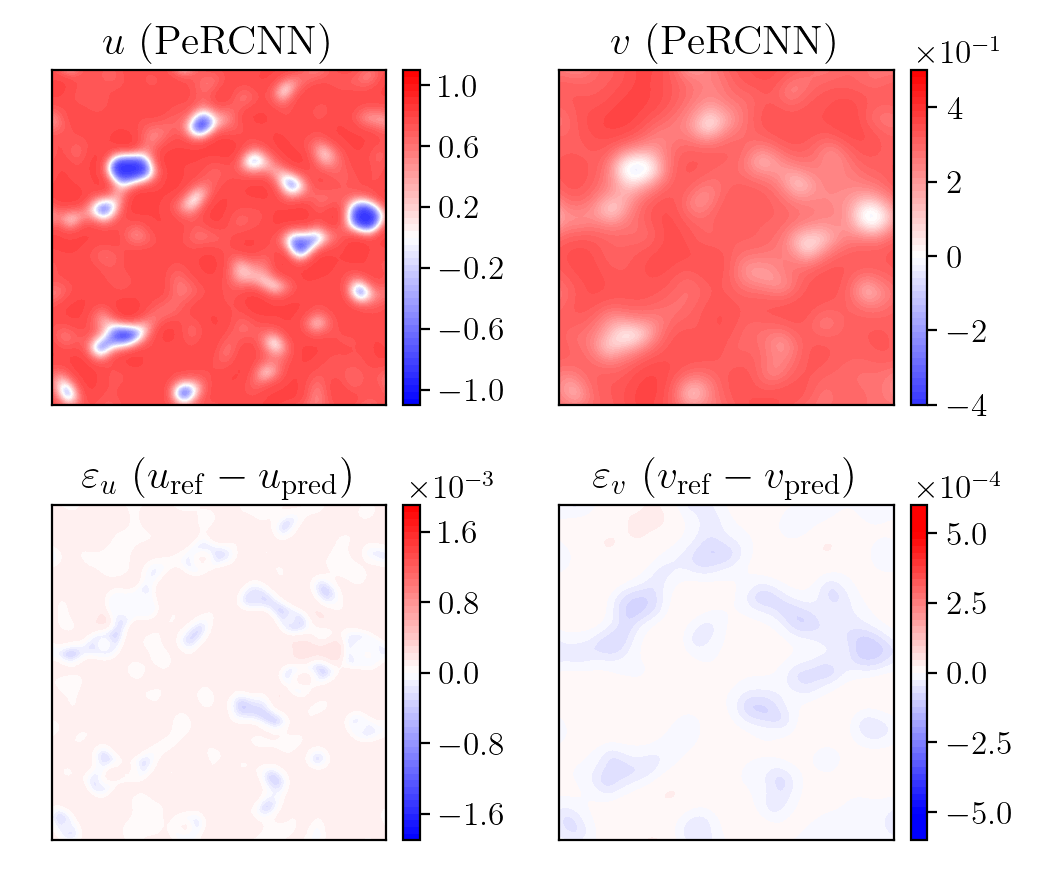}
\end{minipage}%
}%
\subfigure[$3.0$s]{
\begin{minipage}[t]{0.24\linewidth}
\centering
\includegraphics[width=\linewidth]{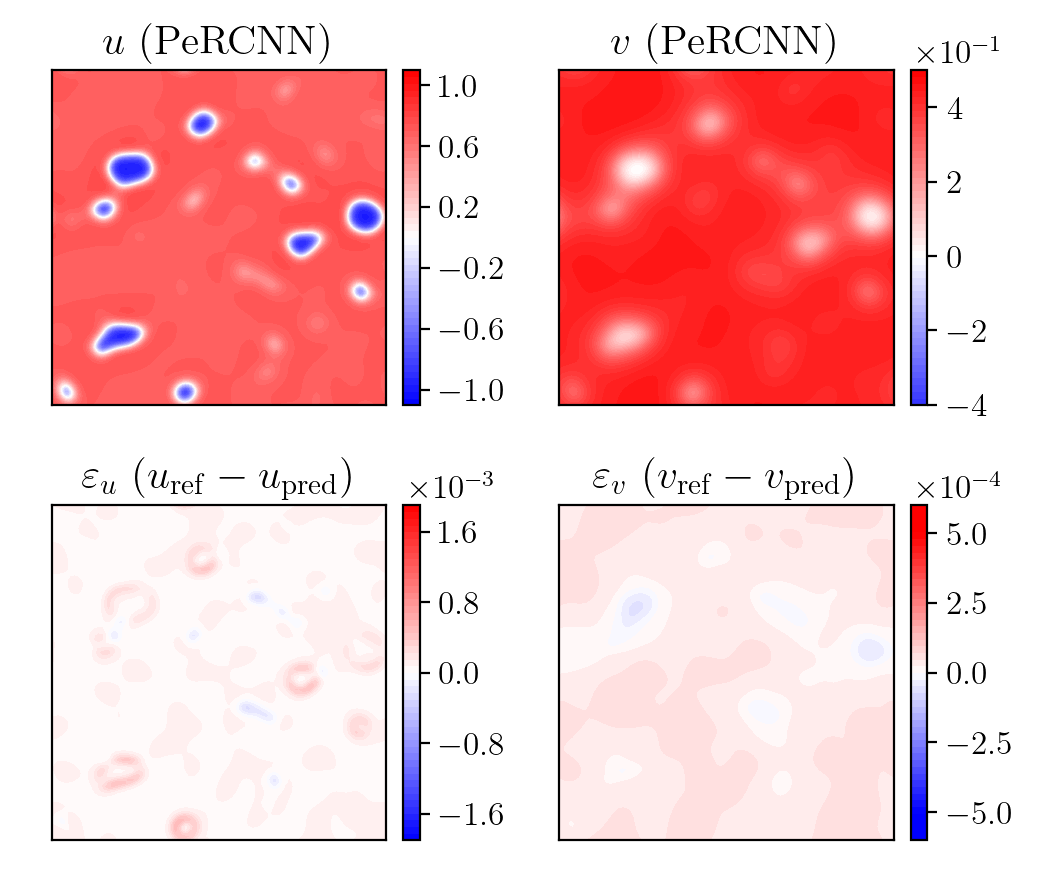}
\end{minipage}%
}%
\subfigure[$4.5$s]{
\begin{minipage}[t]{0.24\linewidth}
\centering
\includegraphics[width=\linewidth]{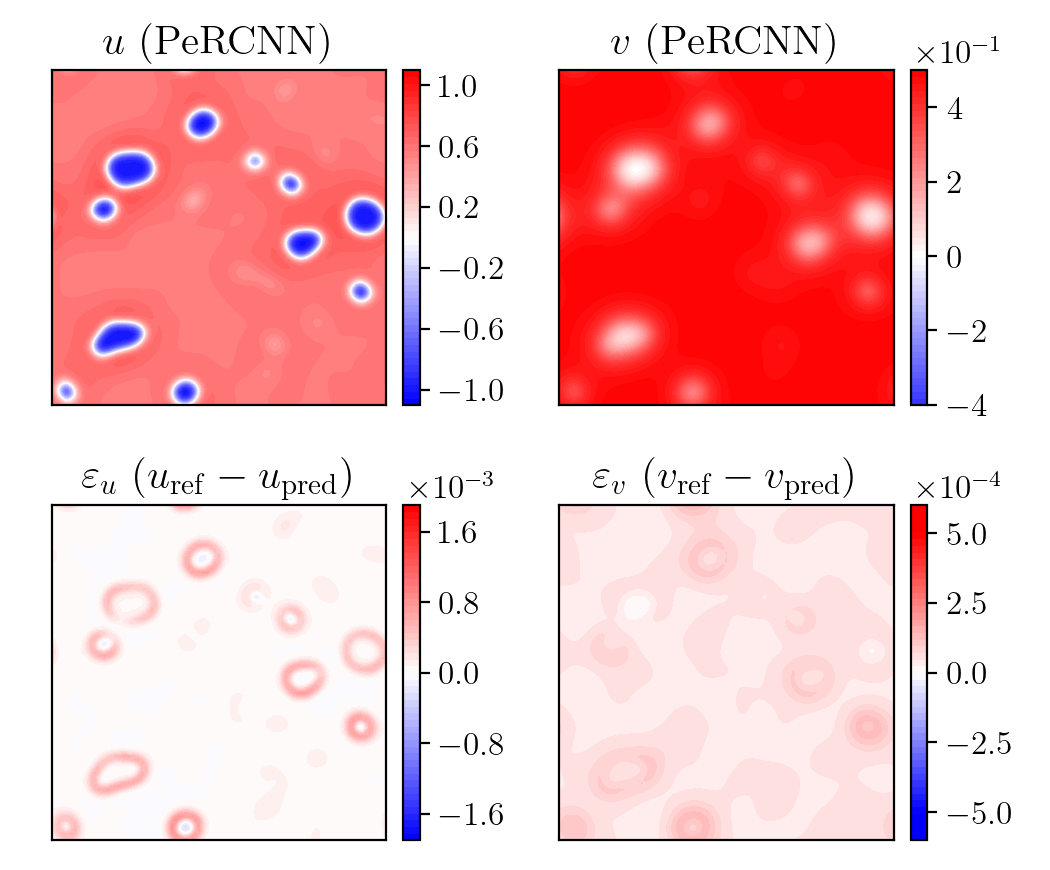}
\end{minipage}%
}%
\subfigure[$6.0$s]{
\begin{minipage}[t]{0.24\linewidth}
\centering
\includegraphics[width=\linewidth]{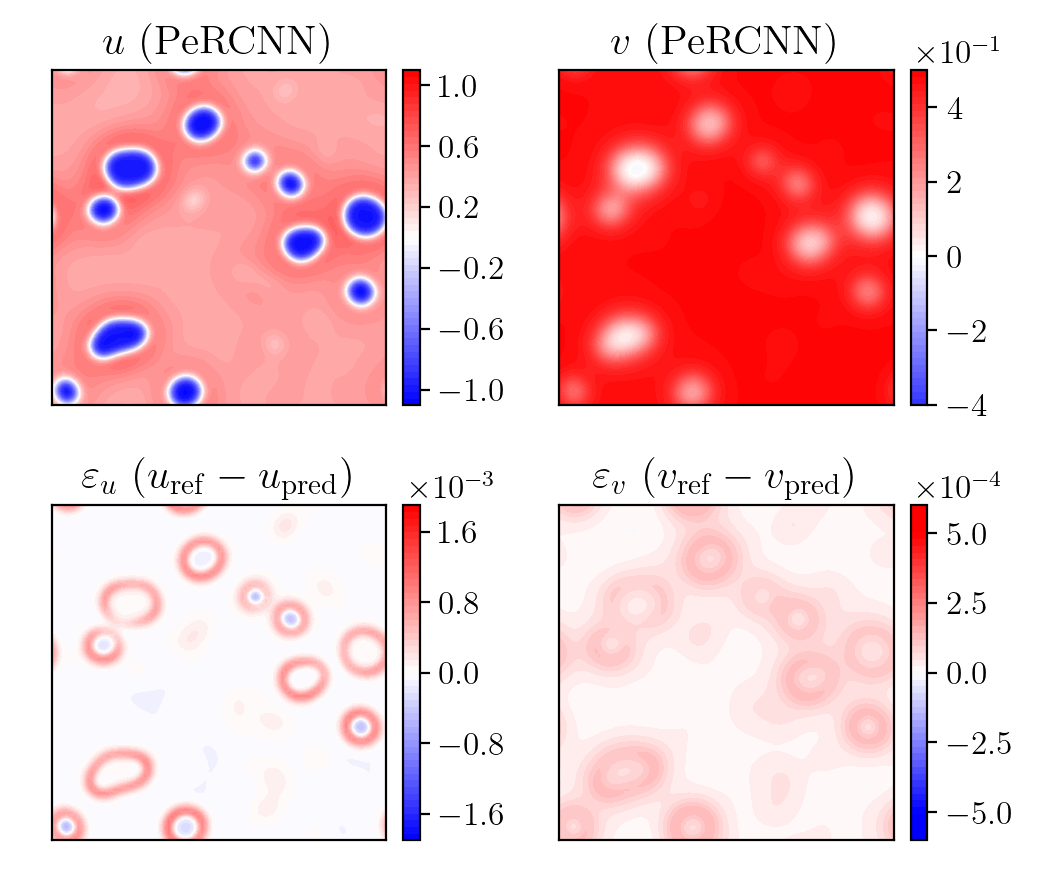}
\end{minipage}%
}%
\hfill \\
\vskip -0.1in
\subfigure[$7.5$s]{
\begin{minipage}[t]{0.24\linewidth}
\centering
\includegraphics[width=\linewidth]{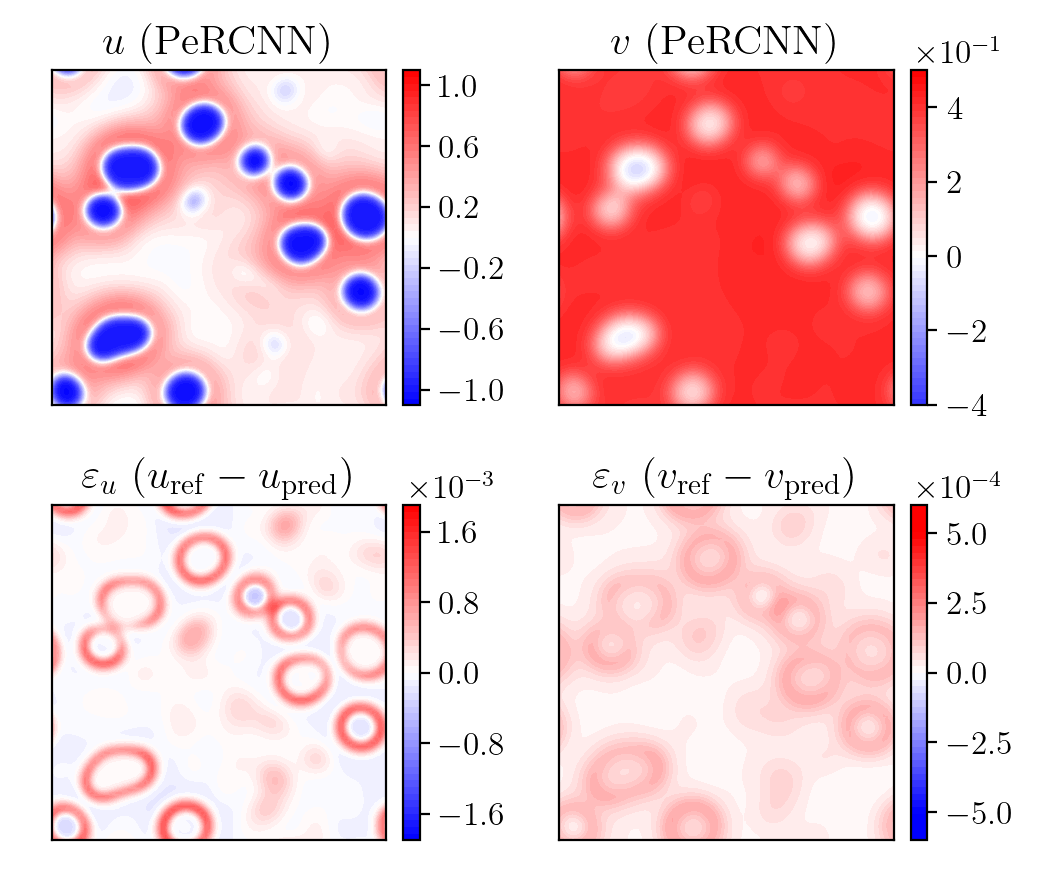}
\end{minipage}
}%
\subfigure[$9.0$s]{
\begin{minipage}[t]{0.24\linewidth}
\centering
\includegraphics[width=\linewidth]{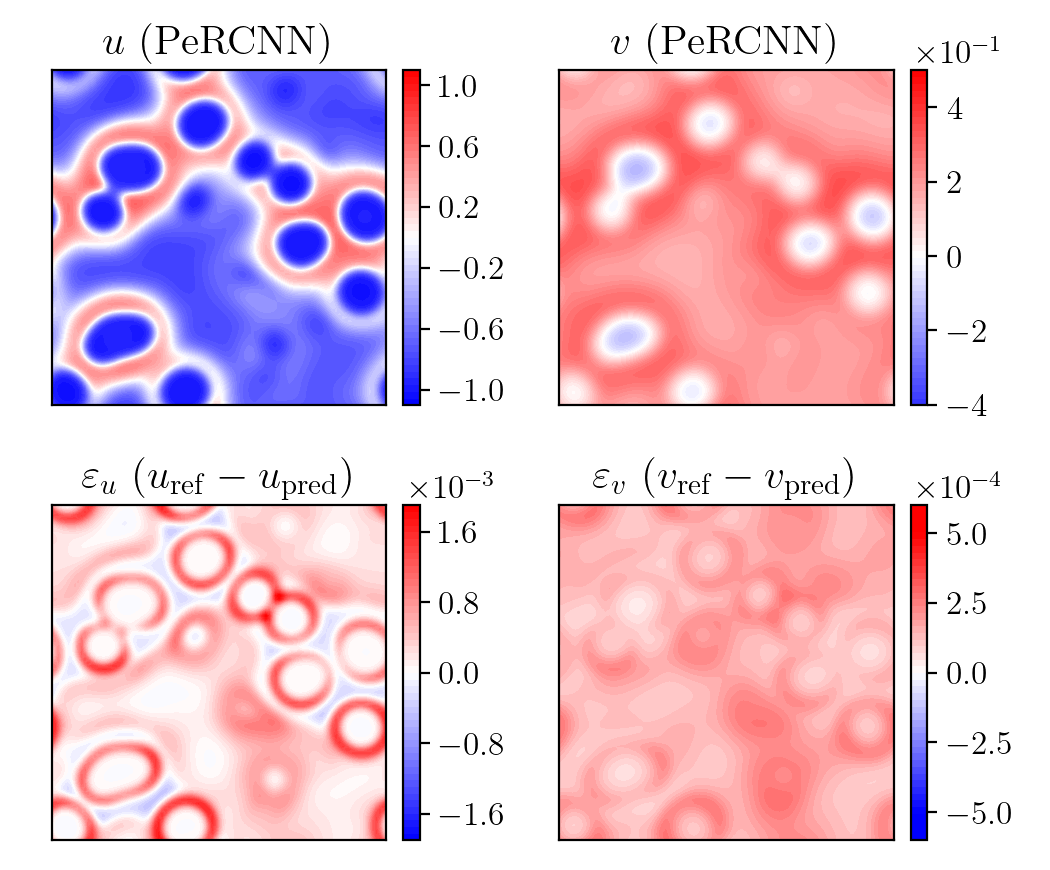}
\end{minipage}
}%
\subfigure[$10.5$s]{
\begin{minipage}[t]{0.24\linewidth}
\centering
\includegraphics[width=\linewidth]{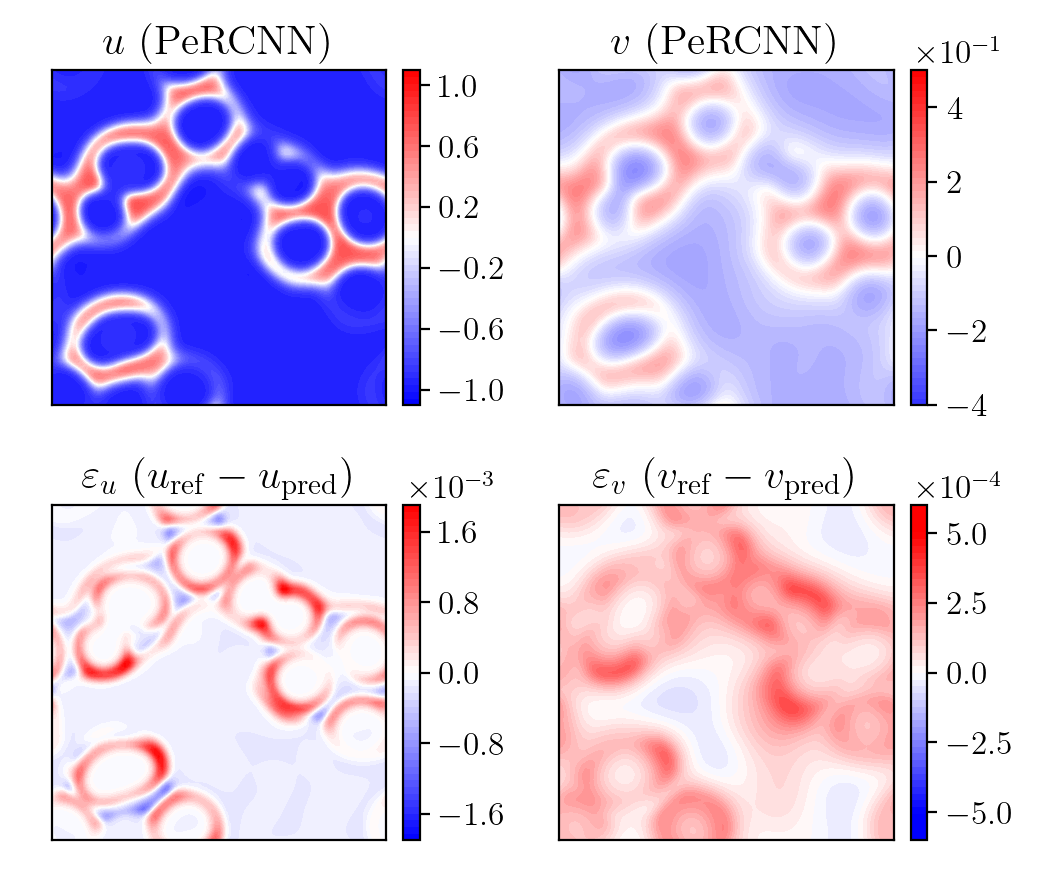}
\end{minipage}
}%
\subfigure[$12.0$s]{
\begin{minipage}[t]{0.24\linewidth}
\centering
\includegraphics[width=\linewidth]{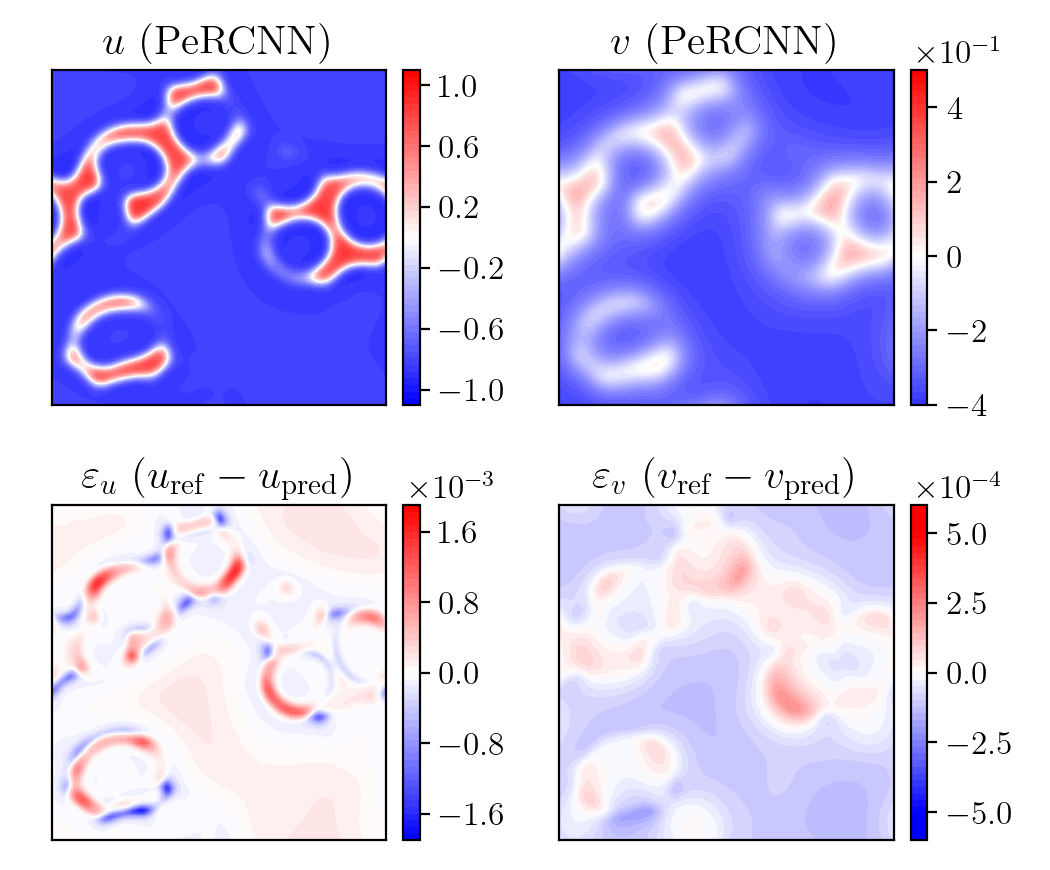}
\end{minipage}
}%
\vskip -0.1in
\caption{PeRCNN prediction ($\mathbf{u}$) of the FN reaction diffusion system and corresponding error ($\varepsilon_\mathbf{u} =\mathbf{u}_\text{pred}-\mathbf{u}_\text{ref}$). }
\label{fig:pde_solve_2d_FN}
\end{figure}

For the case of $\lambda$--$\Omega$ model, we consider a spatial and time domain of $\Omega\in[-10,10]^2$ and $\mathcal{T}\in[0,10]$. The system's parameters $\mu_u$, $\mu_v$ and $\beta$ are selected to be 0.1, 0.1 and 1.0 respectively. The computational domain is discretized to a Cartesian grid with $\delta x$=0.2 and $\delta t$=0.0125, making the shape of PeRCNN prediction to be $\boldsymbol{\widehat{\mathcal{U}}}\in \mathbb{R}^{801\times2\times101\times 101 }$. The periodic boundary condition is employed while the initial condition is prescribed by a well developed field with spiral pattern. The physics loss (see Eq. \eref{eq:loss_func_solve_pde}) would be computed from the PeRCNN prediction using FD method. The network architecture used to solve the PDE consists of 3 parallel Conv layers, each of which has 4 channels. During the training of the network, 10,000 iterations of the Adam optimizer are employed. The initial learning rate is set to be $5\times 10^{-3}$ and decays with the rate of 0.98 every 50 iterations. 

Once the training of the PeRCNN is finished, we retrieve the prediction for plotting  against the reference solution\footnote[1]{The ground truth is generated with the finite difference (FD) solver.}, which is shown in Fig. \ref{fig:pde_solve_lam_omg}. It can be seen that the spiral pattern of the solution is well predicted by the PeRCNN. In the meanwhile, the error of the prediction is much smaller (absolute value up to $10^{-7}$) compared with the scale of the ground truth, which shows the accuracy of the PeRCNN as a PDE solver.

We then consider the FN RD system that features a much slower evolution. This system has the parameters of $\mu_u=1.0$, $\mu_v=10.0$, $\alpha=0.01$ and $\beta=0.25$. The computational domain of $\Omega\in[-50,50]^2$ and $\mathcal{T}\in[0,12]$ is discretized by spacing of $\delta x=1.0$ and $\delta t=0.002$, which leads to the predicted solution $\boldsymbol{\widehat{\mathcal{U}}}\in \mathbb{R}^{6001\times 2\times101\times 101 }$. The initial condition of the problem is sampled from the Gaussian distribution with zero mean and standard deviation of 0.05. The PeRCNN used to approximate the solution has 3 parallel hidden layers while each layer has 4 channels. Considering the large number of time steps in this problem, we conduct the training with two time batches using Adam optimizer. It is noteworthy that the pretraining with a smaller number of time steps is found to accelerate the whole training process. Therefore, we start training the recurrent network with 501 time steps, and then extend gradually from 1001, 3001 steps to the total 6001 steps. Due to the good extrapolation capability of the PeRCNN, which would be discussed with more details in Section \ref{sec:data_driven_model}, the number of iterations required for training can be reduced significantly as the number of time steps increases. Specifically, 5000, 3000, 1000 and 500 iterations are employed in each stage of the training process. The learning rate starts at $2\times 10^{-2}$ and decays by 0.98 every 50 iterations throughout all 4 stages. The prediction given by our PeRCNN and the corresponding error are provided in Fig. \ref{fig:pde_solve_2d_FN}. Despite the very large number of simulated time steps, PeRCNN still gives satisfactory prediction compared with reference solution. This example demonstrates the excellent scalability of the proposed PeRCNN for problems with long time pattern. 


\begin{figure}[t!]
\begin{center}
\begin{varwidth}{0.14\textwidth}
\centering\bfseries
{\small PeRCNN \\prediction\par}
\end{varwidth}%
\begin{tabular}{@{}c@{}}
\includegraphics[width=0.75\textwidth]{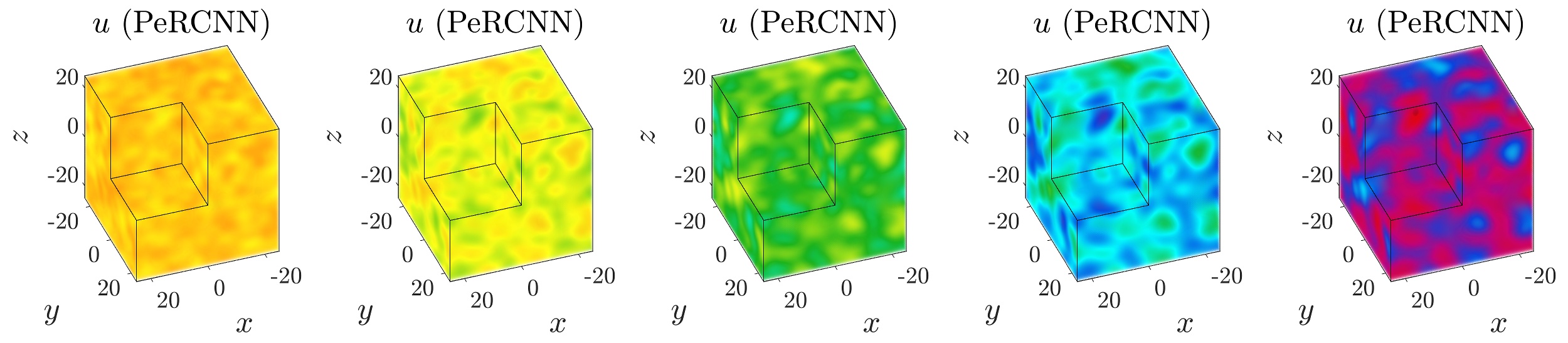}
\end{tabular}%
\hfill \\
\begin{varwidth}{0.14\textwidth}
\centering\bfseries
{\small Reference\par}
\end{varwidth}%
\begin{tabular}{@{}c@{}}
\includegraphics[width=0.75\textwidth]{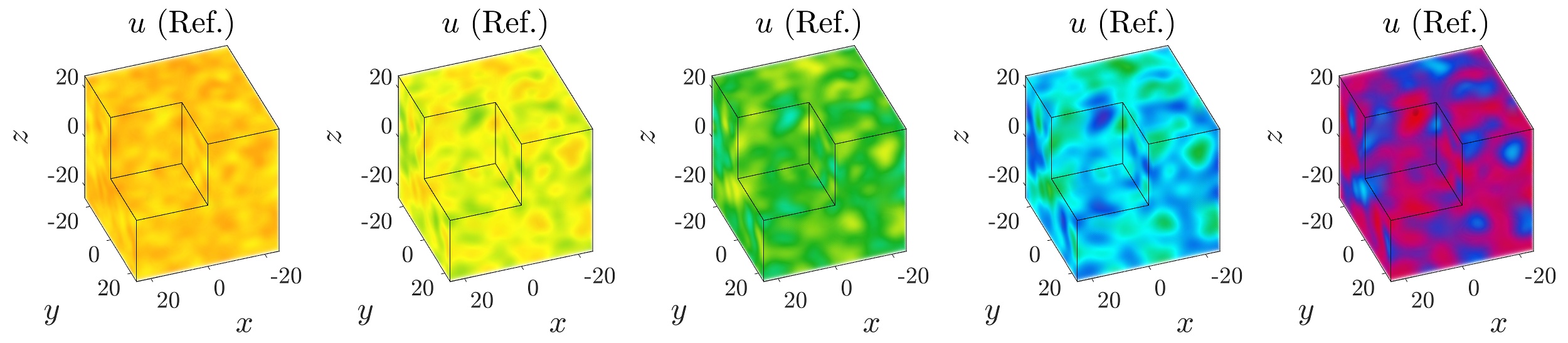}
\end{tabular}%
\hfill \\
\begin{varwidth}{0.14\textwidth}
\centering\bfseries
{\small PeRCNN \\prediction\par}
\end{varwidth}%
\begin{tabular}{@{}c@{}}
\includegraphics[width=0.75\textwidth]{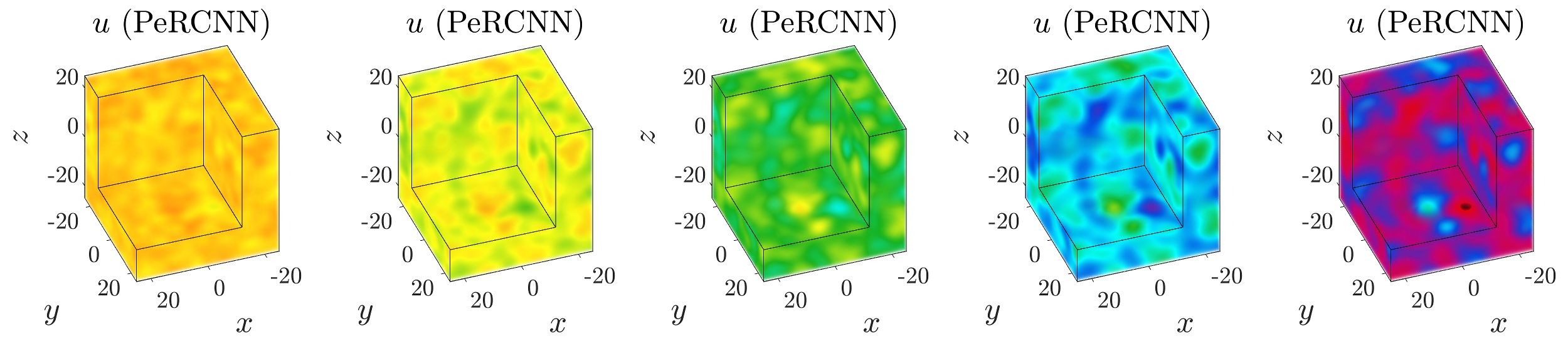}
\end{tabular}%
\hfill \\
\begin{varwidth}{0.14\textwidth}
\centering\bfseries
{\small Reference\par}
\end{varwidth}%
\begin{tabular}{@{}c@{}}
\includegraphics[width=0.75\textwidth]{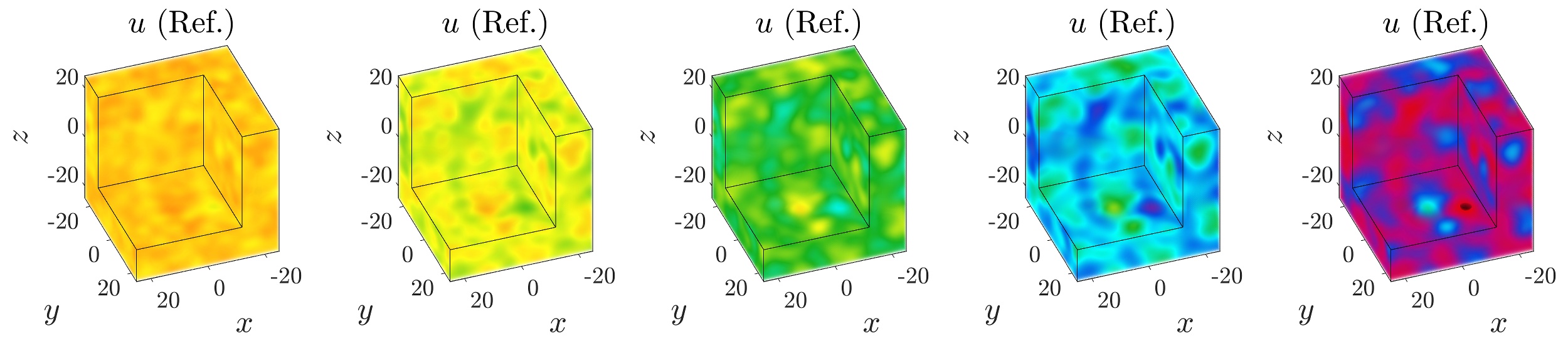}
\end{tabular}%
\hfill \\
\begin{tabular}{@{}c@{}}
\centering
\includegraphics[width=0.14\textwidth]{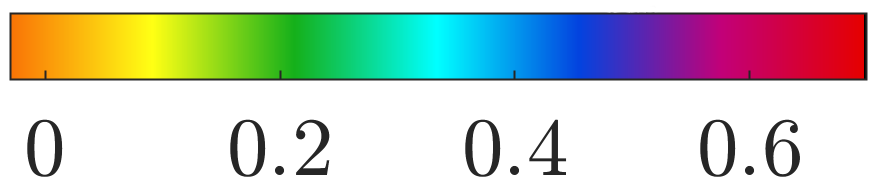}
\end{tabular}%
\hfill \\
\small
\begin{tabular*}{0.8\textwidth}{@{\extracolsep{\fill}} c c c c c c}
~~~~~~~~~ & (a) 0.8 s &(b) 1.6 s &(c)  2.4 s &(d) 3.2 s &(e) 4.0 s
\end{tabular*}
\hfill
\end{center}
\caption{PeRCNN prediction ($u$) of the 3D FitzHugh-Nagumo reaction diffusion system and the reference solution (Top two rows: sliced at $1/2$ of the edges; Bottom two rows: sliced at $3/4$ of the edges).}
\label{pde_solve_3d_FN_u}
\end{figure}

\begin{figure}[h]
\begin{center}
\begin{varwidth}{0.14\textwidth}
\centering\bfseries
{\small PeRCNN \\prediction\par}
\end{varwidth}%
\begin{tabular}{@{}c@{}}
\includegraphics[width=0.75\textwidth]{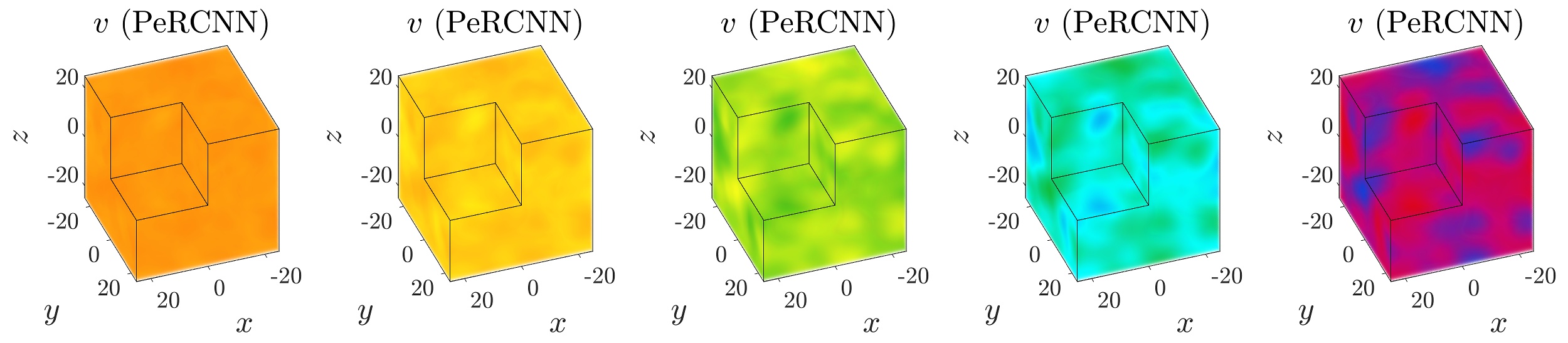}
\end{tabular}%
\hfill \\
\begin{varwidth}{0.14\textwidth}
\centering\bfseries
{\small Reference\par}
\end{varwidth}%
\begin{tabular}{@{}c@{}}
\includegraphics[width=0.75\textwidth]{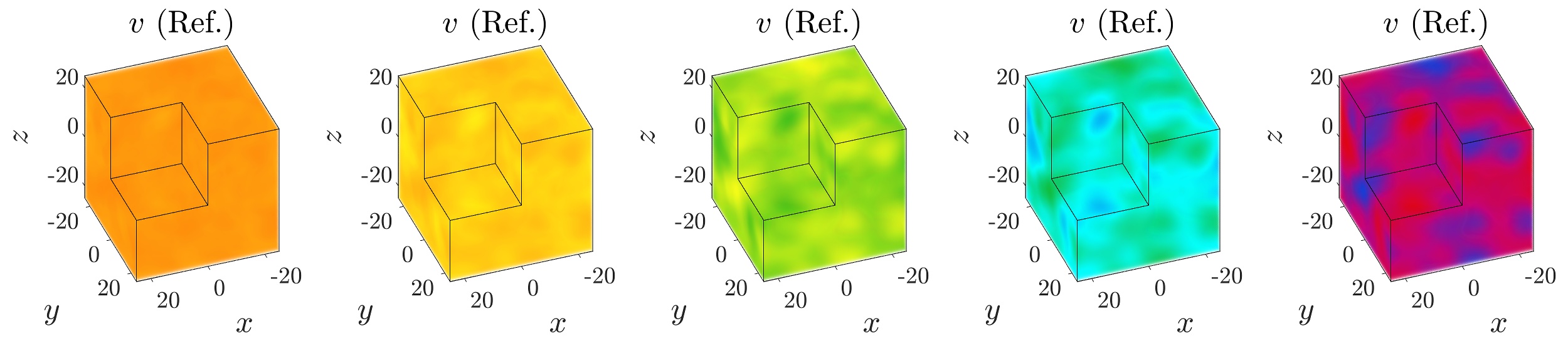}
\end{tabular}%
\hfill \\
\begin{varwidth}{0.14\textwidth}
\centering\bfseries
{\small PeRCNN \\prediction\par}
\end{varwidth}%
\begin{tabular}{@{}c@{}}
\includegraphics[width=0.75\textwidth]{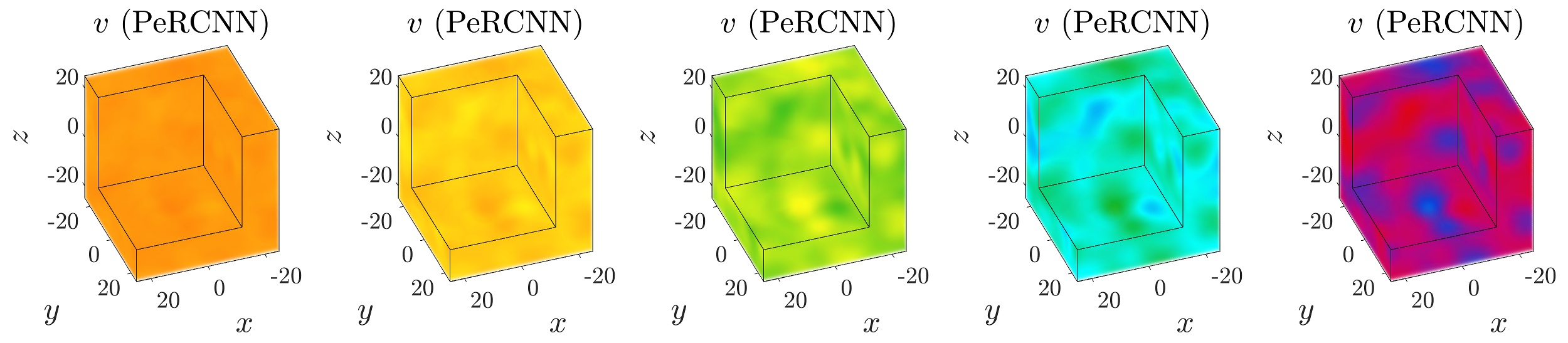}
\end{tabular}%
\hfill \\
\begin{varwidth}{0.14\textwidth}
\centering\bfseries
{\small Reference\par}
\end{varwidth}%
\begin{tabular}{@{}c@{}}
\includegraphics[width=0.75\textwidth]{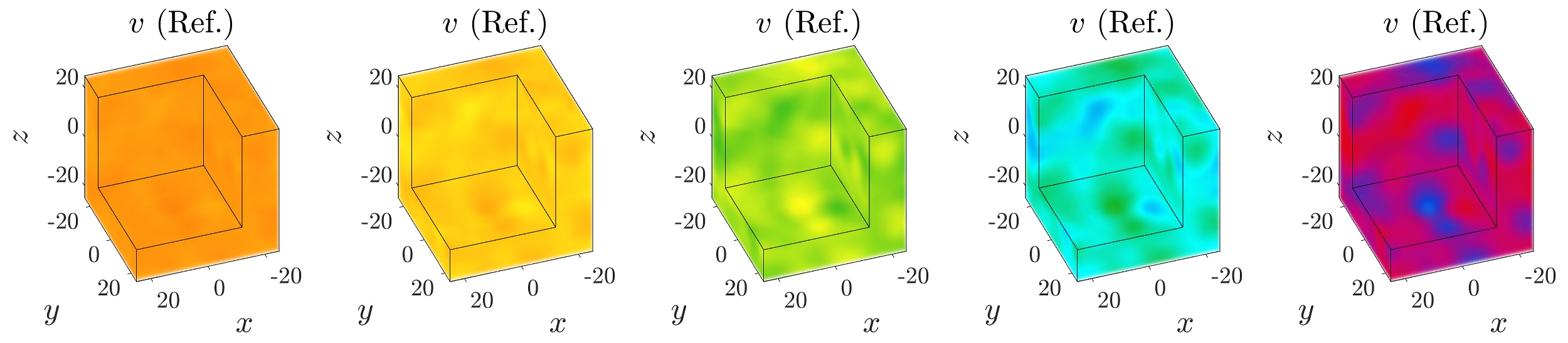}
\end{tabular}%
\hfill \\
\begin{tabular}{@{}c@{}}
\includegraphics[width=0.14\textwidth]{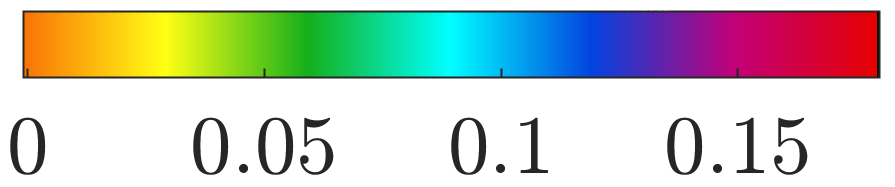}
\end{tabular}%
\hfill \\
\small
\begin{tabular*}{0.8\textwidth}{@{\extracolsep{\fill}} c c c c c c}
~~~~~~~~~ & (a) 0.8 s &(b) 1.6 s &(c)  2.4 s &(d) 3.2 s &(e) 4.0 s
\end{tabular*}
\hfill
\end{center}
\caption{PeRCNN prediction ($v$) of the 3D FitzHugh-Nagumo reaction diffusion system and the reference solution (Top two rows: sliced at $1/2$ of the edges; Bottom two rows: sliced at $3/4$ of the edges).}
\label{pde_solve_3d_FN_v}
\end{figure}

\begin{figure}[t]
\begin{tabular}{m{0.15\textwidth}  r }
{\bfseries \small Prediction ($u$)} & \begin{minipage}{0.8\textwidth}\includegraphics[width=\textwidth]{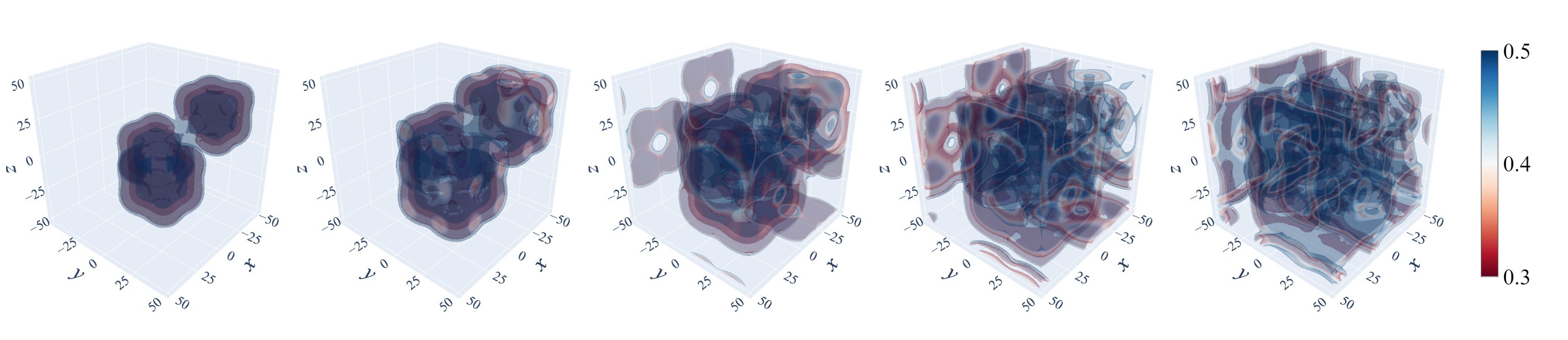}\end{minipage} \\
{\bfseries \small Error ($\varepsilon_u$)}  & \begin{minipage}{0.8\textwidth}\includegraphics[width=\textwidth]{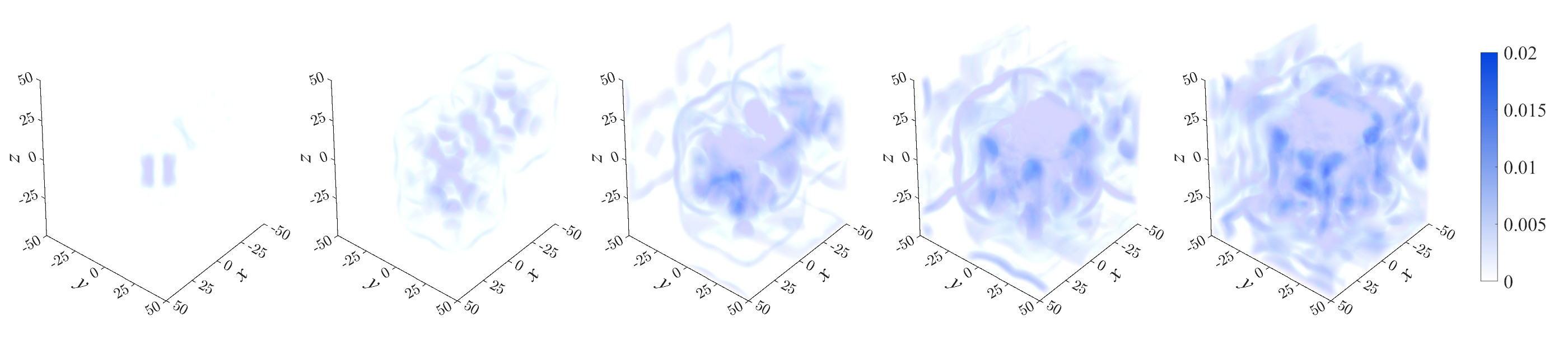}\end{minipage}  \\
{\bfseries \small Prediction ($v$)} & \begin{minipage}{0.8\textwidth}\includegraphics[width=\textwidth]{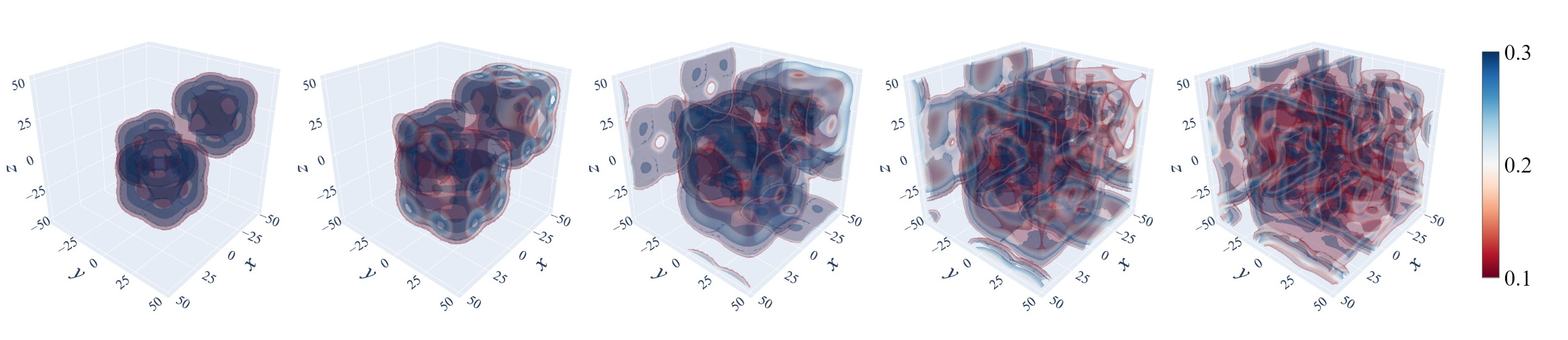}\end{minipage}\\
{\bfseries \small Error ($\varepsilon_v$)}  & \begin{minipage}{0.8\textwidth}\includegraphics[width=\textwidth]{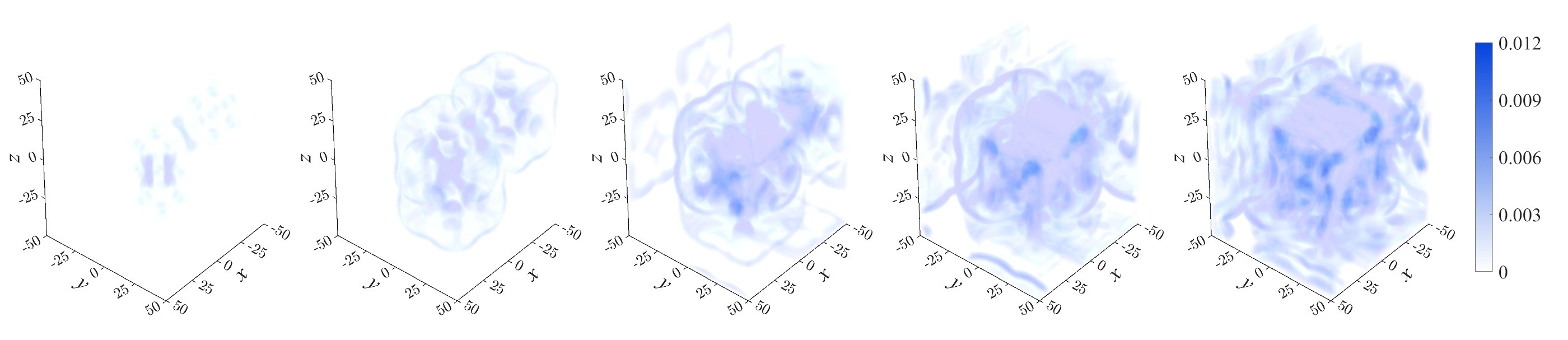}\end{minipage} 
\end{tabular}
\caption{Isosurfaces of $u$ and $v$ for PeRCNN prediction (3D Gary-Scott model) and corresponding error against reference solution plotted on point cloud.}
\label{pde_solve_3d_GS_v}
\end{figure}

\subsection{3D problems}\label{sec_3d_pde_solve}

Due to the curse of dimensionality, solving PDEs on 3D spatial domain is considered to more challenging than the 2D problems. In this section, we explore the scalability of the proposed PeRCNN architecture on 3D RD systems. The RD system of Gray-Scott (GS) model described by
\begin{equation} 
\label{eq:gs_eqn} 
\begin{aligned}
u_t&=\mu_u\Delta u-uv^2+F(1-u) \\
v_t&=\mu_v\Delta v+uv^2-(F+\kappa)v
\end{aligned}
\end{equation}
where $\kappa$ and $f$ denote the kill and feed rate respectively, and the aforementioned FN model described by Eq. \eref{eq:fn_eqn} are primarily studied. Through these two numerical examples, we demonstrate that the proposed PeRCNN is able to solve the 3D problems of high dimensionality while some of the state-of-the-art approaches, e.g., ConvLSTM and PINN, struggle.

We first consider the 3D FN RD system with $\mu_u=1.0$, $\mu_v=10.0$, $\alpha=0.01$ and $\beta=0.25$. The computational domain of $\Omega\times \mathcal{T} \in[-25,25]^3\times[0,4]$ is discretized with $\delta x=1.0$ and $\delta t=0.004$ in spatial and time dimension respectively, resulting to the prediction $\boldsymbol{\widehat{\mathcal{U}}}\in \mathbb{R}^{ 1001\times2\times51\times 51 \times51}$. The initial condition of the problem is sampled from the Gaussian distribution with zero mean and standard deviation of 0.05 while periodic BC is adopted. The network architecture employed to solve the PDE features 3 parallel hidden layers and 4 channels each layer. Similar to the 2D FN case in Section \ref{sec:solve_pde_2d}, the pretraining is conducted with 101, 201 and 501 time (prediction) steps sequentially before the final training of 1001 steps. The corresponding number of iterations for Adam optimizer are 4000, 2000, 1000 and 200 respectively. The learning rate starts at $2.5\times10^{-2}$ and decays with the rate of 0.98 every 100 iterations. We plot the prediction of PeRCNN, as well as the reference solution, in Fig. \ref{pde_solve_3d_FN_u} and \ref{pde_solve_3d_FN_v} respectively. Part of the domain is cut for better visualization. It can be seen that the range of both $u$ and $v$ component vary rapidly with regard to time. However, the PeRCNN prediction still agrees well with the reference solution. The root mean squared error (RMSE) of the full-field prediction are $4.4\times10^{-4}$ and $5.8\times10^{-4}$ for $u$ and $v$ respectively.

We then solve the 3D GS RD system described by Eq. \eref{eq:gs_eqn} with $\mu_u=0.2$, $\mu_v=0.1$, $F=0.025$ and $\kappa=0.055$. The computational domain is $\Omega\times\mathcal{T}=[-50, 50]^3\times[0,750]$. As for the initial condition, three initial disturbances are placed randomly on the constant background of $u$ and $v$. The spatial and time spacing for the discretization are $\delta x=2.0$ and $\delta t=0.5$ respectively. That said, the prediction from the PeRCNN would be $\boldsymbol{\widehat{\mathcal{U}}}\in \mathbb{R}^{ 1501\times2\times51\times 51 \times51}$. The pretraining is performed with 101, 251 and 501 time steps sequentially before the final training of whole 1501 time steps split by two batches. The number of iterations with Adam optimizer for each stage are 5000, 3000, 1000 and 200 respectively. The learning rate starts at $10^{-2}$ and decays at the rate of 0.975 every 200 iterations. The isosurfaces of the PeRCNN's prediction are provided in Fig. \ref{pde_solve_3d_GS_v}, along with the corresponding error against the reference solution using points cloud. Overall, PeRCNN prediction has only minor discrepancies and agrees well with the reference solution. The RMSE of the full-field prediction is $4.1\times10^{-3}$. 

\subsection{Comparison with existing methods}

\begin{figure}[h!]
\begin{center}
\includegraphics[width=0.7\textwidth]{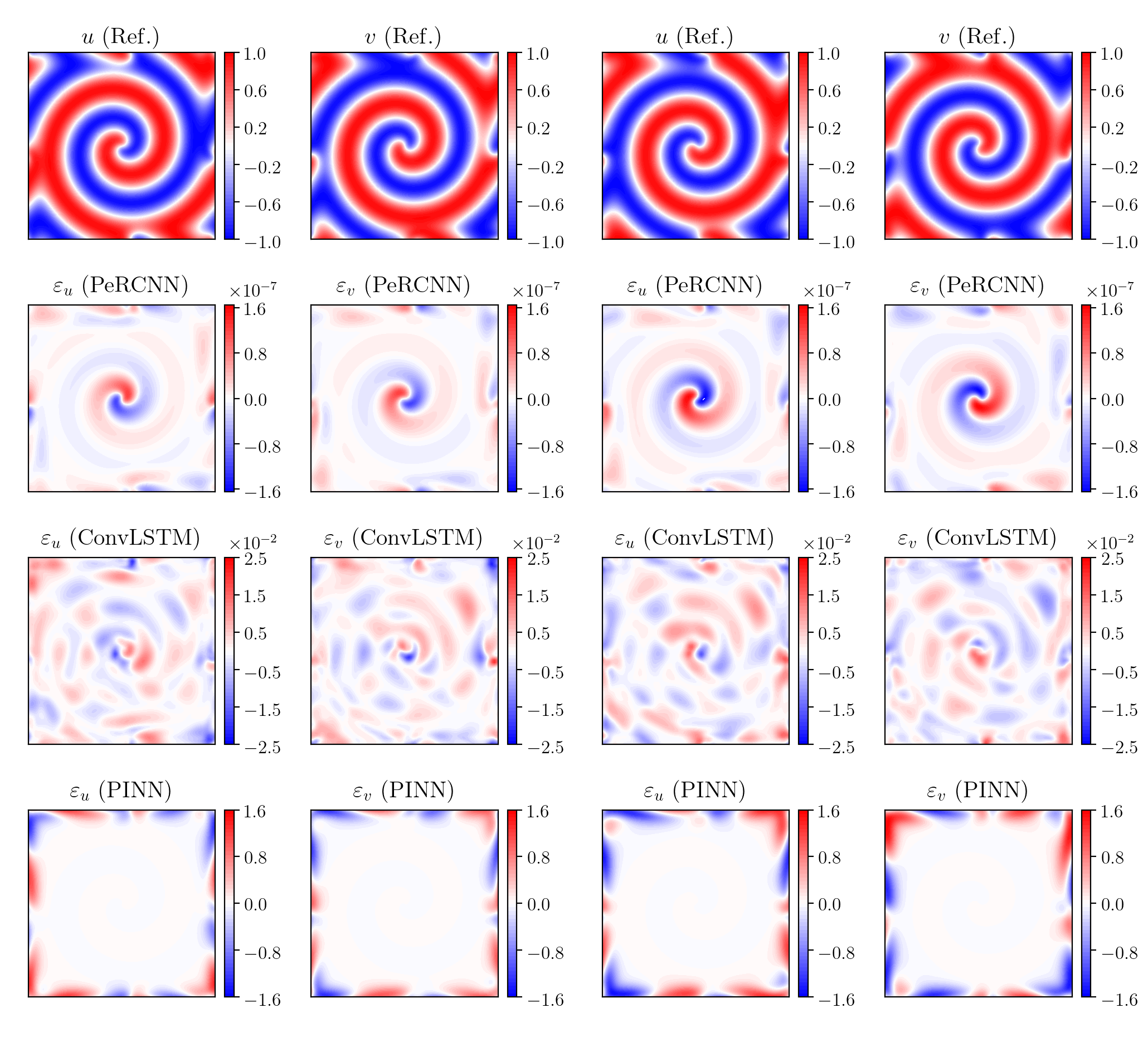}
\small
\begin{tabular*}{0.45\textwidth}{@{\extracolsep{\fill}} c c}
(a)   7.5 s & (b)  10.0 s
\end{tabular*}
\hfill
\end{center}
\vskip -0.2in
\caption{Comparison of the error against reference solution for PeRCNN, ConvLSTM and PINN prediction on $\lambda$-$\Omega$ model.}
\label{fig:contrast_2d_Lam_Omg}
\end{figure}

\begin{figure}[h!]
\begin{center}
\includegraphics[width=0.7\textwidth]{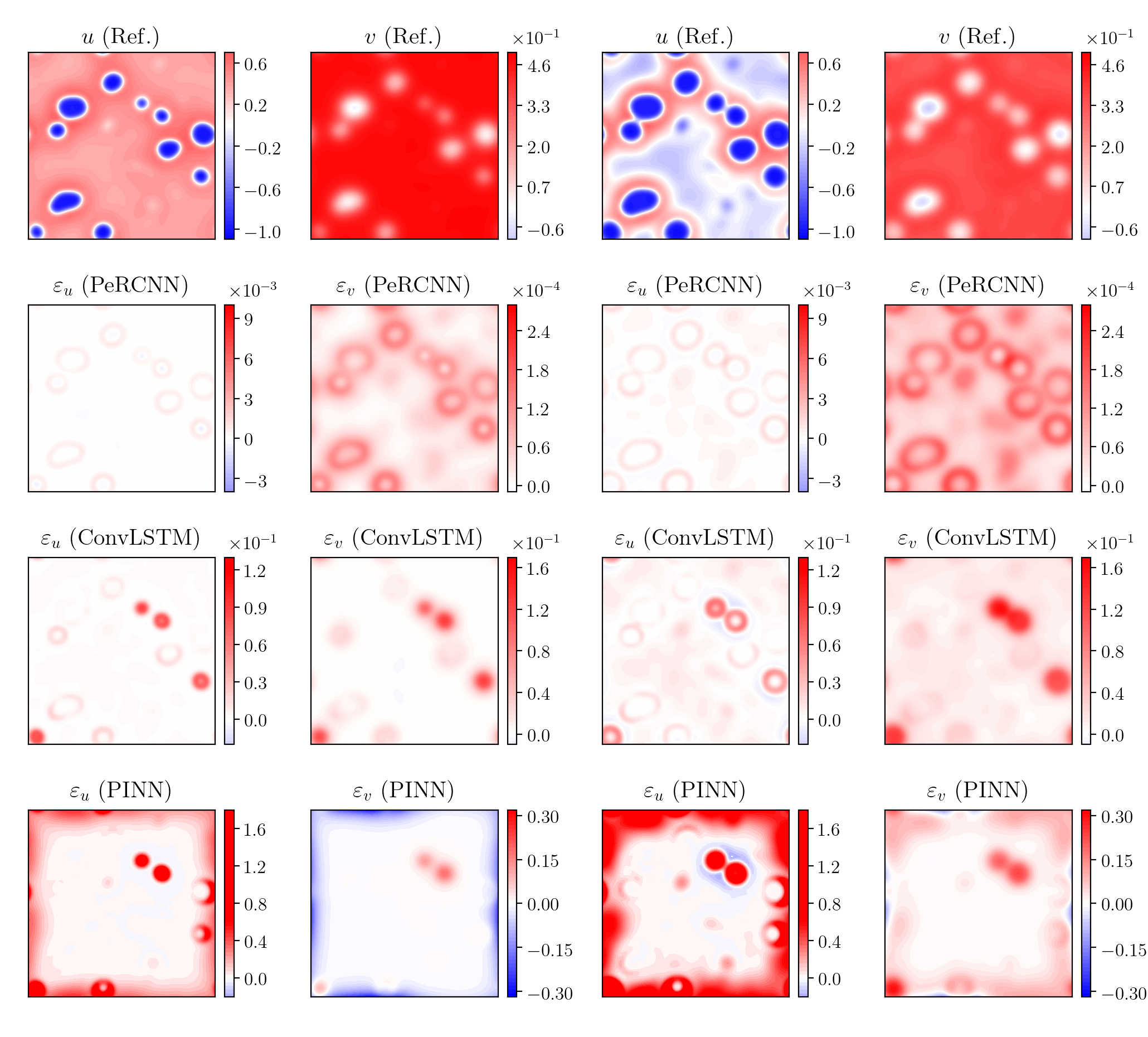}
\small
\begin{tabular*}{0.45\textwidth}{@{\extracolsep{\fill}} c c}
(a)   6 s & (b) 8 s
\end{tabular*}
\hfill
\end{center}
\vskip -0.2in
\caption{Comparison of the error against reference solution for PeRCNN, ConvLSTM and PINN prediction on 2D FitzHugh-Nagumo model.}
\label{fig:contrast_2d_FN}
\end{figure}

In this section, we compare the performance of the proposed PeRCNN on solving the reaction-diffusion (RD) systems with some existing DL-based methods, i.e., PINN \cite{raissi2019physics, raissi2019deep, raissi2020hidden, sun2020surrogate, rao2020physicsElastic, rao2020physicsTAML} and ConvLSTM \cite{shi2015convolutional}. Each method will be used to solve the aforementioned RD equations. The hyperparameters adopted in each method are selected from a range using a hold-out validation set. 

The visual comparisons are performed on the 2D $\lambda$--$\Omega$ and 2D FN RD systems. The ground truth and the corresponding error for each method are presented in Fig. \ref{fig:contrast_2d_Lam_Omg} and \ref{fig:contrast_2d_FN}. It can be seen that the PeRCNN has a much smaller error scale compared with other baselines. It is worth mentioning that the PINN features a large error near the boundaries. This is because PeRCNN and ConvLSTM could impose the BC with periodic padding, while the PINN can only resort to a penalty term in loss function. As introduced in Section \ref{sec:encoding_mechanism}, the periodic padding can impose the periodic BC forcibly by creating connectivity among boundary pairs (e.g., top and bottom boundaries). This observation shows the advantages of our method, or the discrete learning method in a broader sense, for solving the PDE of a system with regular domain. The performance difference between ConvLSTM and PeRCNN is mainly caused by the different recurrent block ($\Pi$-block vs. ConvLSTM cell). As the multiplicative form of $\Pi$-block renders the PeRCNN with better expressiveness for nonlinear terms (see Section \ref{sec:architecture}) compared with the black-box deep network, it is no surprise to see our PeRCNN outperforms the ConvLSTM regarding the prediction accuracy. 

To make a fair comparison, we also modify the ConvLSTM by incorporating a highway diffusion Conv layer into the recurrent block. As this revision increases the computational overhead of the network, we are only able to solve the 2D problems with the modified ConvLSTM. The quantitative metrics in Table \ref{tb:comparison_pde_solve} shows that the diffusion layer could improve the accuracy of the ConvLSTM. However, there still exists a accuracy gap between ConvLSTM and our PeRCNN model. 

\begin{table*}[h!]
\centering
\caption{Rooted mean square error (RMSE) of predictions against ground truth. Results of 3D ConvLSTM with diffusion Conv layer is not available due to large memory demand.}
    \begin{tabular}{lcccc}
    \toprule
    \multirow{2}{*}{System} & \multirow{2}{*}{PeRCNN} & \multirow{2}{*}{PINN} & \multicolumn{2}{c}{ConvLSTM}     \\
    \cmidrule(lr){4-5}
     &   &  & w/o diffusion & w/ diffusion \\
    \midrule
    2D $\lambda$--$\Omega$ & $\bf{2.7\times10^{-8}}$ & $2.1\times10^{-1}$ & $2.9\times10^{-3}$  & $2.2\times10^{-2}$ \\
    2D FN                  & $\bf{1.2\times10^{-4}}$ & $1.3\times10^{-1}$ & $5.1\times10^{-2}$ & $1.7\times10^{-2}$\\
    3D FN                  & $\bf{5.1\times10^{-4}}$ & $5.9\times10^{-3}$ & $8.7\times10^{-3}$ & N/A \\
    3D GS                  & $\bf{4.1\times10^{-3}}$  &  $6.7\times10^{-2}$  & $1.3\times10^{-1}$ & N/A \\
    \bottomrule
    \end{tabular}
\label{tb:comparison_pde_solve} 
\end{table*}

\section{Identification of coefficients in reaction-diffusion system} \label{sec:scalar_id}

\subsection{Problem description}\label{sec:id}

{\color{black} Here, we firstly} focus on the problem of identifying the scalar coefficients in the PDEs given some low-resolution (LR) and noisy snapshots (or measurement data) of the system. The dynamical system we consider is the 2D GS RD system governed by 
\begin{equation} 
\label{eq:gs_eqn_unknow_coef} 
\begin{aligned}
u_t&=\mu_u\Delta u-c_1\cdot uv^2+c_F(1-u) \\
v_t&=\mu_v\Delta v+c_2\cdot uv^2-(c_F+c_\kappa)v
\end{aligned}
\end{equation}
where the $\mu_u, \mu_v, c_1, c_2, c_F$ and $c_\kappa$ are unknown coefficients we seek to identify. 

\begin{figure}[t!]
\centering
\begin{minipage}[t]{1\linewidth}
\centering
\includegraphics[width=\linewidth]{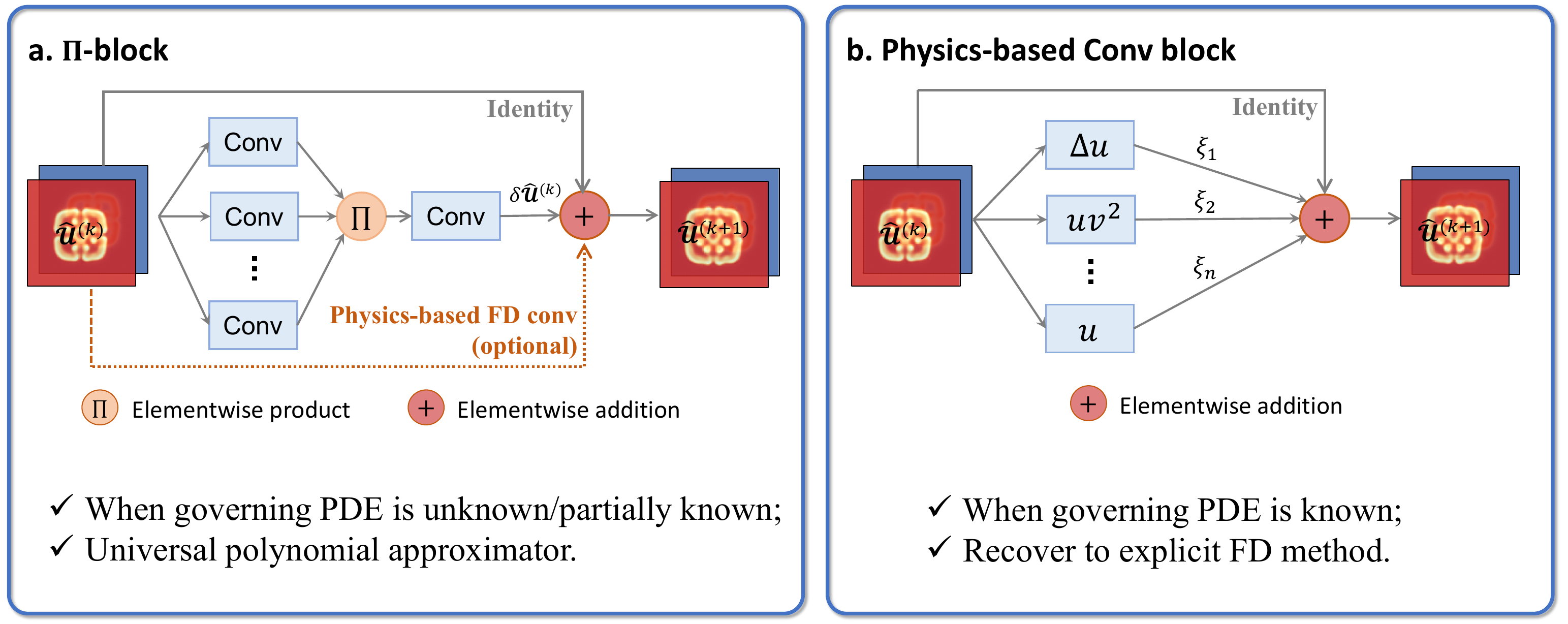}
\end{minipage}%
\caption{The comparison of the $\Pi$-block and completely physics-based block. We could switch between these two learning block depending on the availability of the prior knowledge on governing PDEs.}
\label{fig:two_form_recurrent_block}
\end{figure}

The measurement data used in the identification is synthetized through numerical simulation. We first obtain the numerical solution of the 2D GS RD system using the high-order FD solver. The physical domain of $\Omega\times\mathcal{T}=[-0.5, 0.5]^2$ $\times[0,1000]$ is discretized by $\delta x=0.01$ and $\delta t=0.5$ for the simulation. The synthetic measurement data is obtained by downsampling the numerical solution in both spatial and temporal dimension. The major assumption in this problem is that the explicit form of the governing PDE is known except some scalar coefficients. That being said, the existing terms in the PDE are all available. Therefore, we replace the $\Pi$-block in PeRCNN (see \textcolor{blue}{Main Text Fig. 1\textbf{a}}) with a completed physics-based recurrent network\footnote[1]{The recurrent block is built completely according to the RHS of Eq. \eref{eq:gs_eqn_unknow_coef}}, which is well illustrated by Fig. \ref{fig:two_form_recurrent_block}. In the constructed PeRCNN model, each unknown coefficient is represented by a individual trainable variable within the computational graph for the forward/backward computations. No elementwise product operation is involved in PeRCNNs used in this section since the form of PDE is completely known. As this physics-based Conv block computes the residual of the state variable, the whole recurrent network recovers to the explict FD method. In following subsections, we consider two different scenarios which reflects the trade-off between the spatial and temporal resolution of the measurement. 

\subsection{Scenario 1: with multiple low-resolution snapshots}\label{sec:senario1}
In this numerical example, we consider the scenario in which the measurement data is LR spatially while abundant in temporal dimension. Specifically, the available measurement consists of multiple noisy and LR snapshots of the 2D GS RD system. The ISG introduced in Section \ref{sec:architecture} is employed to generate the initial state for PeRCNN from one LR snapshot. We pretrain the ISG with the interpolated HR data. As we mainly focus on the problem with regular physical domain, the solution to the system can be discretized on a $H\times W$ Cartesian grid at time steps $\{t_1,..., t_k, ..., t_{n_t}\}$, where $n_t$ denotes the total number of time steps. The misfit error between the prediction and the available measurement data is adopted as the loss function for training the whole PeRCNN, as follows
\begin{equation} 
    \label{sys_id_loss} 
    \displaystyle
    \mathcal{J}(\boldsymbol{\theta})=\textrm{MSE}\left(\boldsymbol{\widehat{\mathcal{U}}}(\tilde{\mathbf{x}})-\tilde{\mathbf{u}}\right)+\lambda\cdot \textrm{MSE}\left(\boldsymbol{\widehat{\mathcal{U}}}^{(0)}-\mathcal{P}(\tilde{\mathbf{u}}^{(0)})\right)
\end{equation}
where $\boldsymbol{\theta}=\{\mu_u, \mu_v, c_1, c_2, c_F,c_\kappa\}$ denotes the set of trainable variables in the network; $\boldsymbol{\widehat{\mathcal{U}}}(\tilde{\mathbf{x}})$ denotes the mapping of HR prediction $\boldsymbol{\widehat{\mathcal{U}}}\in \mathbb{R}^{n_t\times n\times H \times W}$ on the coarse grid whose locations are denoted by $\mathbf{\Tilde{x}}$; $\tilde{\mathbf{u}}\in\mathbb{R}^{n_t'\times2\times H'\times W'}$ denotes the LR measurement where $n_t'<n_t$, $H'<H$ and $W'<W$; $\mathcal{P}(\cdot)$ is a spatial interpolation function (e.g., bicubic or bilinear); $\lambda$ is the regularizer weighting. The regularization term denotes the IC discrepancy between the interpolated HR initial state $\mathcal{P}(\tilde{\mathbf{u}}^{(0)})$ and the predicted HR initial state $\widehat{\mathcal{U}}^{(0)}$ from ISG, which is found effective in preventing network overfitting. Compared with the existing work on physics-informed learning \cite{raissi2019physics, raissi2019deep, rao2020physicsElastic, gao2021phygeonet}, one major distinction of the loss function employed here is the absence of the physics loss. This because the prior physics knowledge is already encoded into the network architecture as shown in Section \ref{sec:encoding_mechanism}. This facilitates the learning process of the spatiotemporal system significantly.


The synthetic measurement used for the identification includes 26 LR snapshots from $t=450$ to $t=650$. Each measurement snapshot is downsampled ($4\times$ in both spatial directions) from the HR numerical solution after being added 10\% uncorrelated Gaussian noise. The snapshots of the measurement are shown in \textcolor{blue}{Main Text Fig. 3\textbf{a}}, which is characterized with evident background noise and discontinuity. To prevent the overfitting to noise, early stopping is employed by splitting the dataset to 21 and 5 snapshots as training and validation set. The weighing coefficient for regularizer $\lambda$ is set to be $2.5\times 10^{-2}$ after numerical tests. Bicubic interpolation function is employed for the pretraining of the ISG. The final training process for the whole network consists of 40,000 iterations of Adam optimizer. The learning rate, which decays by 0.98 for every 400 iterations, is set to be $2\times10^{-3}$ at the start to achieve the best convergence after empirical study. 

As for the initialization of each coefficient, we randomly and independently sample from a physically feasible interval. Prior knowledge on the RD system is utilized as well for setting the intervals. For example, it is well studied that in GS RD system, the kill rate $c_\kappa$ and feed rate $c_F$ are bounded to ensure the stability of the solution \cite{BROWN19951713, pearson1993complex, mcgough2004pattern}. Therefore, referring to \cite{pearson1993complex,mcgough2004pattern}, we randomly draw the initial guess of $c_\kappa$ and $c_F$ from the intervals of $(0, 0.25)$ and $(0, 0.07)$ respectively, while the reaction coefficient $c_1$ and $c_2$ are from $(0, 2)$ respectively. To give a rough estimate for diffusion coefficients, we simplify the equation to $\mathbf{u}_t=\mathbf{D}\Delta\mathbf{u}$ where $\mathbf{D}=\text{diag}(\mu_u,\mu_v)$ and apply the least square method to estimate the coefficient matrix $\widehat{\mathbf{D}}$ with the LR measurement. A lower bound $0$ and upper-bound $2\widehat{\mu}$ (where $\widehat{\mu}$ is the estimated diffusion coefficient) is imposed during optimization to ensure the stability of the diffusion system. In this example, the $\widehat{\mu}_u$ and $\widehat{\mu}_v$ are $1.73\times10^{-5}$ and $5.70\times10^{-6}$ respectively.

To demonstrate the capability of the proposed method, we conducted the identification with both noisy and clean data. For each case, we perform 10 runs with various random seed for coefficient initialization. The coefficients identified by the PeRCNN, as well as the ground truth values, are presented in Table \ref{tb:coef_id}. It can be seen that for both this two cases, the PeRCNN is able to identify the unknown coefficients very well. In the absence of the noise, the identified coefficients feature high accuracy, with the mean absolute relative error (MARE) for all the coefficients being 0.6\%. For the case with 10\% Gaussian noise, the MARE is 1.61\% in spite of 10\% Gaussian noise in the measurement. 

\begin{table*}[t!]
\centering
\caption{Summary of the coefficient identification results for 2D GS reaction-diffusion system. The measurement data of $\tilde{\mathbf{u}}\in\mathbb{R}^{26\times2\times26\times26}$ is used in Scenario 1 while $\tilde{\mathbf{u}}\in\mathbb{R}^{2\times2\times51\times51}$ is used in Scenario 2.}
\resizebox{\textwidth}{!}{%
\begin{tabular}{lcccccccc}
\toprule
Data amount & Noise (\%) & $\mu_u (10^{-5})$ & $\mu_v (10^{-6})$ & $c_1$ & $c_2$ & $c_F (10^{-2})$ & $c_\kappa (10^{-2})$ & MARE (\%)   \\
\midrule
Ground truth & -& $2.0$ & $5.0$ & 1.0 & 1.0 & $4.0$ & $6.0$ & - \\
\cmidrule{1-9}
\multirow{2}{*}{Scenario 1} & 0  & $1.987$ & $4.989$ & $0.9920$ & $0.9941$ & $3.970$ & $5.965$ & $0.60$ \\
  & 10 & $1.950$ & $5.010$ & $0.9724$ & $0.9823$ & $3.938$ & $5.941$ & $1.61$ \\
\cmidrule{1-9}
\multirow{2}{*}{Scenario 2} & 0  & $1.981$ & $5.124$ & $0.9886$ & $0.9993$ & $4.014$ & $6.046$ & $0.96$ \\
 & 10 & $1.964$ & $5.111$ & $0.9864$ & $0.9987$ & $4.003$ & $6.044$ & $1.05$ \\
\bottomrule
\end{tabular}}
\label{tb:coef_id} 
\end{table*}

\subsection{Scenario 2: with initial and final high-resolution snapshots}


To further examine the capability of the proposed method, we consider a different scenario in which the first and last snapshots with decent resolution are available. The 2D GS RD system considered here has the same settings as the previous section. However, the training dataset only contains two snapshots (with $10\%$ Gaussian noise and $51\times51$ resolution) at $t=450$ and $650$, which are shown in \textcolor{blue}{Main Text Fig. 3\textbf{b}}. It can be seen that $51\times51$ snapshots reveal much more details on spatial pattern compared with the measurement in previous section. To ensure the pixel-wise supervision, the noisy snapshots are interpolated into full resolution (i.e. $101\times101$) with bicubic interpolation. The loss function used for training is 
\begin{equation} 
\label{sys_id_loss_two} 
\mathcal{J}(\boldsymbol{\theta})=\sum_{i\in\{1,n_t\}}\textrm{MSE}\left(\boldsymbol{\widehat{\mathcal{U}}}^{(i)}-\mathcal{P}(\tilde{\mathbf{u}}^{(i)})\right)
\end{equation}
where $\boldsymbol{\widehat{\mathcal{U}}}$ is the network's prediction, $\tilde{\mathbf{u}}$ is the LR measurement and the subscript (i.e., $1$ and $n_t$) indicates the first and last snapshot. We use the same coefficient initialization and training strategy as in previous section. The identified coefficients are summarized in Table \ref{tb:coef_id}. It can be seen that the network is able to identify the unknown scalar coefficients with accuracy. The MARE of the identified coefficients obtained in the noise-free and 10\% noise case is 0.96\% and 1.05\% respectively.

Although the proposed method uncovers the coefficients accurately in this scenario, we should note that for lots of dynamic systems (e.g., those with second order time derivatives $\mathbf{u}_{tt}$), the unknown coefficients may not be identifiable given merely two snapshots. The rich pattern and the mathematical properties (i.e., diffusion equation has only 1st order time derivative) of the current system make the identification possible.

{\color{black}
\subsection{Comparison with PINN}
Physics-informed neural network (PINN) has been proven to be an effective approach for inverse analysis of PDE systems, e.g., identifying the coefficients of PDEs \cite{raissi2019physics, raissi2019deep, jagtap2020conservative, yang2021b}. In this subsection, we compare the performance of our PeRCNN approach with PINN on inverse analysis of the 2D GS RD system (i.e., Scenario 1 with 10\% Gaussian noise). We employ PINN with 5 fully-connected hidden layers, each with 80 neurons, to approximate the solution. 17,576 points are collected from 26 snapshots (each of $26\times26$ resolution) as the noisy measurement data in the network training. 676 points from the first snapshot are used to enforce the initial condition and 5,000 points are sampled on each of the four edges to enforce the periodic boundary condition (PBC). In addition, we sample 100,000 collocation points within the spatiotemporal domain to calculate the PDE residual. To summarize, the loss function for training PINN is formulated as
\begin{equation}
    \label{eq:loss_func}
    \mathcal{L}(\mathcal{N}, \mathcal{C}) = \mathcal{L}_{pde} + \alpha_1 \cdot \mathcal{L}_{ic} + \alpha_2\cdot \mathcal{L}_{pbc} + \alpha_3 \cdot\mathcal{L}_{mea},
\end{equation}
where $\mathcal{L}_{pde}$, $\mathcal{L}_{ic}$, $\mathcal{L}_{pbc}$ and $\mathcal{L}_{mea}$ denote the loss component of the governing PDE, initial condition, PBC and measurement data, respectively. $\alpha_i$ denotes the weighting coefficient chosen carefully from numerical tests to ensure each component is balanced in scale. Specifically, $\alpha_1$, $\alpha_2$ and $\alpha_3$ are selected to be 5, 2 and 10, respectively. Note that the trainable variables of PINN include both the set of variables in the neural network denoted by $\mathcal{N}$ and the set of unknown coefficients denoted by $\mathcal{C}$. Table \ref{tb:comparision_pinn} provides the coefficients identified by both PINN and our approach (PeRCNN). It can be seen that the PINN struggles to recover the unknown coefficient. This is because (1) the point-wise solution approximation by the fully-connected neural network makes PINN confronted with challenge in modeling very complex patterns of the system, and (2) the sparsity (i.e., $26\times26$ resolution) and high-level noise (i.e., 10\%) of the measurement data further leads to poor prediction by PINN.}

\begin{table*}[t!]\color{black}
\centering
\caption{\hsedit{Summary of the identified coefficients by PINN and PeRCNN on the 2D GS RD system. The PeRCNN result is taken from Section \ref{sec:senario1}.} }
{%
\begin{tabular}{cccccccc}
\toprule
Parameters & $\mu_u (10^{-5})$ & $\mu_v (10^{-6})$ & $c_1$ & $c_2$ & $c_F (10^{-2})$ & $c_\kappa (10^{-2})$ & MARE (\%)   \\
\midrule
Truth & $2.0$ & $5.0$ & 1.0 & 1.0 & $4.0$ & $6.0$ & - \\
\cmidrule{1-8}
PeRCNN  & $1.950$ & $5.010$ & $0.9724$ & $0.9823$ & $3.938$ & $5.941$ & $1.61$\\
\cmidrule{1-8}
PINN &  $0.056$ & $0.054$ & $0.6760$ & $0.8652$ & $2.871$ & $6.109$ & $43.71$ \\
\bottomrule
\end{tabular}}
\label{tb:comparision_pinn} 
\end{table*}

{
\color{black}
\subsection{Identification of space-varying coefficients}

We further consider the problem of identifying non-scalar coefficients in a PDE system. For example, some real-world systems are accompanied by space-varying coefficients (e.g., nonlinear reaction process in a nonhomogeneous media). Fortunately, our approach can be easily extended to solve such problems by introducing a trainable approximator to the space-varying coefficient in the PDE system. Without loss of generality, we utilize the Gaussian mixture to approximate the space-varying coefficient, given by
\begin{equation} 
    \label{eq:gaussian_mixture} 
    \mathcal{G} (\mathbf{x}|\{A_{1:N},\boldsymbol{\mu}_{1:N}, \boldsymbol{\Sigma}_{1:N}\}) = \sum_{i=1}^N \frac{A_i}{2\pi|\boldsymbol{\Sigma}_i|^{1/2} } \exp \left\{- \frac{1}{2} \left(\mathbf{x}-\boldsymbol{\mu}_i\right)^{T} \boldsymbol{\Sigma}_i^{-1} \left(\mathbf{x}-\boldsymbol{\mu}_i \right)  \right\}
\end{equation}
where $A_i$, $\boldsymbol{\mu}_i$ and $\boldsymbol{\Sigma}_i$ are the amplitude, mean vector and covariance metrics of the $i$ th component. These trainable parameters in the Gaussian mixture are optimized by minimizing the loss function (see Eq. \eref{sys_id_loss}).

\begin{figure}[t!]
\centering
\includegraphics[width=0.8\linewidth]{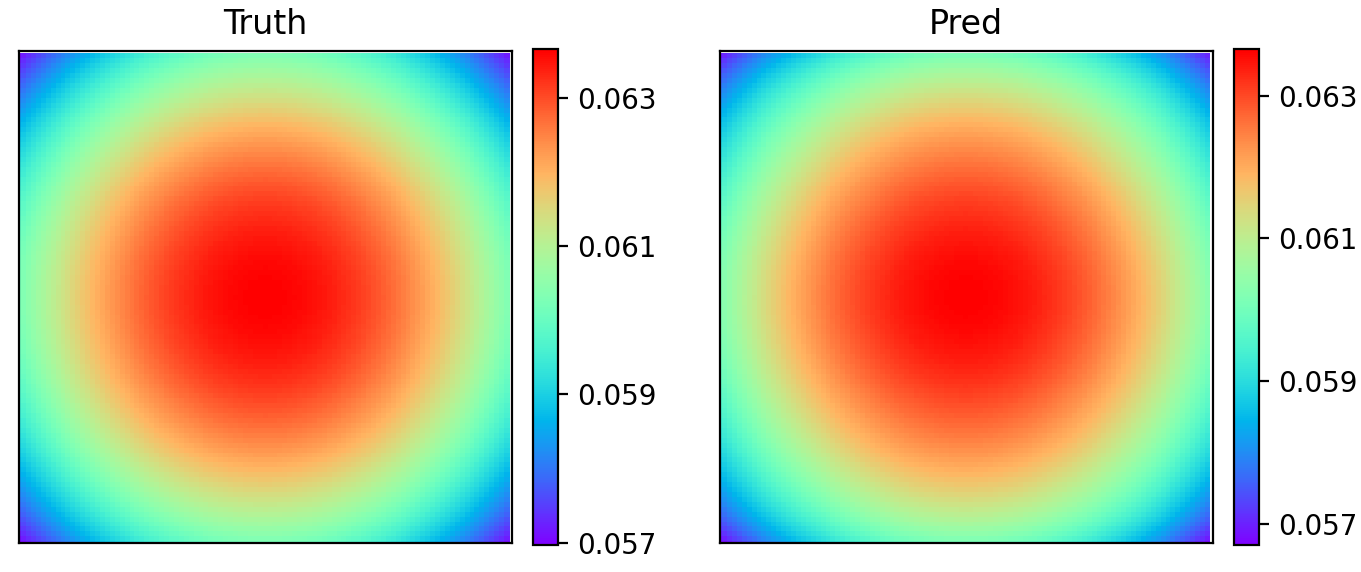}
\caption{\hsedit{\color{black}Contour of the space-varying coefficient $c_\kappa(x,y)$. Left: ground truth, right: prediction.} }
\label{fig:space_varying_coefficient}
\end{figure}

To demonstrate the effectiveness of this extension, we consider the 2D Gray-Scott reaction-diffusion equation
\begin{equation} 
\label{eq:2d_gs_varying_coef} 
\begin{aligned}
u_t&=\mu_u\Delta u-c_1\cdot uv^2+c_F(1-u) \\
v_t&=\mu_v\Delta v+c_2\cdot uv^2-(c_F+c_\kappa)v
\end{aligned}
\end{equation}
where the $\mu_u, \mu_v, c_1, c_2$ and $c_F$ are scalar coefficients, while $c_\kappa(x,y)$ is a space-dependent function, namely,
\begin{equation} 
c_\kappa(x,y)=A\exp \left ( -\frac{x^2+y^2}{2} \right )
\end{equation}
Here, $A$ is set to be 0.0636 in this example. The computational settings to generate the ground truth are the same as the previous section. The measurement data used in the training includes 1500 noise-free snapshots of $100\times100$ resolution. The Gaussian mixture with 5 components is utilized in conjunction with the PeRCNN described in Section \ref{sec:id}. Note that the Gaussian mixture parameters are randomly initialized to ensure the positiveness of $A_i$ and $\boldsymbol{\Sigma}_i$. The network is trained with 20,000 epochs by the Adam optimizer. The learning rate starts at 0.01 and evolves with the decaying rate of 0.98 every 200 epochs. The result is provided in Table \ref{tb:coef_id_space_varying} and Fig. \ref{fig:space_varying_coefficient}. It can be seen that all the unknown coefficients including the space-varying $c_\kappa$ are accurately recovered by the network. It should be noted that this numerical example seeks to demonstrate the extension of the proposed approach for identifying space-varying coefficients in PDE systems. For more complex problems, one can utilize other approximation functions such as polynomial basis, Fourier basis, or neural networks to parameterize the unknown coefficient.

\begin{table*}[t!]
\centering
\caption{\color{black}The identified scalar coefficients. The identified space-varying coefficient is given in Fig. \ref{fig:space_varying_coefficient}.}
\begin{tabular}{lccccccc}
\toprule
  & $\mu_u (10^{-5})$ & $\mu_v (10^{-6})$ & $c_1$ & $c_2$ & $c_F (10^{-2})$ & MARE (\%)   \\
\midrule
Ground truth & $2.0$ & $5.0$ & 1.0 & 1.0 & $4.0$ & - \\
\cmidrule{1-7}
Identified & $2.0008$ & $5.0000$ & $1.0004$ & $1.0001$ & $4.0017$ & $0.0265$ \\
\bottomrule
\end{tabular}
\label{tb:coef_id_space_varying} 
\end{table*}

}

\section{Physics-encoded data-driven modeling of RD system}
\label{sec:data_driven_model}

\subsection{Problem description}

In this section, we employ the PeRCNN to establish a generalizable and robust data-driven model for predicting HR nonlinear spatiotemporal dynamics based on very limited LR and noisy training data. One of the biggest challenges in this problem comes from the limited prior knowledge on the system. To formulate this problem, we would again consider the spatiotemporal dynamical system introduced in Section \ref{sec:residual_pde}, which is described by Eq. \eref{eq:dynamic_system_pde}. 

Similar to the settings in Section \ref{sec:senario1}, we assume it is provided a small amount of and potentially noisy measurements over a coarser spatiotemporal grid $\Tilde{\mathbf{u}}\in\mathbb{R}^{n_t'\times2\times H'\times W'}$. The goal of the data-driven modeling is to establish a reliable model that gives the most likely full-field solution $\boldsymbol{\widehat{\mathcal{U}}}\in \mathbb{R}^{n_t\times 2\times H \times W}$, and possesses satisfactory extrapolation ability over the temporal horizon (e.g., for $t > t_{n_t}$). To this end, we employ the PeRCNN introduced in Section \ref{sec:architecture} to learn the system's dynamics from measurements data. The loss function used to train the network is defined by Eq. \eref{sys_id_loss} where $\boldsymbol{\theta}=\{\mathbf{W},\mathbf{b},\mu_u,\mu_v\}$ denotes the set of trainable parameters (i.e., network's weights $\mathbf{W}$, biases $\mathbf{b}$ and diffusion coefficients $\mu_u$ and $\mu_v$ in the highway diffusion Conv layers) in the network. 


\subsection{Synthetic measurement data}
\label{sec:syn_meas_data}

Table \ref{tb:data_set} summarizes the computational parameters of the datasets (or numerical solutions) involved in the paper. Each dataset denotes the numerical solution obtained by solving an IBVP of the corresponding PDE. The synthetic measurement data we used to establish the data-driven model is obtained through downsampling the dataset. In this section, the 2D and 3D GS RD system would be primarily studied while the 2D Burgers' equation system would also be tested as a supplementary example. The 3D GS RD system introduced in Section \ref{sec_3d_pde_solve} is reused in this section while the 2D GS RD system features the parameters of $\mu_u=2\times10^{-5},~\mu_v=5\times10^{-6},~\kappa=0.06$ and $F=0.04$. The Burgers' equation has a wide range of applications in fluid or traffic flow modeling. Its governing equation can be written as
\begin{equation}
    \label{eq:burgers}
    \begin{aligned}
    u_t&=\nu\Delta u - (uu_x + vu_y) \\
    v_t&=\nu\Delta v - (uv_x + vv_y)
    \end{aligned}
\end{equation}
where $\mathbf{u}=[u,~v]^\texttt{T}$ denotes the fluid velocities and $\nu$ is the viscosity coefficient, which is set to be 0.005 in this example.

\begin{table}[t!]
\caption{Computational parameters for datasets generation.}
\vspace{-15pt}
\label{tb:data_set}
\begin{center}
\begin{small}
\begin{tabular}{lccccc}
\toprule 
\multirow{2}{*}{Dataset} & \multirow{2}{*}{PDE} & \multirow{2}{*}{Parameters} &  Size & \multirow{2}{*}{$\delta t$} & \multirow{2}{*}{$\delta x$} \\
& & & ($n_t\times n\times H\times W$) &&\\
\midrule
2D $\lambda$-$\Omega$& \eref{eq:LO_eqn} &  $\mu_u$=0.1, $\mu_v$=0.1 $\beta$=1.0  & $801\times2\times 101^2$ & 0.0125 & 0.2 \\
2D FN & \eref{eq:fn_eqn} & $\mu_u=1.0$, $\mu_v=10.0$, $\alpha=0.01$, $\beta=0.25$  & $ 6001\times2\times101^2$ & 0.002 & 1.0 \\
3D FN & \eref{eq:fn_eqn} & $\mu_u=1.0$, $\mu_v=10.0$, $\alpha=0.01$, $\beta=0.25$  & $1001\times2\times51^3$ & 0.004 & 1.0 \\
\multirow{2}{*}{2D GS} & \multirow{2}{*}{\eref{eq:gs_eqn}} & $\mu_u=2.0\times 10^{-5}$, $\mu_v=5.0\times 10^{-6}$,  &  \multirow{2}{*}{$2501\times 2\times101^2$} &  \multirow{2}{*}{0.5}&  \multirow{2}{*}{$0.01$} \\
&&$F=0.04$, $\kappa=0.06$ &&&\\
3D GS & \eref{eq:gs_eqn} &  $\mu_u=0.2$, $\mu_v=0.1$, $F=0.025$, $\kappa=0.055$  & $1501\times2\times49^3$  & 0.5 & 25/12\\
2D Burgers' & \eref{eq:burgers} &  $\nu=0.005$  & $ 1601\times2\times101^3$  & $2.5\times10^{-4}$ & 0.01\\
\bottomrule
\end{tabular}
\end{small}
\end{center}
\end{table}

\subsection{Evaluation metrics}
\paragraph{Accumulative rooted-mean-square error (RMSE):} Accumulative RMSE is defined by 
\begin{equation} 
    \label{eq:accum_rmse} 
    \begin{aligned}
    \text{RMSE}(t_i) = \sqrt{\text{MSE} \left ( \widehat{\boldsymbol{\mathcal{U}}}^{(1:i)}-\boldsymbol{\mathcal{U}}_{\text{ref}}^{(1:i)}\right )}
    \end{aligned}
\end{equation}
where $\widehat{\boldsymbol{\mathcal{U}}}^{(1:i)}$ and $\boldsymbol{\mathcal{U}}_{\text{ref}}^{(1:i)}$ denote the predicted and reference solution from $1^\text{st}$ to $i^\text{th}$ step respectively. It measures the error of all snapshots til a time instance $t_k$. We use this metrics to evaluate the error propagation of the model's prediction. 

\paragraph{Physics error:} For the data-driven modeling of the physical system, we are also interested in the residual of the governing PDE, which reflects to what degree the prediction obeys the underlying physical law. As the prediction $\widehat{\boldsymbol{\mathcal{U}}}$ is given on a Cartesian grid, we are able to compute the physics error (see Eq. \eref{eq:loss_func_solve_pde}) based on the PDEs through FD method. The physic error would be used as a supplementary metrics to evaluate the accuracy of the model.

\subsection{Sensitivity tests on major hyperparameters}
\label{sec:data_driven_sensitivity}
The proposed PeRCNN has several hyperparameters, including the number of parallel convolutional layers, the filter size and the number of channels, which could affect the performance of the trained model. Therefore, we herein explore the sensitivity of these hyperparameters w.r.t. model performance. The 2D GS RD system (see Table \ref{tb:data_set}) is used as the testbed for the sensitivity study. The measurement data consists of 41 LR ($26\times 26$) snapshots uniformly downsampled from the time interval from $t=0$ to $200$ after 10\% Gaussian noise is added to the raw dataset. Each PeCRNN model is constructed to produce the prediction on full resolution, i.e., $\widehat{\boldsymbol{\mathcal{U}}}\in\mathbb{R}^{401\times 2\times 101\times 101}$. 

\begin{figure*}[htbp]
\centering
\subfigure[Filter size]{
\begin{minipage}[t]{0.63\linewidth}
\centering
\includegraphics[width=\linewidth]{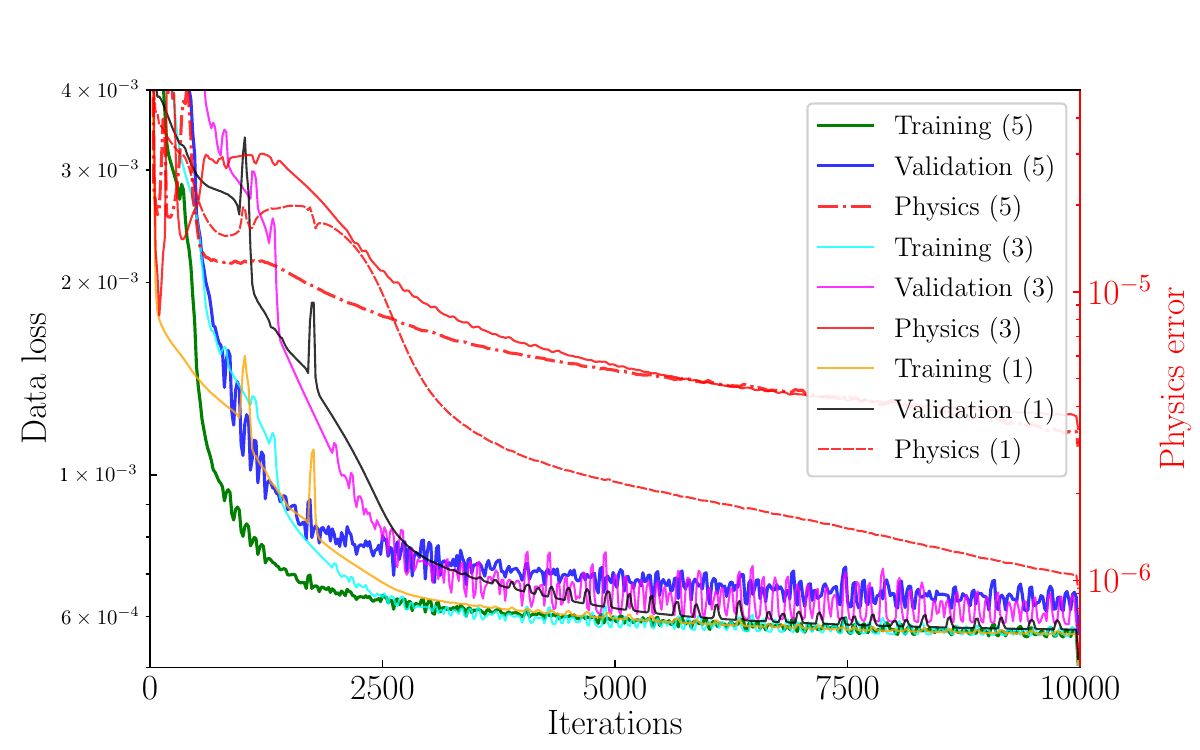}
\end{minipage}%
}%
\hfill\\
\centering
\subfigure[Number of channels. ]{
\begin{minipage}[t]{0.63\linewidth}
\centering
\includegraphics[width=\linewidth]{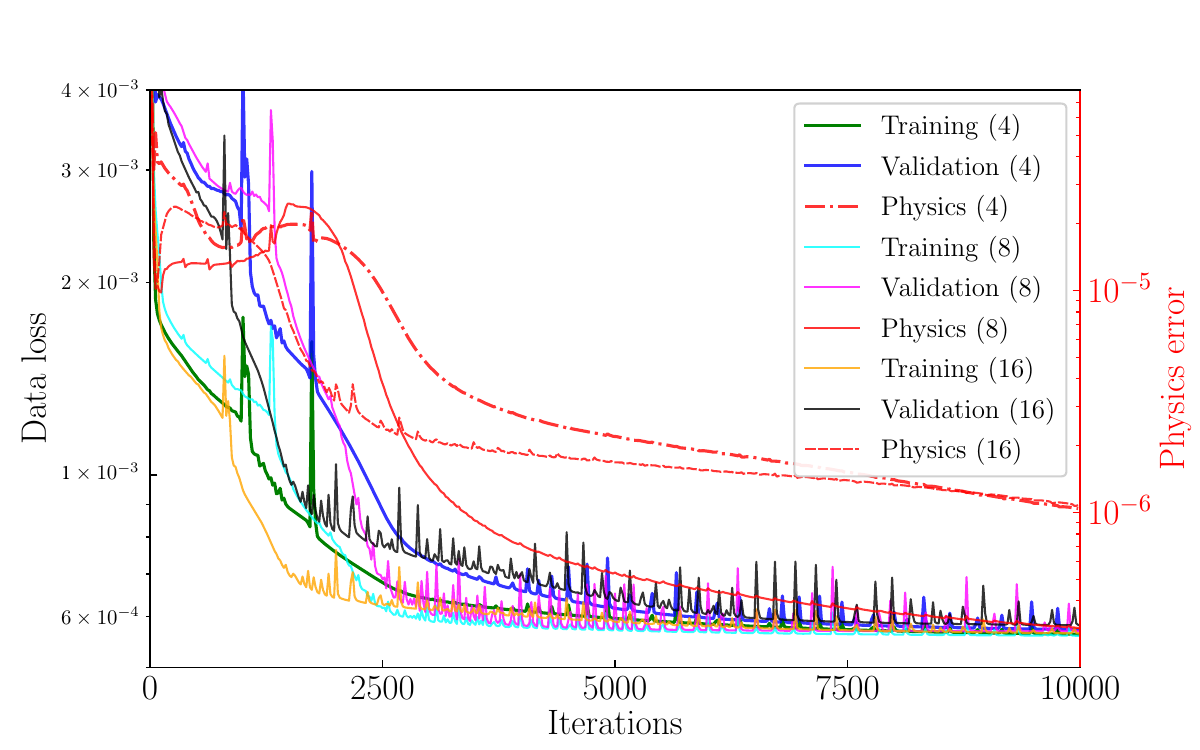}
\end{minipage}%
}%
\hfill\\
\centering
\subfigure[Number of Conv layers]{
\begin{minipage}[t]{0.63\linewidth}
\centering
\includegraphics[width=\linewidth]{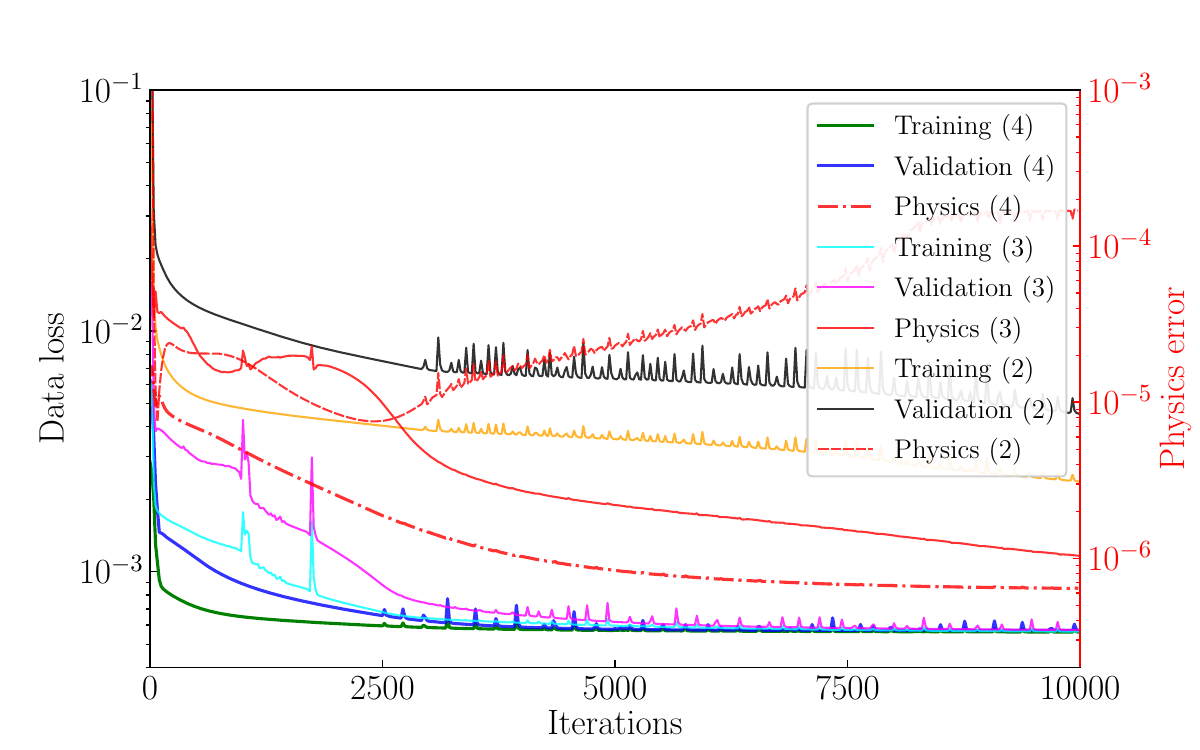}
\end{minipage}
}%
\vspace{-5pt}
\caption{Comparison of learning curves for different hyperparameter settings.}
\label{fig:learning_curve}
\end{figure*}

\paragraph{Filter size:} The close connection between the convolution operation and numerical differentiation has been well studied \cite{cai2012image}. In general, the size of the Conv filter (or stencil in FD) decides the highest order of spatial derivatives it can approximate. In this test, we compare the learning process and model accuracy for PeRCNN with various filter size, e.g., 5, 3 and 1. Apart from the filter size, each network has 3 parallel Conv layers with 4 channels in addition to the physics-based Conv layer defined by the diffusion term. Figure \ref{fig:learning_curve}(a) compares the learning curves for networks with different filter sizes. It can be observed that all networks are able to fit the data with a small training error. However, the network with $3\times 3$ filters generalizes the physics best according to the physical error. In the meanwhile, it is interesting that the network with filter size of 1 performs as good as other two networks. This is because the reaction term of GS model contains no spatial derivatives and $1\times 1$ filters would be sufficient for the representation.

\paragraph{Number of channels:} In Section \ref{sec:architecture}, we argue that the number of channels of the feature map decides how many nonlinear terms the $\Pi$-block could express. Before we configure the tests, we recognize that the reaction term of $u$ for the GS model, i.e.,  $-uv^2+F(1-u)$, would require two channels to represent $uv^2$ and $F(1-u)$ respectively. However, empirical study shows that the redundant trainable parameters could boost the convergence of the training. Therefore, we set the number of channels to be 4, 8 and 16 in the tests. Besides, the Conv layers of 3 and the filter size of 1 are adopted. The learning curves for different numbers of channels are shown in Figure \ref{fig:learning_curve}(b). It is seen that all the three settings are sufficient to ensure the network converges while the 8-channel network delivers the best data loss and physical error.

\paragraph{Number of Conv layers:} In our design of the PeRCNN architecture, more parallel Conv layers would lead to the approximation capability higher degree polynomial for the unknown function $\mathcal{F}$. However, in typical data-driven modeling, we have no prior knowledge on the highest degree of the function. Therefore, the number of Conv layers for PeRCNN will be determined through parametric tests. In this test, we explore how largely the performance of PeRCNN relies on the number of Conv layers adopted. The networks with 2, 3 and 4 layers are considered for comparison, with the filter size of 1, the channel number of 4 and the diffusion term encoded. Figure \ref{fig:learning_curve}(c) compares the learning curves for PeRCNNs with different numbers of Conv layers. It is seen that the loss for both the 3-/4-layer network converges to a same level while the 4-layer network converge faster at the beginning. By contrast, the 2-layer network is unable to fit the data due to the fact that the  reaction term in the GS model is third degree polynomial. This test shows that more parallel layers could help improve the model performance, though slightly in this case, at the cost of more trainable parameters. 

\subsection{Numerical examples}

In this part, we use the proposed PeRCNN to solve the data-driven modeling problem. Through several numerical examples, we demonstrate that our method outperforms some state-of-the-art data-driven modeling methods regarding the computational efficiency and accuracy. More importantly, PeRCNN is found to possess extraordinary capability in extrapolation (or generalization) on which the black-box models (e.g. DNN, ConvLSTM, etc.) are known to perform poorly \cite{marcus2018deep}. This claim is verified by comparing the PeRCNN with some widely used data-driven models, including the recurrent ResNet \cite{liao2016bridging, zhang2017deep}, ConvLSTM \cite{shi2015convolutional}, PDE-Net \cite{long2018pde} and deep hidden physics model (DHPM) \cite{raissi2018deep}, on multiple dimensions. 

In the comparisons, we fix the amount of measurement data, training/validation dataset splitting, the number of prediction steps, the Gaussian noise level and the random seed among all methods. Specifically, 10\% validation data is split out from the training data for hyperparameters selection and early stopping. The measurement data used for constructing the data-driven model is downsampled from the full-resolution numerical solution after being added 10\% Gaussian noise. Details of the measurement data in each example would be provided in the corresponding subsection. Once the model is finalized, extrapolation over time (i.e., inference beyond the time interval of training dataset) would be performed to examine the generalizability of each model. 

\subsubsection{Baselines}
\label{sec:data_driven_baseline}

To make a comparison of PeRCNN with other widely used models for spatiotemporal predictions, we also implemented the recurrent ResNet \cite{liao2016bridging, zhang2017deep}, ConvLSTM \cite{shi2015convolutional}, PDE-Net \cite{long2018pde} and DHPM \cite{raissi2018deep}. A very brief introduction to each method is given below for readers to grasp the major characteristics of each model.

\paragraph{ConvLSTM} \cite{shi2015convolutional} is a convolutional variant of Long-short Term Memory (LSTM) which exploits multiple self-parameterized controlling gates, such as input, forget and output gates, to capture the spatiotemporal correlations among the data. It has been extensively used in applications such as video super-resolution \cite{tao2017detail,liang2017dual}, traffic prediction \cite{yuan2018hetero} and climate forecasting \cite{shi2015convolutional}, among many others. 

\paragraph{Recurrent ResNet} is another model adopted widely by researchers \cite{ liao2016bridging, zhang2017deep} for the spatiotemporal prediction of dynamical systems. One main characteristic distinguishes the recurrent ResNet with the conventional ResNet \cite{he2016deep} is that the weights are shared across time. 

\paragraph{PDE-Net} \cite{long2018pde} is a recurrent deep network that is able to predict dynamics of complex systems. Its basic idea is to learn the differential operators from the measurement data using the convolutional filters and approximate the the unknown dynamics of the system. 

\paragraph{DHPM} \cite{raissi2018deep} differs from the previous three models as it utilizes fully connected neural networks (FCNNs) to approximate the prediction, on which hidden physics prior is exerted. In DHPM, one deep FCNN is employed to fit the measurement (or labeled) data, i.e., pairs of the spatiotemporal location $[x,y,t]$ and state variable $[u,v]$, while another shallow FCNN is used to impose a hidden physical constraint on the fitted solution. 

\vspace{15pt}

Due to the different network architecture of each method, it is impossible to adopt an exact same set of hyperparameters. Therefore, the configurations that achieves the lowest validation loss from a range of hyperparameters are adopted for the comparison, which are summarized in Table \ref{tb:hyerpara_range_percnn}--\ref{tb:hyerpara_range_dhpm}. In addition to the listed hyperparameters, all other hyperparameters are kept the same, e.g., training/validation dataset split, the number of prediction steps, the optimizer (Adam), the max number of epochs, the Gaussian noise level (10\%) and the random seed. In the network architecture design, we assume the solution within the domain is periodic while the dynamical system of interest is accompanied with the ubiquitous diffusion phenomenon. Hence, a highway diffusion Conv layer with fixed filters will be created in the PeRCNN models. 

\begin{table*}[h!]
\caption{Range of hyperparameters for PeRCNN.}
\label{tb:hyerpara_range_percnn}
\vspace{-15pt}
\begin{center}
\begin{small}
    \begin{tabular}{lcccccc}
    \toprule 
    \multirow{2}{*}{Dataset} & \multirow{2}{*}{Filter size} & \multirow{2}{*}{$\#$ layers} & $\#$ channels  & $\#$ channels & \multirow{2}{*}{Learning rate} & \multirow{2}{*}{$\lambda$}  \\
        &  &  &  ($\Pi$-block)  &(ISG) &  &  \\
    \midrule
    2D GS       & 1$\sim$5 (1) & 2$\sim$4 (3) & 4$\sim$16 (8) & 4$\sim$16 (8) & 0.001$\sim$0.01 (0.002) & 0.001$\sim$1 (0.005)\\
    3D GS       & 1$\sim$5 (1) & 2$\sim$4 (3) & 2$\sim$8 (4) & 4$\sim$8 (4) & 0.001$\sim$0.01 (0.005) & 0.001$\sim$1 (0.5) \\
    \bottomrule
    \end{tabular}
\end{small}
\end{center}
\vskip -0.1in
\end{table*}

\begin{table*}[h!]
\caption{Range of hyperparameters for ConvLSTM.}
\label{tb:hyerpara_range_convlstm}
\vspace{-15pt}
\begin{center}
\begin{small}
\begin{tabular}{lcccccc}
\toprule 
Dataset & Filter size & $\#$ layers & $\#$ channels & Learning rate & Weight decay   \\
\midrule
2D GS & 3$\sim$5 (5) & 1$\sim$2 (2) & 16$\sim$32 (32) & 0.0005$\sim$0.01 (0.001) & e-5$\sim$e-3 (e-5) \\
3D GS & 3$\sim$5 (5)  & 1   & 8$\sim$16 (16) & 0.0005$\sim$0.01 (0.0005) & e-5$\sim$e-3 (e-5)  \\
\bottomrule
\end{tabular}
\end{small}
\end{center}
\vskip -0.1in
\end{table*}

\begin{table*}[h!]
\caption{Range of hyperparameters for recurrent ResNet.}
\label{tb:hyerpara_range_resnet}
\vspace{-15pt}
\begin{center}
\begin{small}
\begin{tabular}{lcccccc}
\toprule 
Dataset & Filter size & $\#$ layers & $\#$ channels & Learning rate & Weight decay   \\
\midrule
2D GS & 3$\sim$5 (3) & 2$\sim$4 (2) & 16$\sim$128 (64) & 0.0001$\sim$0.01 (0.0002) & e-5$\sim$e-2 (e-3) \\
3D GS & 3$\sim$5 (3)  & 2$\sim$3 (2)  & 8$\sim$32 (32) & 0.0001$\sim$0.01 (0.0002) & e-5$\sim$e-2 (e-4)  \\
\bottomrule
\end{tabular}
\end{small}
\end{center}
\vskip -0.1in
\end{table*}

\begin{table*}[h!]
\caption{Range of hyperparameters for PDE-Net.}
\label{tb:hyerpara_range_pdenet}
\vspace{-15pt}
\begin{center}
\begin{small}
\begin{tabular}{lccccc}
\toprule 
Dataset & Filter size & $\#$ channels & Learning rate & Weight decay   \\
\midrule
2D GS & 3$\sim$5 (5) & 5$\sim$15 (15) & 0.0001$\sim$0.01 (0.0002) & e-5$\sim$e-2 (e-5) \\
3D GS & 3$\sim$5 (3) & 5$\sim$15 (15) & 0.0001$\sim$0.01 (0.001) & e-5$\sim$e-2 (e-4)  \\
\bottomrule
\end{tabular}
\end{small}
\end{center}
\vskip -0.1in
\end{table*}

\begin{table*}[h!]
\caption{Range of hyperparameters for DHPM.}
\label{tb:hyerpara_range_dhpm}
\vspace{-15pt}
\begin{center}
\begin{small}
    \begin{tabular}{lcccccc}
    \toprule 
    Dataset & $\mathcal{N}_1$ width & $\mathcal{N}_1$ depth & $\mathcal{N}_2$ width & $\mathcal{N}_2$ depth & Input for $\mathcal{N}_2$ & Learning rate    \\
    \midrule
    2D GS & 60$\sim$100 (80)  & 4$\sim$5 (5) & 10$\sim$30 (10) & 2$\sim$3 (2)& $(\Delta u,\Delta v,u,v)$ & 0.001 $\sim$0.02 (0.01)  \\
    3D GS & 60$\sim$100 (80)  & 4$\sim$5 (5) & 10$\sim$30 (10) & 2$\sim$3 (2)& $(\Delta u,\Delta v,u,v)$ & 0.001 $\sim$0.02 (0.01)  \\
    \bottomrule
    \end{tabular}
\end{small}
\end{center}
\end{table*}

\subsubsection{2D GS RD system}\label{sec:data_driven_2dgs}

We first perform the data-driven modeling on 2D GS RD system, which exhibits complicated maze-like pattern. The problem setups, computational parameters and discretization to obtain the synthetic measurement data are given in Table \ref{tb:data_set}. The measurement data employed for training the data-driven model includes 41 LR snapshots ($26\times26$) from $t=0$ to $400$ downsampled from the raw data. In the meanwhile, 10\% Gaussian noise is added into the measurement data. We assume that no prior knowledge on system is available except the existing diffusion phenomenon from the measurement. Therefore, we create a highway diffusion layer in PeRCNN to encode the $\Delta\mathbf{u}$ term. To accelerate the training process and ensure the stability of diffusion system, we impose a lower bound of $0$ and upper bound of $2\Tilde{\mu}$ on the coefficient in the diffusion layer, where $\Tilde{\mu}$ is the estimated diffusion coefficient from solving a linear regression problem of $u_t=\Tilde{\mu}\Delta u$ with the measurement data. Specifically, the estimated diffusion coefficient is $3.01\times10^{-5}$. Each model is trained for predicting 801 full-resolution snapshots $\widehat{\boldsymbol{\mathcal{U}}}\in\mathbb{R}^{801\times2\times  101\times 101}$ during the training phase while 1700 extra snapshots are inferred for extrapolation once the model is finalized. The hyperparameters adopted by each model are provided in Section \ref{sec:data_driven_baseline}.
 
\begin{figure}[t!]
\centering
\includegraphics[width=0.99\textwidth]{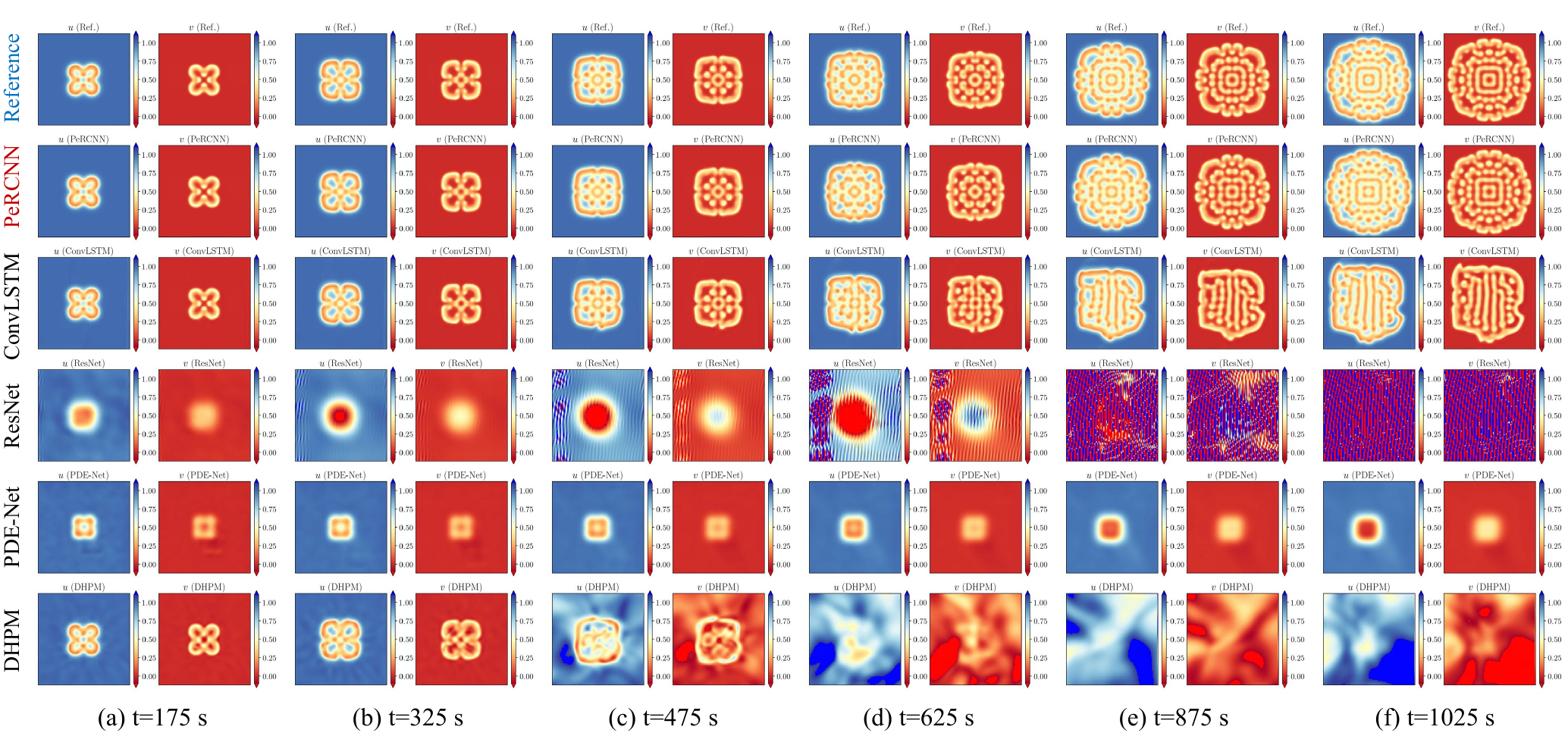}
\caption{Comparison of the predictions among multiple models. The training measurement has the resolution of $26\times26$ and is corrupted by 10\% Gaussian noise. The first four columns ($t=175,~325$) fall into the interval with measurement while the remainder ($t=475,~625,~875$ and $1025$) corresponds to extrapolation results.}
\label{fig:contrast_2d_GS_data_driven}
\end{figure}

Predictions at various times are presented in Fig. \ref{fig:contrast_2d_GS_data_driven}. We report that the recurrent ResNet is unable to reconstruct the full-resolution snapshots of the training data due to the limited data after all the hyperparameter combinations are tried. Apart from the recurrent ResNet, all the models are able to fit the training data and give satisfactory prediction in the supervised time period. However, for the prediction beyond the time interval of the training data (or extrapolated prediction), the accuracy of all models except PeRCNN deteriorates sharply as time goes. It shows that PeRCNN generalizes the unknown underlying physics well from the data. To quantitatively measure the extrapolation capability of our model, we also plot the evolution of accumulative RMSE defined by Eq. \eref{eq:accum_rmse} in \textcolor{blue}{Main Text Fig. 4\textbf{a}}. It can been seen that the PeRCNN outperforms the competitors at all stages regarding the error propagation, which further confirms the generalizability of the PeRCNN. We may notice that the accumulative RMSE starts from an initial high value. This is due to the fact that the training data is corrupted by 10\% Gaussian noise and the metrics is computed from one single snapshot at the beginning. The effect of the unrelated noise gradually fades out as more time steps are considered.

\subsubsection{3D GS RD system}\label{sec:data_driven_3d_rd}
\begin{figure}[t!]
\centering
\includegraphics[width=0.99\textwidth]{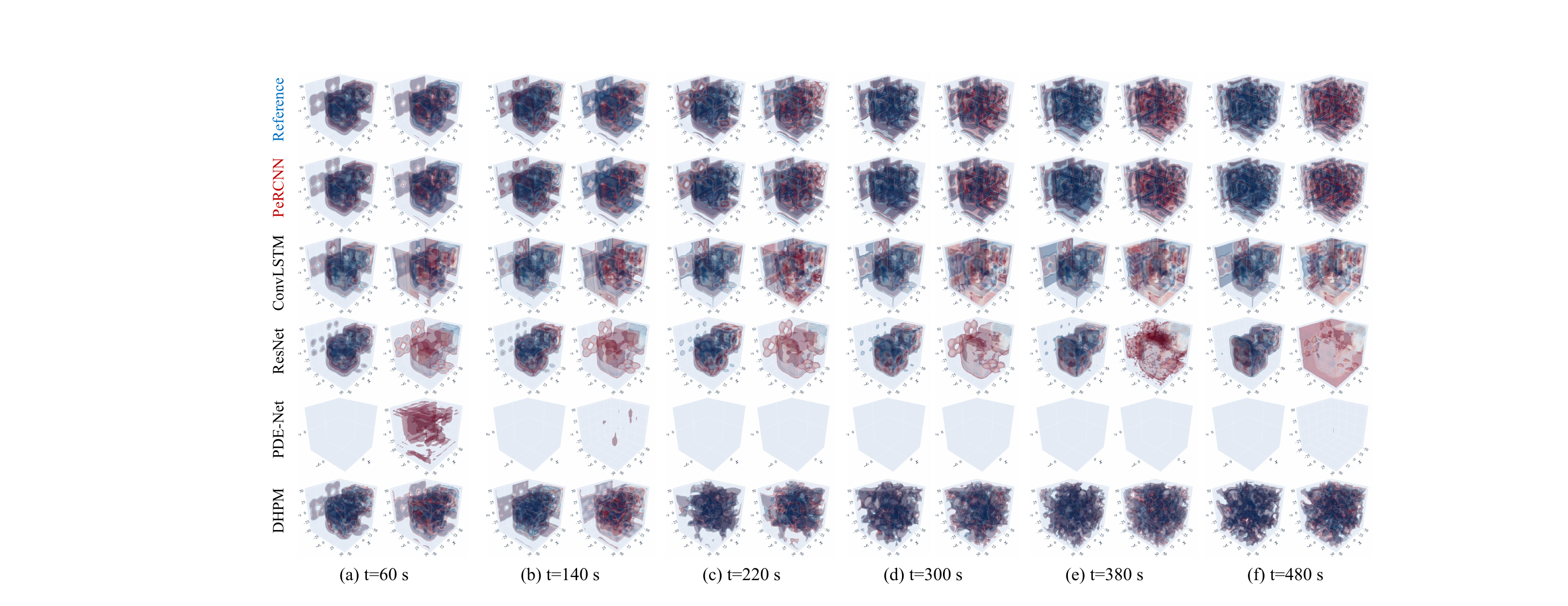}
\caption{Comparison of the predictions on 3D GS RD system. The training measurement has the resolution of $25\times25\times25$ and is corrupted by 10\% Gaussian noise. The first four columns ($t=60,~140$) fall into the interval with measurement while the remainder ($t=220,~300,~380$ and $480$) corresponds to extrapolation results.}
\label{contrast_3d_GS_data_driven}
\end{figure}

In the last example, we test our method on 3D GS RD system. The physical domain ($\Omega\times\mathcal{T}=[-50, 50]^3$ $\times[0,500]$) and the system parameters ($\mu_u=0.2$, $\mu_v=0.1$, $F=0.025$ and $\kappa=0.055$) are the same as in Section \ref{sec_3d_pde_solve}. As the computational intensity of this higher dimensional example brings challenges to the existing methods, we aim to scrutinize the performance of our PeRCNN regarding the scalability and computational efficiency. The training data used to establish the data-driven model includes 21 noisy LR snapshots ($25^3$) sampled from $t=0$ to $150$. The assumptions on the system and the estimate of the diffusion coefficients in the previous 2D GS RD example are applied here as well. Each trained model produces 301 full-resolution ($49^3$) snapshots during learning stage while 700 extrapolation steps are predicted after each model is finalized. The predicted isosurfaces of two levels are plotted in Fig. \ref{contrast_3d_GS_data_driven}. The conclusion we draw in the previous examples applies here as well since the PeRCNN is the only model that gives a satisfactory long-term prediction. The flat error propagation curve of PeRCNN, as shown in \textcolor{blue}{Main Text Fig. 4\textbf{b}}, also demonstrates the remarkable generalization capability of PeRCNN. 

Table \ref{tb:compare} compares the number of trainable parameters, the training time per epoch, and the RMSE of both training and extrapolation for each model. It can be seen that PeRCNN is characterized with good model efficiency as it uses least amount of training parameters. For the 3D case where the training efficiency of the network is of great concern, the elapsed time for training one epoch by PeRCNN is comparable to that of the ResNet, which is widely acknowledged to be an efficient network architecture. It should be noted that the DHPM achieves the lowest time cost for training a epoch. This is because no recurrent computation is involved in DHPM as it relies on FCNN for learning the dynamics. As for the accuracy of the training and extrapolation, our model outperforms the baselines consistently across different examples. In a nutshell, PeRCNN outperforms the other three baselines with much less trainable parameters and higher accuracy. 

\begin{table}[t!]
\caption{Number of trainable parameters, training time and RMSE of prediction for each model.}
\vspace{-15pt}
\label{tb:compare}
\begin{center}
\begin{small}
\begin{tabular}{lccccc}
\toprule 
\multirow{2}{*}{Dataset} & \multirow{2}{*}{Model} & \multirow{2}{*}{No. of Parameters} & \multirow{2}{*}{Time per Epoch (s)} & \multicolumn{2}{c}{RMSE} \\
\cmidrule(lr){5-6}
&&&  &Training&Extrapolation \\
\midrule
\multirow{4}{*}{2D GS} & PeRCNN & 7,430 & 3.68 & $1.0\times10^{-2}$ & $5.1\times10^{-2}$ \\
      & ConvLSTM & 110,530 & 3.70 & $1.9\times10^{-2}$ & $1.5\times10^{-1}$ \\
      & ResNet  &  38,274 &  2.16 &   $1.0\times10^{-1}$    &  $1.6\times10^{2}$  \\
      & PDE-Net  &  3,614 &  4.71 &   $1.1\times10^{-1}$    &  $1.3\times10^{1}$  \\
      & DHPM & 26,624 & 0.11 & $2.7\times10^{-2}$ & $2.9\times10^{-1}$ \\
\midrule
\multirow{4}{*}{3D GS} & PeRCNN & 10,118 & 2.43 & $1.9\times10^{-2}$ & $3.7\times10^{-2}$ \\
    & ConvLSTM & 32,034 & 22.73 & $8.3\times10^{-2}$ & $1.6\times10^{-1}$ \\
    & ResNet & 29,506 &  2.87 & $1.2\times10^{-1}$ & $1.9\times10^{-1}$ \\
    & PDE-Net  &  17,614 &  3.08 &   $2.0\times10^{-1}$    &  $2.0\times10^{-1}$  \\
    & DHPM & 26,624 &  0.19 & $1.0\times10^{-1}$ & $4.5\times10^{-1}$ \\
\bottomrule
\end{tabular}
\end{small}
\end{center}
\end{table}

\subsection{Interpretability of the learned model}
\label{sec:interpretability}

One major drawback of the traditional deep neural neural network is the lack of interpretability because the output of the network is usually expressed as a prolonged nested function. However, since each channel of the input to the $\Pi$-block (i.e., $\widehat{\boldsymbol{\mathcal{U}}}^{(k)}$) corresponds to a solution component (i.e., $[u,v]$), the multiplicative form of $\Pi$-block makes it possible to extract (or interpret) the explicit form of learned $\mathcal{F}$ from the model's weights and biases via symbolic computations. The purpose of this subsection is to briefly demonstrate how the learned model can be interpreted as an explicit expression, which would be useful for researcher to disentangle the underlying physics. 

We first interpret the learned model for 3D GS RD system from Section \ref{sec:data_driven_3d_rd}. Since the parallel Conv layers (in $\Pi$-block) have filter size of 1, which means that each output channel would represent the linear combination of $u$, $v$ and a constant, the elementwise product of three Conv layers would produce third degree polynomials, which correspond to the reaction term of the 3D GS RD system. The extracted reaction term reads 
\begin{equation}
    \label{eq:factor_out_terms}
    \resizebox{0.99\linewidth}{!}{$
    \mathbf{R(u)}=
    \begin{bmatrix}
    -0.0074u^3 - 0.0051u^2v - 0.2uv^2 - 0.0386v^3 - 0.0018u^2 - 0.11uv- 0.055v^2 - 0.016u  - 0.022v + 0.025 \\
    0.0005u^3 - 0.013u^2v + 0.54uv^2 - 0.087v^3 - 0.0076u^2 + 0.023uv + 0.046v^2 + 0.017u - 0.036v - 0.0097
    \end{bmatrix} 
    $}
\end{equation}
In addition, the identified diffusion coefficient matrix from the highway diffusion Conv layer shows $\mu_u=0.18$ and $\mu_v=0.080$. Comparing the extracted terms with the ground truth, we can see there are some distracting terms as a result of the 10\% noise in the measurement data and the model's redundancy. This result shows the promise of our model on the data-driven discovery of the governing PDE, which would be studied in-depth in the next section.

To interpret terms involving partial derivatives (e.g., $u\Delta u$, $uu_x$) would require us to freeze or constrain a portion of the convolutional filters in the network, like in \cite{long2018pde}. Here a simple experiment is conducted using the 2D Burgers' dataset, whose governing equation is described by Eq. \eref{eq:burgers}. The network employed to construct the data-driven model has two parallel Conv layers. Two channels of the first Conv layer is associated with $\partial (\cdot)/\partial x$ and $\partial (\cdot)/\partial y$ respectively, by fixing the Conv filter with the corresponding FD stencils. The other Conv layer uses filters of size 1 to represent the linear combination of $u$, $v$ and 1. The training measurement data includes 21 $51\times51$ snapshots from $t=0$ to $0.1$ downsampled from the original numerical solution (see Table \ref{tb:data_set}). The noise-free synthetic measurement data is used in this example. Adam optimizer is employed for the training while learning rate starts at 0.002 and decays by 0.02 every 200 iterations. 

We extract the analytical expression from the trained PeRCNN model, as follows
\begin{equation}
    \label{eq:burgers_terms}
    \resizebox{0.9\linewidth}{!}{$
    \mathbf{u}_t=
    \begin{bmatrix}
        \begin{aligned}
    &0.0051 \Delta u - 0.95u_x(1.07u - 0.0065v - 0.17) + 0.98u_y(0.0045u - 1.01v + 0.17) + 0.053 \\
    &0.0051 \Delta v -0.82v_x(1.22u + 0.0078v - 0.18) - 0.91v_y(0.0063u + 1.08v - 0.17) + 0.058
        \end{aligned} 
    \end{bmatrix} 
    $}
\end{equation} 
which shows that the equivalent expression of the learned model is close to the genuine governing PDE (see Eq. \eref{eq:burgers}), which helps to explain the extraordinary generalization capability of our model. Although the selection of differential operators to be encoded is crucial for identifying the genuine form of the $\mathcal{F}$, the above two examples demonstrate better interpretability of PeRCNN over the traditional FCNN or CNN based methods. 

\subsection{Generalization to different initial condition}
In addition to the remarkable extrapolation capability, we find the trained PeRCNN model can generalize to different initial conditions. To verify this, we employ the trained model from previous sections to perform inference with a different initial condition. The computational parameters and boundary condition are kept the same as that of the training measurement data. It should be noted that the DHPM, which roots in FCNN, is incapable of performing inference with different IC. Therefore, it is not considered in this discussion. 

\begin{figure}[h!]
\centering
\includegraphics[width=0.99\textwidth]{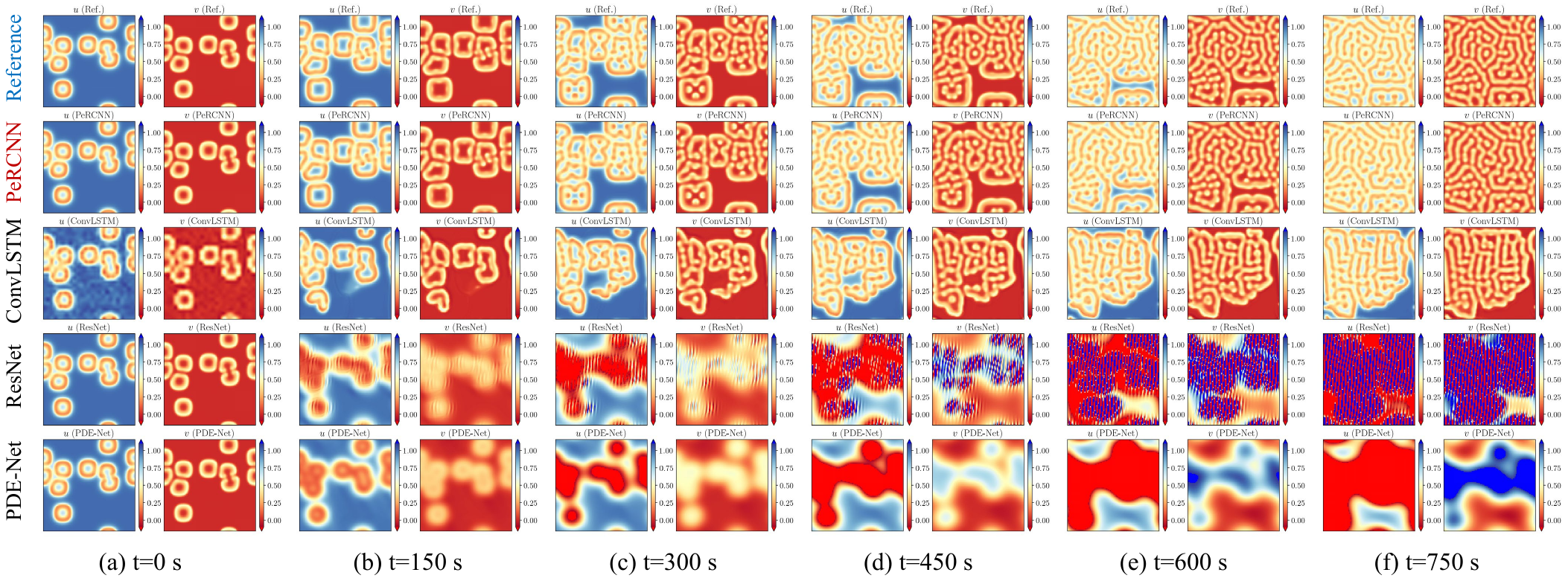}
\caption{Predictions of learned models on a different initial condition. Trained models are from Section \ref{sec:data_driven_2dgs}.}
\label{fig:2dgs_infer_diff_ic}
\end{figure}


\begin{figure}[h!]
\centering
\includegraphics[width=0.99\textwidth]{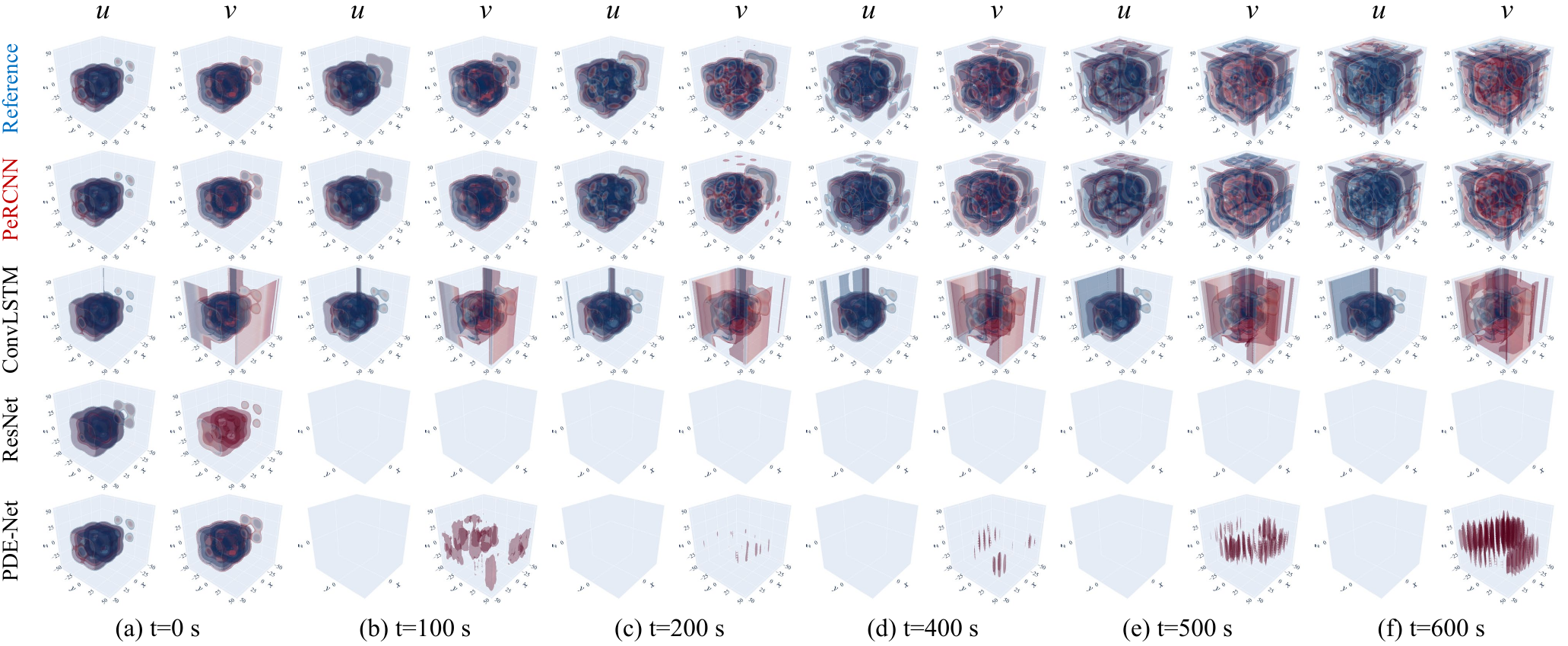}
\caption{Predictions of learned models on a different initial condition. Trained models are from Section \ref{sec:data_driven_3d_rd}.}
\label{fig:3dgs_infer_diff_ic}
\end{figure}

Full-resolution IC is used for the inference. The prediction given by trained models, as well as the ground truth generated with FD solver, is presented in Fig. \ref{fig:2dgs_infer_diff_ic} and \ref{fig:3dgs_infer_diff_ic}. It can be seen that PeRCNN gives consistent prediction compared with the ground truth while the recurrent ResNet and ConvLSTM give wild prediction. From Section \ref{sec:interpretability}, we know that each learned model is associated with an equivalent expression, which is independent of IC. Therefore, we can assume the learned model provides a parameterized template for predicting various systems given the IC. 

\section{Data-driven PDE discovery from scarce data with PeRCNN}

\subsection{Problem description}
\label{sec:disc_pd}

The governing PDEs of many underexplored processes, such as climate systems, biochemical reaction and epidemiology, remains uncertain or partially unknown. To find the governing PDE that is consistent with the observed measurement data is of significance for scientists to understand those underexplored processes. In this section, we primarily focus on the data-driven PDE discovery problem with scarce measurement data. To formalize this problem, let us again consider the dynamical system described by Eq. \eref{eq:dynamic_system_pde}. The objective of the data-driven PDE discovery is to uncover the closed form of $\mathcal{F}$ provided time series measurements on some fixed spatial locations. Although this problem has been studied extensively for decades, the combinatorically large search space of the possible PDE appears as a major obstacle to hinder the discovery of the governing PDE from measurements data. Even worse, the existence of sensing noise and coarse sampling for the measurement also add to the difficulty of solving this problem. To address these challenges, we extend our PeRCNN for discovering the explicit form of the governing PDE of a dynamical system by integrating it with sparse regression. We make a key assumption (based on observation) that the RHS of governing PDEs for many systems of interest consists of only a few terms. That is to say, the $\mathcal{F}$ demonstrates sparsity in the space of possible functions. Based on this assumption, we are able to utilize the sparse regression \cite{brunton2016discovering} as a crucial helper to identify the possible terms in a PDE. 
\textcolor{blue}{Main Text Fig. 5} demonstrates the procedures of the data-driven discovery of governing PDEs with the PeRCNN, namely the hybrid physics-encoded DL (PeDL) framework. This framework consists of three distinct procedures including data reconstruction, sparse regression and coefficients fine-tuning. Provided some noisy and LR snapshots of the system, the data reconstruction process (same as in Section \ref{sec:data_driven_model}) is first conducted to obtained a high-resolution data-driven predictive model. In this step, prior physics knowledge could be encoded into the network to facilitate the training, as introduced in Section \ref{sec:encoding_mechanism}. Once the data-driven model is established, we generate the high-fidelity data (or reconstructed data) for the subsequent sparse regression, which would be detailed in the following section. Sparse regression tells us the active terms from a pool of candidate functions, on which the explicit form of the governing PDE can be assembled. Afterward, we perform the fine-tuning process to further improve the accuracy of the scalar coefficient associated with each term. The fine-tuning step follows the same routine as described in Section \ref{sec:scalar_id}. Note that the PeRCNN models employed in data reconstruction and coefficients fine-tuning have different recurrent blocks. The former involves the $\Pi$-block as a general data approximator with little prior knowledge encoded, while the latter is completely defined by the form of the PDE from the sparse regression result. The differences between these two networks are well illustrated by Fig. \ref{fig:two_form_recurrent_block}.


\subsection{Methodology}

To discover the governing PDE from measurement data, we proposed a hybrid physics-encoded learning framework for equation discovery, which consists of three distinct steps. We would use the 2D GS RD equation as an example to demonstrate how the proposed framework works. In-depth discussions on the methodological details and results, e.g., effects of noise level, importance of the fine-tuning step, selection of hyperparameters and candidate set in the library matrix are provided in the following sections. 
\paragraph{Data reconstruction:} As we primarily seek to solve the equation discovery problem with noisy and scarce (or low-resolution) measurement data, to construct the linear system directly using FD on raw data would introduce large truncation error. Therefore, people usually employ various approximators (e.g., polynomials or splines) to reconstruct the high-fidelity data. As we have seen in Section \ref{sec:data_driven_model}, on a variety of nonlinear systems, PeRCNN model shows effectiveness in reconstructing the high-fidelity solution given some low-resolution snapshots of the system. Therefore, in the first step of the proposed framework, we employ PeRCNN to establish a data-driven model for high-fidelity data reconstruction. Specifically, in this 2D GS RD case, we utilize the $\Pi$-block with 3 parallel Conv layers, 8 channels and filter size of 1. Other settings (e.g., train/validation dataset split, hyperparameters, etc.) are kept the same as in Section \ref{sec:data_driven_2dgs} except the noise level of the training dataset is 5\% here. Once the training is finished, we generate the reconstructed data, which is spatiotemporally high-resolution, for the subsequent sparse regression. The reconstructed data compared with the ground truth is shown in Fig. \ref{fig:recontruct}(c). 

\paragraph{Sparse regression:} With the reconstructed high-fidelity data ($\widehat{\boldsymbol{\mathcal{U}}}\in\mathbb{R}^{801\times2\times101\times101}$) from the previous step, we are able to establish the linear system for the sparse regression, which is elaborated in Section \textcolor{blue}{Data-driven Discovery of PDEs} of the \textcolor{blue}{Main Text}. Specifically, the Sequential Threshold Ridge regression (STRidge) algorithm (see Algorithm \ref{alg:STR} for pseudocode) \cite{rudy2017data} is adopted to obtain a sparse representation of $\mathcal{F}$ for a given tolerance $\delta$ that filters the entries of $\mathbf{\Xi}$. Iterative search with STRidge can be performed to find the optimal tolerance according to the selection criteria (see \textcolor{blue}{Main Text Eqn (8)}). The derivative terms involved in the linear system are computed through numerical differentiation. Since we employ the $\Pi$-block with 3 parallel Conv layers and filter size of 1, the equivalent expression of the approximated $\mathcal{F}$ would be a third degree polynomial. Therefore, power functions of $\mathbf{u}$ up to third degree, i.e., $\mathbf{\Theta}(u,v)=[1,u, v, u^2, uv, v^2, u^3, u^2v, uv^2, v^3]$, are considered in the library matrix. Note that this way of choosing candidate terms in the library matrix is different from the traditional sparse regression in which the candidate set is predefined \cite{brunton2016discovering, rudy2017data}. A detailed discussion on it is provided in Section \ref{sec:disc:selection_library}. In this 2D GS RD example, the coefficient vector obtained from the sparse regression is given below
\begin{equation}
    \widehat{\mathbf{\Xi}}=
    \begin{bmatrix}
    0.034 & -0.034 & 0 & 0 & 0 & 0 & 0 &  & -0.87 & 0 \\
    0 & 0 & -0.095 & 0 & 0 & 0 & 0 & 0 & 0.95 & 0
    \end{bmatrix} ^\texttt{T}
\end{equation} 
which indicates the set of active terms for each component are $\mathcal{S}_u=\{\Delta u,uv^2,u,1\}$ and $\mathcal{S}_v=\{\Delta v,uv^2,v\}$. Note that the governing equations for $u$ and $v$ are discovered separately. The selection of hyperparameter in sparse regression is discussed further in Section \ref{sec:disc:pareto_analysis}. 

\paragraph{Coefficients fine-tuning: } The obtained coefficients from sparse regression may not fully exploit all the available measurement as the regression is performed on subsampled reconstructed HR data. To further improve the result of equation discovery, we present a fine-tuning step to produce the final explicit governing equation. With the explicit form of the PDE from sparse regression, we establish a completely physics-based recurrent network except that some coefficients are treated as trainable variables (see Fig. \ref{fig:two_form_recurrent_block}\textbf{b}). The same settings in Section \ref{sec:senario1} are followed here except that the trainable coefficients are initialized with the nonzero entries obtained from sparse regression. By minimizing the MSE of the network's prediction and the measurement data, we can obtain the governing PDE with the fine-tuned coefficients, which reads

\begin{equation}
    \label{eq:fine_tined_eqn}
    \mathbf{u}_t=
    \begin{bmatrix}
        \begin{aligned}
    &2.001\times10^{-5} \Delta u - 1.003uv^2 - 0.04008u + 0.04008 \\
    &5.042\times10^{-6} \Delta v + 1.009uv^2 - 0.1007v
        \end{aligned}
    \end{bmatrix} 
\end{equation} 
Compared with the values in Table \ref{tb:data_set}, it can be seen that the fine-tuned coefficients are of high accuracy. In Section \ref{sec:disc:importance_ft}, we argue the necessity of the fine-tuning step and the accuracy boost brought by it.

\begin{algorithm}[t!]
	\caption{\small{Sequential threshold ridge regression: $\widehat{\mathbf{\Xi}} = \texttt{STRidge}\big(\mathbf{U}_t, \mathbf{\Theta}, \delta\big)$}}
	\label{alg:STR}
    \footnotesize
    \begin{algorithmic}[1]
		
		\STATE {\bfseries Input:} Left hand side (LHS) of the linear system $\mathbf{U}_t$, library matrix assembled from candidate functions $\mathbf{\Theta}(\mathbf{U})$, and threshold tolerance $\delta$. \vspace{1pt} 
		
		\textcolor{gray}{\# weighing coefficient $\lambda$ is set to be $10^{-3}$}
		\STATE Initialize coefficients $\widehat{\boldsymbol{\Xi}}$ from the standard Ridge regression result, i.e., $\big ( \mathbf{\Theta^\text{T}\Theta}+\lambda \mathbf{I} \big )^{-1} \mathbf{\Theta^\text{T}} \mathbf{U}_t$. \vspace{1pt}      
		
		\REPEAT \vspace{1pt} 
		
		\STATE Determine indices of coefficients in $\widehat{\boldsymbol{\Xi}}$ falling below or above the sparsity threshold $\delta$: \par \vspace{1pt} {\begin{center} $\textit{bigidx} = \{i: |\widehat{\mathbf{\Xi}}[i]| \geq \delta\}$ and $\textit{smallidx} = \{j: |\widehat{\mathbf{\Xi}}[j]| < \delta\}$. \end{center}} \vspace{1pt} 
		
		\STATE Enforce sparsity to small values by setting them to zero: $\widehat{\boldsymbol{\Xi}}[\textit{smallidx}] = \mathbf{0}$. \vspace{1pt} 
		
		\STATE Update remaining non-zero values with standard Ridge regression:
		\par \vspace{1pt}  {\begin{center}$\widehat{\mathbf{\Xi}}[\textit{bigidx}] = \operatorname*{arg\,min}_{\mathbf{\Xi}[\textit{bigidx}]} \big\{ \big\|\mathbf{\Theta}[bigidx]\mathbf{\Xi}[bigidx] - \mathbf{U}_t \big\|_2^2 + \lambda\big\|\mathbf{\Xi}[bigidx]\big\|^2_2\big\}$. ~~~  \end{center}}
		
		\UNTIL{maximum number of iterations reached or size of \textit{bigidx} unchanged for two consecutive iteration.} \vspace{2pt} 
		
		\STATE Finalize non-zero coefficients with standard least square regression.
		
		{\begin{center}$\widehat{\mathbf{\Xi}}[\textit{bigidx}] = \operatorname*{arg\,min}_{\mathbf{\Xi}[\textit{bigidx}]} \big\{ \big\|\mathbf{\Theta}[bigidx]\mathbf{\Xi}[bigidx] - \mathbf{U}_t \big\|_2^2 \big\}$. ~~~  \end{center}}
		
		\STATE {\bfseries Output:} the best solution $\widehat{\mathbf{\Xi}}$.
	
	\end{algorithmic} 
\end{algorithm}

\subsection{Experimental setup}
\label{sec:exper_settings}

\paragraph{Numerical examples:} We validate our approach with three different examples, including the 2D $\lambda$--$\Omega$, 2D GS RD and 2D Burgers' equation system. The 2D Burgers' case is considered in this section as a complementary example to demonstrate the effectiveness of our method. The governing equation of each system, as well as the computational parameters employed for generating the numerical solutions, is provided in Table \ref{tb:data_set}. 

\paragraph{Baselines:} To compare our approach with some existing methods, we also experiment with the widely used PDE-FIND \cite{rudy2017data}, in which the FD is used to obtain the derivatives for noiseless data while the derivatives of the fitted Chebyshev polynomial are used for noisy data. In addition, we also consider the baselines of the sparse regression (SR) coupled with a fully connected neural network (namely FCNN+SR) or PDE-Net (namely PDE-Net+SR). In these two baselines, the FCNN and PDE-Net \cite{long2018pde} are used respectively to fit the LR measurement and perform inference for HR data while SR is used to discover the PDE. As we assume the ubiquitous diffusion phenomenon exists in the concerned system, terms that are known to exist in the PDE (e.g., the diffusion term $\Delta\mathbf{u}$) would be exempted from being filtered in sparse regression for each method. 

\paragraph{Evaluation metrics:} The relative $\ell_2$ error, defined as $ E=||\mathbf{\Xi}_\text{id}- \mathbf{\Xi}_\text{true}||_2/||\mathbf{\Xi}_\text{true}||_2$, measures the relative distance between the identified coefficient vector $\mathbf{\Xi}_\text{id}$ and the ground truth $\mathbf{\Xi}_\text{true}$ where $||\cdot||_2$ denotes the $\ell_2$ norm. However, when the magnitude of coefficients vary significantly (e.g., the 2D GS RD equation), $E$ cannot reflect the result well as the small coefficients get overwhelmed. Hence, we introduce the following non-dimensional measures to gain a more insightful evaluation of the results. Since the PDE discovery can also be considered as a binary classification problem (e.g., whether a term exists or not) given a candidate library, we introduce the precision and recall to evaluate the performance of the proposed method. The recall measures the percentage of the successfully identified coefficients among the true coefficients, defined as $R=||\mathbf{\Xi}_\text{id}\odot \mathbf{\Xi}_\text{true}||_0/||\mathbf{\Xi}_\text{true}||_0$ where $\odot$ denotes elementwise product of two vectors and $||\cdot||_0$ denotes the number of nonzeros in a vector. Here a successful identification occurs when the entries in both identified and true vectors are nonzero. Similarly, the precision has the definition of $P=||\mathbf{\Xi}_\text{id}\odot \mathbf{\Xi}_\text{true}||_0/||\mathbf{\Xi}_\text{id}||_0$.

\paragraph{Generation of the measurement data:} Our approach seeks to overcome the challenges brought by the noisy and LR measurement in the governing equation discovery problem. Therefore, we synthetized the measurement data by downsampling (both spatially and temporally) and incorporating artificial Gaussian noise into the numerical solutions. In the FD simulation, 9-point stencil is used for computing the spatial derivatives while Runge-Kutta scheme is used for time stepping. Periodic boundary condition is adopted for all three systems. The measurement data used in equation discovery is obtained by downsampling the high-resolution numerical solution and adding Gaussian noise. In both the 2D Burgers' and $\lambda$--$\Omega$ RD case, we downsampled the numerical solution clip of shape ${201\times2\times101\times101}$ to the measurement data as $\Tilde{\mathbf{u}}\in\mathbb{R}^{41\times2\times51\times51}$. In the 2D GS RD case, we considered an extremely LR measurement, i.e., $\Tilde{\mathbf{u}}\in\mathbb{R}^{161\times2\times26\times26}$ downsampled from the raw data of shape ${801\times2\times101\times101}$. To mimic the measurement collected in real world, we also add different levels (i.e., noise-free, 5\% and 10\%) of Gaussian noise to the original numerical solution before downsampling for the LR measurement. 

\subsection{Results and comparison with existing methods}

\begin{figure}[t!]
    \centering
	\includegraphics[width=1.0\linewidth]{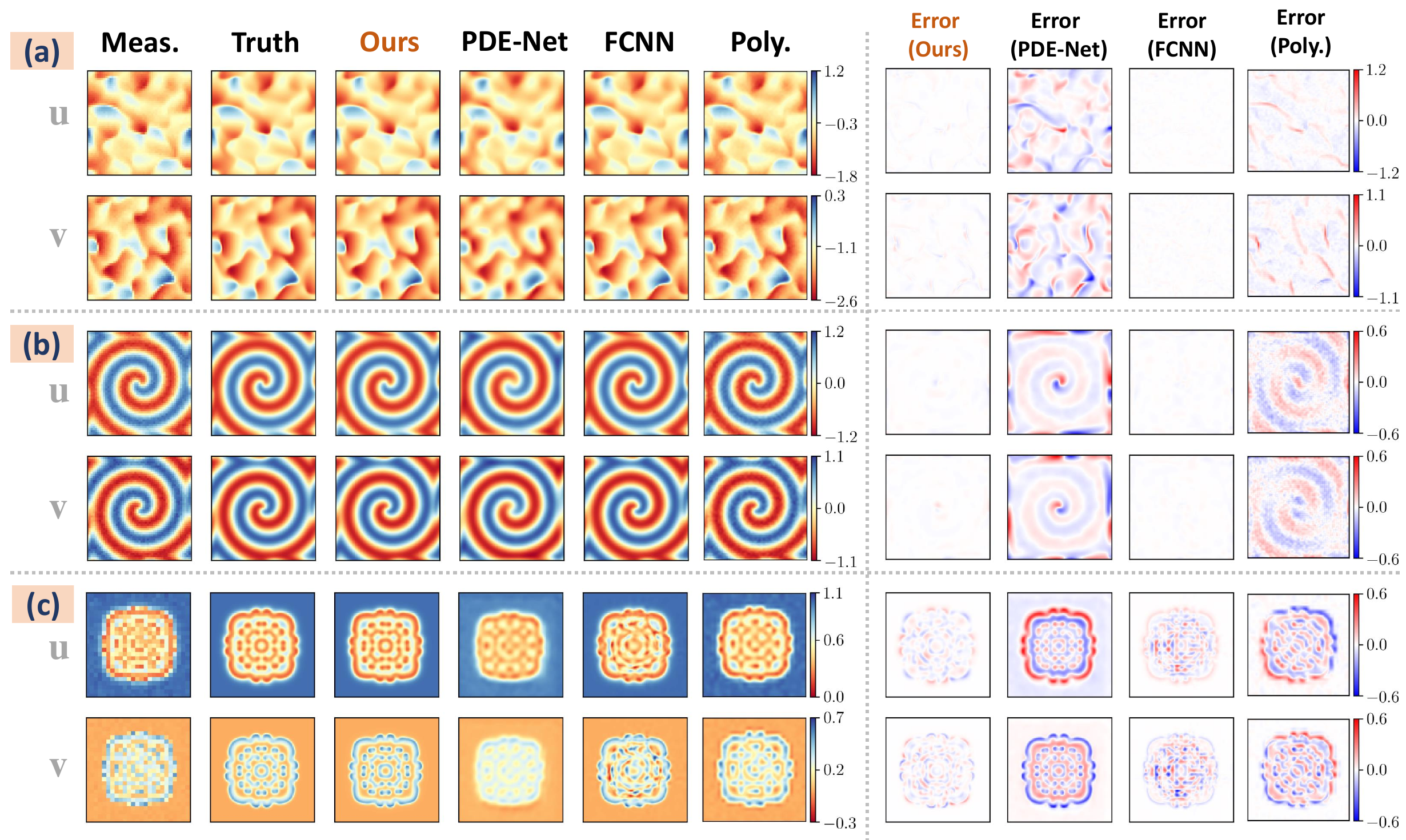} 
	\caption{Snapshots of the state variable at one time instance \cite{rao2022discovering}. (a)-(c) represent the 2D Burgers', $\lambda$--$\Omega$ and GS RD systems. Six columns starting from the left denote the LR measurement, HR ground truth, reconstructed HR solution from our model, PDE-Net, FCNN and Chebyshev polynomial fitting (i.e., PDE-FIND) respectively. The reconstruction error of each model is also provided. }
	\vspace{0pt}
	\label{fig:recontruct}
\end{figure}

\begin{table*}[t!]
\center
\footnotesize
\caption{ {Performance comparison between our proposed framework and baselines \cite{rao2022discovering}.}}
\begin{tabular}{lcccccccccc}
\toprule
\multirow{2}{*}{Cases} & Metrics & \multicolumn{3}{c}{Relative $\ell_2$ error~/$\times10^{-2}$} & \multicolumn{3}{c}{Precision~/\%} & \multicolumn{3}{c}{Recall~/\%}  \\
\cmidrule(lr){3-5}\cmidrule(lr){6-8}\cmidrule(lr){9-11}
 &  Noise level& 0\% & 5\% & 10\% & 0\% & 5\% & 10\% & 0\% & 5\% & 10\% \\
\midrule
\multirow{4}{*}{Burgers'} & Ours & 0.50 & 0.54& \textbf{0.59} & \textbf{100} & \textbf{100} & \textbf{100} & \textbf{100} & \textbf{100} & \textbf{100}\\
\cmidrule{2-11}
&  {FCNN+SR} & \textbf{ {0.37}} & \textbf{ {0.43}} &  {1.34} & \textbf{ {100}} & \textbf{ {100}}  &  {75.0} & \textbf{ {100}} & \textbf{ {100}} & \textbf{ {100}}\\
\cmidrule{2-11}
&  {PDE-Net+SR} &  {339.7} &  {442.3} &  {372.0} &  {40.0} &  {25.0}  &  {28.5} &  {66.7} &  {50.0} &  {33.3}\\
\cmidrule{2-11}
 & PDE-FIND & 3.32 & 36.80 & 45.09  & 75 & \textbf{100}  & 62.5 & \textbf{100} & 83.3 & 83.3\\
\midrule
\multirow{4}{*}{$\lambda$--$\Omega$ RD} &  Ours & \textbf{1.18}& \textbf{2.69}&  \textbf{5.44} &\textbf{100} & \textbf{100} & \textbf{91.6} &\textbf{100} &\textbf{100} & \textbf{100}\\
\cmidrule{2-11}
 &  {FCNN+SR} &  {5.19} &  {7.50} &  {15.85} & \textbf{ {100}} &  {85.7}  &  {85.7} & \textbf{ {100}} & \textbf{ {100}} & \textbf{ {100}}\\
\cmidrule{2-11}
&  {PDE-Net+SR} &  {62.23} &  {107.45} &  {104.63} &  {81.82} &  {64.3}  &  {50.0} &  {75.0} &  {75.0} &  {50.0}\\
\cmidrule{2-11}
 & PDE-FIND & 1.70 & 92.52 & 99.15 & \textbf{100} & 83.3  & 77.8 & \textbf{100} & 62.5 & 58.3\\
\midrule
\multirow{4}{*}{GS RD} & Ours & \textbf{1.59} & \textbf{2.85} & \textbf{10.03} & \textbf{100} & \textbf{100} & \textbf{85.7} &\textbf{100} & \textbf{100} & \textbf{85.7}\\
\cmidrule{2-11}
&  {FCNN+SR} &  {95.14}  &  {143.55}  &  {162.98}  &   {37.5} &   {33.3} &  {33.3}  &  {60.0}  &  {57.1} &  {57.1} \\
\cmidrule{2-11}
&  {PDE-Net+SR} &  {204.61} &  {100.00} &  {382.20} &  {30.8} &  {40.0}  &  {30.0} &  {57.1} &  {28.6} &  {42.9}\\
\cmidrule{2-11}
 & PDE-FIND & 113.05 & 89.99 & 108.1 & 45.5 &  50.0 & 42.9 & 85.7 & 57.1& 60.0\\
\bottomrule
\end{tabular}
\vspace{-6pt}
\label{tb:summary_result}
\end{table*}

\paragraph{2D Burgers' equation:} We first test the proposed approach on the 2D Burgers' equation. To reconstruct the high-fidelity data, we use the PeRCNN with 3 parallel Conv layers of 16 channels and filter size of 5. Furthermore, in this and the following examples, we assume the measurement exhibits the ubiquitous diffusion phenomenon. That said, the diffusion layer would be encoded into the network architecture. Also in the sparse regression stage, the coefficient of diffusion term would be exempted from being filtered. We train the network with Adam optimizer for 15,000 iterations. The learning rate is initialized to be $0.002$ and decreases to 97\% of the previous for every 200 iterations. 

Once the training is completed, the HR solution (i.e., 201 snapshots of $101\times101$) can be inferred from the trained model. Figure \ref{fig:recontruct}(a) provides the snapshots of the reconstructed HR data under 5\% noise from each method. It can be seen that our method and the FCNN has a much smaller reconstruction error, which would give rise to more accurate derivatives for constructing the library matrix. We also observe that PDE-Net struggles in reconstructing the HR data. Based on our experiments and the claim PDE-Net paper \cite{long2018pde} made, this is because the additive operation limits PDE-Net to only learning the response of linear PDEs. Therefore, unlike our model, PDE-Net lacks the capability to exactly express nonlinear terms like $uu_x$ and $u^2v$. As we cannot extract an explicit expression from the PeRCNN model of $5\times5$ Conv filters for specifying the candidate set, here we predefine a group of 70 candidate functions that consists of polynomial terms $\{1,u,v,u^2,uv,v^2,u^3,u^2v,uv^2,v^3\}$, derivatives $\{1,u_x,u_y,v_x,v_y,\Delta u, \Delta v\}$ and their combinations. In addition to the diffusion terms known $a~priori$, the result of sparse regression indicates the presence of several other terms in $\mathcal{F}$, i.e., $\mathcal{S}_u=\{uu_x,vu_y\}$ and $\mathcal{S}_v=\{uv_x,vv_y\}$ respectively. As the last step, the fine-tuning is performed using the network built completed based on the discovered PDE structure in the second stage, which gives us the final discovered equation:
\begin{equation}
    \label{eq:fine_tined_burgers2}
    \begin{aligned}
    u_t&=5.0113\times10^{-3} \Delta u - 1.004uu_x - 1.004vu_y, \\
    v_t&=4.9953\times10^{-3} \Delta v - 1.009uv_x - 1.002vv_y.
    \end{aligned}
\end{equation} 
The precision, recall and relative $\ell_2$ error of the discovered PDE under various noise levels are provided in Table \ref{tb:summary_result}. It can be observed that all methods except PDE-Net+SR perform well when the measurement is noise-free. Furthermore, the performance of baselines deteriorate when the noise level increases to 10\% while our approach demonstrates better robustness against the noise. Our approach achieves ${5.9\times10^{-3}}$ relative $\ell_2$ error, 100\% recall and precision under the 10\% noise level.

\paragraph{2D $\boldsymbol{\lambda}$--$\boldsymbol{\Omega}$ RD equation:}
In the first step of data reconstruction, the recurrent network has 3 parallel Conv layers, 16 channels and $1\times1$ kernels. We train the network with Adam optimizer for 15,000 iterations. The learning rate is initialized to be $0.002$ and decreases to 97\% of the previous for every 200 iterations. Once the training is completed, we infer from the network the HR solution at finer spatiotemporal grid. The reconstructed data and the corresponding error against reference solution of our method and the considered baselines are shown in Fig. \ref{fig:recontruct}(b). Since 3 parallel Conv layers and $1\times1$ kernels are adopted, the resultant expression of the recurrent network would be a third degree polynomial, which is used to establish the library matrix $\mathbf{\Theta}(u,v)=[1, u, v, u^2, uv, v^2, u^3, u^2v, uv^2, v^3]$. We randomly subsample 10\% of the HR points for the sparse regression, whose result (5\% noise case) gives the set of terms with nonzero coefficients, i.e., $\mathcal{S}_u=\{\Delta u,u,u^3,u^2v,uv^2,v^3\}$ and $\mathcal{S}_v=\{\Delta v,u,u^3,u^2v,uv^2,v^3\}$. With the set of existing terms in the governing PDE, the fine-tuning step is performed using all the available measurement. The obtained scalar coefficients from the fine-tuning renders us with the final governing equation, which reads as
\begin{equation}
    \label{eq:fine_tined_lam_omg}
    \begin{aligned}
    u_t=& 0.096 \Delta u + 1.038u - 1.048u^3 + 1.004u^2v - 1.050uv^2 + 0.998v^3, \\
    v_t=& 0.100 \Delta v + 1.014v - 0.998u^3 - 1.025u^2v - 0.999uv^2 - 1.015v^3.
    \end{aligned}
\end{equation} 
The quantitative evaluation metrics computed from the discovered PDE under various noise levels are provided in Table \ref{tb:summary_result}. Our approach achieves $5.44\times10^{-2}$ relative error, 100\% recall and 91.6\% precision under 10\% noise. Empirical study in Section \ref{s:larger_noise} shows our method is also capable of handling the measurement data with larger noise level.

\paragraph{2D GS RD equation:} In this example, we assume the measurements are some extremely LR (i.e., ${26\times26}$) snapshots (see measurements in Fig. \ref{fig:recontruct}(c)). As the first step, we employ the network with 3 parallel Conv layers, 8 channels and filter size of 1 to reconstruct the data. From the reconstruction model, we infer the HR (i.e., ${101\times101}$) solution at finer time instances. Snapshots of HR solution in Fig. \ref{fig:recontruct}(c) show our method is able to restore the high-fidelity data from the LR and noisy measurement very well. Given the network configurations, a library of polynomials up to the third degree $\mathbf{\Theta}(u,v)=[1, u, v, u^2, uv, v^2, u^3, u^2v, uv^2, v^3]$ is constructed for sparse discovery. The sparse regression is performed on subsampled (10\%) linear system, which tells us the set of existing terms $\mathcal{S}_u=\{\Delta u,1,u,uv^2\}$ and $\mathcal{S}_v=\{\Delta v,v,uv^2\}$. With the existing terms in the PDE, we perform the fine-tuning using the completely physics-based Conv block. The final obtained PDEs in the case of 5\% noise are
\begin{equation}
    \label{eq:fine_tined_gs}
    \begin{aligned}
    u_t=&2.001\times10^{-5} \Delta u - 1.003uv^2 - 0.04008u + 0.04008, \\
    v_t=&5.042\times10^{-6} \Delta v + 1.009uv^2 - 0.1007v.
    \end{aligned}
\end{equation} 
As presented in Table \ref{tb:summary_result}, our proposed approach performs well on discovering the PDE as a result of fully utilizing the prior physics knowledge and the powerful expressiveness of the model, while PDE-FIND, PDE-Net+SR and FCNN+SR struggle due to the extremely LR measurements.

\begin{table}[t!]
\caption{Discovered PDE from our method compared with the ground truth. The false positive terms are highlighted in red. }
\vspace{-15pt}
\label{tb:result_pde_discovery}
\begin{center}
\begin{small}
\begin{tabular}{lll}
\toprule 
Example & Noise & Discovered PDE \\
\midrule
\multirow{8}{*}{2D Burgers'} & 0\% & 
    $\begin{aligned}
    u_t&=-4.973\times10^{-3} \Delta u + 1.002uu_x + 1.003vu_y \\
    v_t&=-4.983\times10^{-3} \Delta v + 1.009uv_x + 1.002vv_y
    \end{aligned}$\\
                \cmidrule(lr){2-3}
                & 5\%  &  
    $\begin{aligned}
    u_t&=-5.011\times10^{-3} \Delta u + 1.004uu_x + 1.004vu_y \\
    v_t&=-4.995\times10^{-3} \Delta v + 1.009uv_x + 1.002vv_y
    \end{aligned}$\\
                \cmidrule(lr){2-3}
                & 10\% & 
    $\begin{aligned}
    u_t&=-5.136\times10^{-3} \Delta u + 1.006uu_x + 1.008vu_y \\
    v_t&=-5.056\times10^{-3} \Delta v + 1.006uv_x + 1.001vv_y
    \end{aligned}$\\
                \cmidrule(lr){2-3}
                &  Truth & 
    $\begin{aligned}
    u_t&=-5.0\times10^{-3} \Delta u + 1.0uu_x + 1.0vu_y \\
    v_t&=-5.0\times10^{-3} \Delta v + 1.0uv_x + 1.0vv_y
    \end{aligned}$\\
\midrule
\multirow{8}{*}{2D $\lambda$--$\Omega$ RD} & 0\%  &  
    $\begin{aligned}
    u_t&=0.096\Delta u + 1.013u - 1.019u^3 + 1.001u^2v -1.021uv^2 + 0.9977v^3 \\
    v_t&=0.096\Delta v + 1.006v - 0.998u^3 -1.0139u^2v -1.002uv^2 - 1.012v^3
    \end{aligned}$\\
    \cmidrule(lr){2-3}
    & 5\%  &  
    $\begin{aligned}
    u_t=& 0.096 \Delta u + 1.038u - 1.048u^3 + 1.004u^2v - 1.050uv^2 + 0.998v^3 \\
    v_t=& 0.100 \Delta v + 1.014v - 0.998u^3 - 1.025u^2v - 0.999uv^2 - 1.015v^3
    \end{aligned}$\\
    \cmidrule(lr){2-3}
    & 10\% &  
    $\begin{aligned}
    u_t&=0.101\Delta u + 1.079u - 1.090u^3 + 1.008u^2v -1.090uv^2 + 0.9982v^3\\
    v_t&=0.105\Delta v + 1.033v - 0.965u^3 - 1.046u^2v - 0.967uv^2 - 1.029v^3 + {\color{black}0.029u}
    \end{aligned}$\\
    \cmidrule(lr){2-3}
    & Truth &      
    $\begin{aligned}
    u_t=&0.1\Delta u + (1-u^2-v^2)u+ (u^2+v^2)v \\
    v_t=&0.1\Delta v - (u^2+v^2)u + (1-u^2-v^2)v
    \end{aligned}$\\
\midrule
\multirow{8}{*}{2D GS RD} & 0\%  &  
    $\begin{aligned}
    u_t=&1.999\times10^{-5} \Delta u - 0.992uv^2 - 0.04003u + 0.03999 \\
    v_t=&5.008\times10^{-6} \Delta v + 1.021uv^2 - 0.1001v
    \end{aligned}$\\
    \cmidrule(lr){2-3}
    & 5\%  &  
    $\begin{aligned}
    u_t=&2.001\times10^{-5} \Delta u - 1.003uv^2 - 0.04008u + 0.04008 \\
    v_t=&5.042\times10^{-6} \Delta v + 1.009uv^2 - 0.1007v
    \end{aligned}$\\
    \cmidrule(lr){2-3}
    & 10\% &   
    $\begin{aligned}
    u_t=&1.846\times10^{-5} \Delta u - 0.904uv^2 - {\color{black}0.0863u^3} + 0.04019 \\
    v_t=&5.438\times10^{-6} \Delta v + 1.051uv^2 - 0.1174v
    \end{aligned}$\\
    \cmidrule(lr){2-3}
    & Truth &  
    $\begin{aligned}
    u_t=&2.0\times10^{-5} \Delta u - 1.0uv^2 + 0.04(1-u) \\
    v_t=&5.0\times10^{-6} \Delta v + 1.0uv^2 - 0.1v
    \end{aligned}$\\
\bottomrule
\end{tabular}
\end{small}
\end{center}
\end{table}

\subsection{Discussions}

\subsubsection{Formation of the library matrix}
\label{sec:disc:selection_library}

In most existing works, e.g., \cite{brunton2016discovering, rudy2017data, chen2020deep}, people resort to a predefined candidate set for constructing the library matrix. The set of candidates in the library is usually selected to be large enough for representing the governing PDE. However, as we have seen in Section \ref{sec:interpretability}, an analytical expression could be extracted from the learned $\Pi$-block. In our proposed framework, this expression is utilized subsequently to establish the library matrix in the sparse regression, i.e., the candidate set covers all terms in the expression. This practice could guarantee the candidate set used in sparse regression is consistent with the network configurations used in data reconstruction. As a result, it would help improve the computational efficiency of our method by utilizing a compact candidate set, i.e., only those terms exist in the equivalent expression are considered in the linear system. Note that this practice only applies when an analytical expression could be extracted from the $\Pi$-block. An alternative and more general way of constructing the library matrix is to utilize a predefined candidate set. 

To demonstrate the alternative way of forming the library matrix, we experiment on the 2D Burgers' equation. To reconstruct the high-fidelity data, we use the PeRCNN with 3 parallel Conv layers of 16 channels and filter size of 5. These hyperparameters of network architecture are selected through hold-out validation. Furthermore, in this and the following examples, we assume the measurement exhibits the ubiquitous diffusion phenomenon. That said, the diffusion layer would be encoded into the network architecture. Also in the sparse regression stage, the coefficient of diffusion term would be exempted from being filtered. We train the network with Adam optimizer for 15,000 iterations. The learning rate is initialized to be $0.002$ and decreases to 97\% of the previous for every 200 iterations. 

Once the training is completed, the HR solution (i.e., 201 snapshots of $101\times101$) can be inferred from the trained model. Figure \ref{fig:recontruct}(a) provides the snapshots of the reconstructed HR data under 5\% noise from each method. It can be seen that our method and the FCNN has a much smaller reconstruction error, which would give rise to more accurate derivatives for constructing the library matrix. To prepare for the subsequent sparse regression, we pre-define a group of 70 candidate functions that consists of polynomial terms $\{1,u,v,u^2,uv,v^2,u^3,u^2v,uv^2,v^3\}$, derivatives $\{1,u_x,u_y,v_x,v_y,\Delta u, \Delta v\}$ and their combinations. In addition to the diffusion terms known $a~priori$, the result of sparse regression indicates the presence of several other terms in $\mathcal{F}$, i.e., $\mathcal{S}_u=\{uu_x,vu_y\}$ and $\mathcal{S}_v=\{uv_x,vv_y\}$ respectively. As the last step, the fine-tuning is performed using the network built completed based on the discovered PDE structure in the second stage (see \textcolor{blue}{Main Text Fig. 5\textbf{b}}), which gives us the final discovered equation:
\begin{equation}
    \label{eq:fine_tined_burgers1}
    \begin{aligned}
    u_t&=5.0113\times10^{-3} \Delta u - 1.004uu_x - 1.004vu_y, \\
    v_t&=4.9953\times10^{-3} \Delta v - 1.009uv_x - 1.002vv_y.
    \end{aligned}
\end{equation} 
The precision, recall and relative $\ell_2$ error of the discovered PDE under various noise levels are provided in Table \ref{tb:summary_result}. It can be observed that all methods except PDE-Net+SR perform well when the measurement is noise-free. Furthermore, the performance of baselines deteriorate when the noise level increases to 10\% while our approach demonstrates better robustness against the noise. Our approach achieves ${5.9\times10^{-3}}$ relative $\ell_2$ error, 100\% recall and precision under the 10\% noise level.

\subsubsection{Effects of noise level and candidate functions}
\label{sec:disc:effect_noise_terms}

In this part, we will explore the effects of the noise level in measurement data and the candidate set on the discovered equation. In the first experiment, the cases of 2D GS RD system with clean data and noisy data (5\% and 10\% level) are considered. We follow the same procedure as described in previous section and obtain the results shown in Table \ref{tb:effect_noise}. It can be seen that, for the case free of noise and with 5\% Gaussian noise, the sparse regression produces the genuine PDE with good accuracy from the reconstructed HR data. However, for the 10\% noise level, the obtained result includes a false positive term $u^3$ while ignoring genuine $u$ term. This is because the state variables (i.e., $u$ and $v$) of 2D GS RD system are in the range of $[0,~1]$ (see Fig. \ref{fig:recontruct}(c)). As a result, the difference between numerical values of $u$ and $u^3$ is too subtle to be distinguished, especially considering the 10\% Gaussian noise in the data. 

To examine the effects of candidate set adopted in sparse regression, we experiment on the 2D GS RD case with an enlarged candidate sets, i.e., the polynomials up to fourth order (15 functions). In this experiment, the noise level of 5\% is considered while the remaining settings are kept the same as in previous section. The identified coefficients from the sparse regression is presented in Table \ref{tb:4th_order}. It can be seen that nonzero terms match exactly with the result of third order candidate set (see 5\% noise case in Table \ref{tb:effect_noise}) which shows the adaptability of the sparse regression on various candidate sets.

\begin{table}[t!]
\caption{Identified coefficients from sparse regression for each term with different noise level.}
\vspace{-15pt}
\label{tb:effect_noise}
\begin{center}
\begin{small}
\begin{tabular}{lccccccccccc}
\toprule 
Component & Noise & $1$ & $u$ & $v$ & $u^2$ & $uv$ & $v^2$ & $u^3$ & $u^2v$ & $uv^2$ & $v^3$ \\
\midrule
$u_t$ & 0\%  & 0.043 & -0.043 & 0 & 0 & 0 & 0 & 0 & 0 & -1.09 & 0 \\
    & 5\%  & 0.034 & -0.034 & 0 & 0 & 0 & 0 & 0 & 0 & -0.87 & 0 \\
    & 10\% & 0.024 & 0 & 0 & 0 & 0 & 0 & -0.024 & 0 &  -0.95 & 0 \\
    & \bf{Ref.}  & \bf{0.040} & \bf{-0.040} & \bf{0} & \bf{0} & \bf{0} & \bf{0} & \bf{0} & \bf{0} & \bf{-1.00} & \bf{0} \\
\midrule
$v_t$ & 0\%  & 0 & 0 & -0.099 & 0 & 0 & 0 & 0 & 0 & 0.99 & 0 \\
    & 5\%  & 0 & 0 & -0.095 & 0 & 0 & 0 & 0 & 0 & 0.95 & 0 \\
    & 10\% & 0 & 0 & -0.11 & 0 & 0 & 0 & 0 & 0 &  1.16 & 0 \\
    & \bf{Ref.}  & \bf{0} & \bf{0} & \bf{-0.10} & \bf{0} & \bf{0} & \bf{0} & \bf{0} & \bf{0} & \bf{1.00} & \bf{0} \\
\bottomrule
\end{tabular}
\end{small}
\end{center}
\end{table}

\begin{table}[t!]
\caption{Identified coefficients with polynomial basis functions up to 4th order (5\% noise). }
\vspace{-15pt}
\label{tb:4th_order}
\begin{center}
\begin{small} 
\begin{tabular}{lccccccccccccccc}
\toprule 
Component & $1$ & $u$ & $v$ & $u^2$ & $uv$ & $v^2$ & $u^3$ & $u^2v$ & $uv^2$ & $v^3$ & $u^4$ & $u^3v$ & $u^2v^2$ & $uv^3$ & $v^4$  \\
\midrule
$u_t$ & 0.034 & -0.034 & 0 & 0 & 0 & 0 & 0 & 0 & -0.87 & 0 & 0 & 0 & 0 & 0 & 0\\
\midrule
$v_t$  & 0 & 0 & -0.095 & 0 & 0 & 0 & 0 & 0 & 0.95 & 0 & 0 & 0 & 0 & 0 & 0 \\
\bottomrule
\end{tabular}
\end{small}
\end{center}
\end{table}

\subsubsection{Pareto analysis for selecting hyperparameters}
\label{sec:disc:pareto_analysis}

STRidge algorithm aims to solve the optimization problem described by \textcolor{blue}{Eqn (8)} in \textcolor{blue}{Main Text}. The optimization objective aggregates two components, $||\mathbf{U}_t-\mathbf{\Theta(U)\Xi}||_2$ and $||\mathbf{\Xi}||_0$, to balance the model complexity and accuracy. Therefore, to select a suitable hyperparameter (or weighting coefficient) $\gamma$ is important to the final discovered PDE. So far, Pareto analysis is considered to be the most effective way for selecting the $\gamma$. To ensure the two components at the same scale, we introduce an normalized hyperparameter $\kappa$ with the definition $\kappa=\gamma/||\mathbf{U}_t-\mathbf{\Theta(U)\Tilde{\Xi}}||_2$ where $\Tilde{\mathbf{\Xi}}$ denotes the Least Squares solution to \textcolor{blue}{Main Text Eqn (7)}.

\begin{figure}[t!]
    \centering
	\includegraphics[width=\linewidth]{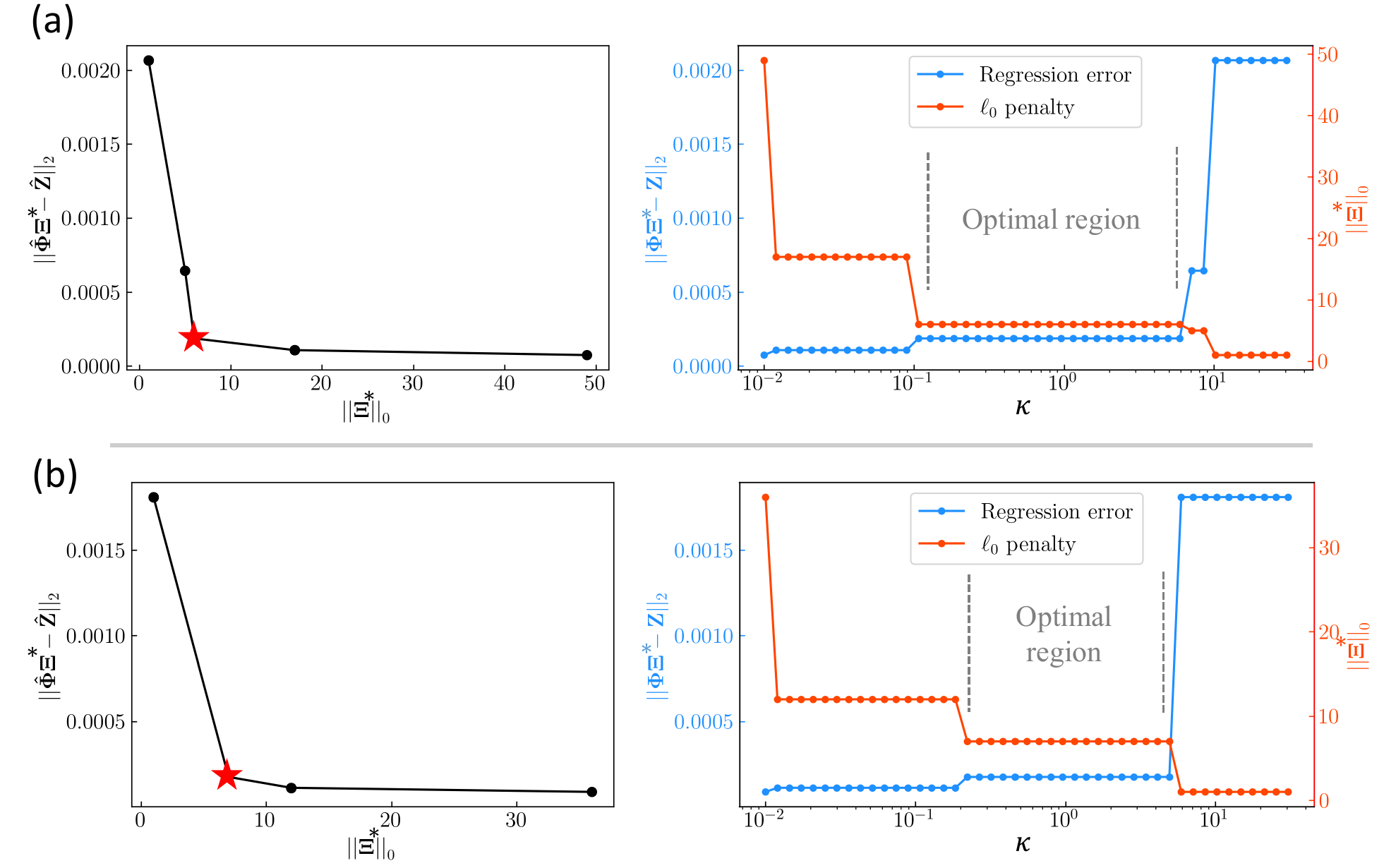} 
	\caption{Pareto analysis for selecting the weighting coefficient $\kappa$ (or $\gamma$) in 2D $\lambda$--$\Omega$ GS system (10\% noise level). (a) and (b) correspond to the $\mathcal{F}_u$ and $\mathcal{F}_v$ respectively. The Pareto front that represents the optimal $\kappa$ is marked with the red star.}
	\label{fig:pareto_analysis_LO}
\end{figure}

In the Pareto analysis, we first sample exponentially uniform points of $\kappa$ within the interval $[10^{-2},~20]$. Each sampled $\kappa$ can be passed to the sparse regression routine, which gives us an optimal coefficient vector $\mathbf{\Xi}^*$. Accordingly, the regression error and number of nonzero coefficients can be computed for Pareto analysis. As an example, the Pareto analysis for 2D $\lambda$--$\Omega$ equation system (10\% noise case) is provided in Fig. \ref{fig:pareto_analysis_LO}. From the left subfigure, we are able to identify the Pareto front, which corresponds to the optimal region of $\kappa$ on the right subfigure. We performed Pareto analyses for each case to find the suitable hyperparameter $\kappa$ for sparse regression which is summarized in Table \ref{tb:kappa_selection}.

\begin{table*}[h!]
\centering
\caption{The coefficient $\kappa$ employed in sparse regression.}
\begin{tabular}{lccccccccc}
\toprule
\multirow{2}{*}{Method}  & \multicolumn{3}{c}{Burgers'}   & \multicolumn{3}{c}{$\lambda$--$\Omega$ RD} & \multicolumn{3}{c}{GS RD}  \\
\cmidrule(lr){2-4}\cmidrule(lr){5-7}\cmidrule(lr){8-10}
 & 0\% & 5\% & 10\% & 0\% & 5\% & 10\% & 0\% & 5\% & 10\% \\
\midrule
Ours &  1 & 1 & 1 & 1 & 1 & 1 & 0.1 & 0.1 & 1 \\
\cmidrule{2-10}
PDE-FIND & 1  & 1 & 1  & 1 & 1 & 1 & 0.1 & 0.05 & 0.1   \\
\bottomrule
\end{tabular}
\label{tb:kappa_selection}
\end{table*}

\subsubsection{Importance of the coefficients fine-tuning}
\label{sec:disc:importance_ft}

The proposed approach consists of three steps including the data reconstruction, sparse regression and the coefficients fine-tuning. However, the sparse regression has already given us the explicit form of the PDE. The last fine-tuning step is introduced to further improve the accuracy of the discovered PDE for two reasons: (1) The result of sparse regression is mainly based on the assumption that the coefficient vector $\mathbf{\Xi}$ is sparse. In fact, the discovered PDE can be further considered as an assumption, on which a completely physics-based recurrent network can be established for the coefficient identification. In the fine-tuning step, the optimization is performed on a much smaller parameter space as the functional form of the PDE is fixed. (2) Due to the high dimensionality of the reconstructed data (e.g., $\mathbf{U}\in\mathbb{R}^{2040200\times 1}$ for 2D Burgers' case), the sparse regression is conducted on subsampled data. Therefore, the fine-tuning step could improve the result by fully exploiting all available measurement data via the recurrent network. 

To validate the above arguments, we compute the relative $\ell_2$ error of the coefficient vector before and after the fine-tuning step. The results are provided in Table \ref{tb:importance_fine_tune}. It can be observed that the fine-tuning step consistently improves the accuracy of coefficient vector under various noise levels. Interestingly, we also find that for the GS and $\lambda$--$\Omega$ RD case with 10\% noise, the improvement brought by the fine-tuning step is not so significant compared with other cases. It is because in these two cases, the sparse regression does not give the correct form of the PDE. That said, the fine-tuning step is conducted based on an assumption that deviates from the fact. This observation also reveals a limitation of our approach -- the final discovered PDE depends largely on the sparse regression result. 

\begin{table*}[h!]
\centering
\caption{Relative $\ell_2$ error (unit: $10^{-2}$) of the coefficient vector before and after the fine tuning.}
\begin{tabular}{lcccc}
\toprule
\multirow{2}{*}{Case}  & \multirow{2}{*}{Stage}  & \multicolumn{3}{c}{Noise level}  \\
\cmidrule(lr){3-5}
 &   & 0\% & 5\% & 10\% \\
\midrule
\multirow{2}{*}{Burgers'} & Before & 0.93 & 1.57& 2.18 \\
\cmidrule{2-5}
         & After & 0.50 & 0.54& 0.59  \\
 \midrule
\multirow{2}{*}{$\lambda$--$\Omega$ RD} &  Before & 1.52 & 3.33 & 6.65 \\
\cmidrule{2-5}
                       & After & 1.18& 2.69&  5.44\\
\midrule
\multirow{2}{*}{GS RD} & Before & 6.40 & 9.84 & 12.33 \\
\cmidrule{2-5}
      & After & 1.59& 2.85& 10.03 \\
\bottomrule
\end{tabular}
\label{tb:importance_fine_tune}
\end{table*}

\subsubsection{Scalability to the noise level and data sparsity}
\label{s:larger_noise}

To examine the scalability of our method with regard to the Gaussian noise level, we performed additional experiments with the noise level of 20\% and 30\% for all three cases (2D Burgers', $\lambda$--$\Omega$ RD and GS RD system). The same settings from Section \ref{sec:exper_settings} is adopted in these experiments. The performance of our method on various noise levels is summarized in Table \ref{tb:tolerance_noise}. We observed that larger noise affects the accuracy of our method differently. For the 2D GS RD equation, the accuracy of our method deteriorates drastically as the noise level increases. That is because the measurement data used in this example has a relatively low resolution (i.e., $26\times26$). Even worse, the phenomenon in 2D GS RD system is characterized with local patterns and sharp gradients (see Fig. \ref{fig:recontruct}(c)). Increasing the Gaussian noise would further blur the spatial information in the measurement data. However, for the case of 2D $\lambda$-$\Omega$ RD and Burgers' equation where the measurement data has a decent resolution (i.e., $51\times51$) and smoother spatial patterns, our method stays competitive even for 30\% Gaussian noise. Overall, there is a trade-off between the sparsity and noise level of the measurement data. 

\begin{table*}[h!]
\center
\footnotesize
\caption{ {Performance of the proposed method on larger noise level.}}
\begin{tabular}{lcccc}
\toprule
 {Cases} &  {Noise level~/\%} &  {Precision~/\%} &  {Recall~/\%}  &  {Relative $\ell_2$ error~/$\times10^{-2}$}  \\
\midrule
\multirow{5}{*}{ {Burgers'}}  &  {0} &  {100} &  {100} &  {0.50}  \\
                        &  {5} &  {100} &  {100} &  {0.54}  \\
                        &  {10} &  {100} &  {100} &  {0.59}  \\
                        &  {20} &  {75.0} &  {100} &  {7.33}  \\
                        &  {30} &  {75.0} &  {100} &  {9.10}  \\
\midrule
\multirow{5}{*}{ {$\lambda$--$\Omega$ RD}}  &  {0} &  {100} &  {100} &  {1.18}  \\
                        &  {5} &  {100} &  {100} &  {2.69}  \\
                        &  {10} &  {91.6} &  {100} &  {5.44}  \\
                        &  {20} &  {91.6} &  {100} &  {7.65}  \\
                        &  {30} &  {91.6} &  {100} &  {15.89}  \\
\midrule
\multirow{5}{*}{ {GS RD}}  &  {0} &  {100} &  {100} &  {1.59}  \\
                        &  {5} &  {100} &  {100} &  {2.85}  \\
                        &  {10} &  {85.7} &  {85.7} &  {10.03}  \\
                        &  {20} &  {46.2} &  {85.7} &  {86.05}  \\
                        &  {30} &  {33.3} &  {37.5} &  {238.31}  \\
\bottomrule
\end{tabular}
\label{tb:tolerance_noise}
\end{table*}

As for the sparsity of the data, our method assumes the measurement data is collected on a fixed coarse sensor grid (e.g., $21\times21$ sensors), captured either by point-wise sensor units or by imaging techniques. Although the ground truth data (or numerical solution) is of high fidelity, our measurement data is heavily spatiotemporally-downsampled and, as a result, remains sparse (e.g., a finite number of LR snapshots). To examine the scalability of the proposed method regarding the data sparsity, we performed numerical experiments on the 2D GS and $\lambda$--$\Omega$ RD equations using various spatial resolutions, namely, $51\times51$, $26\times26$, $21\times21$ and $11\times11$. The observation is that the tolerable sparsity of our method depends on the spatial pattern of the system. For example, in the 2D GS RD example, the lowest resolution of data to guarantee the discoverability of the governing PDE is $26\times26$ since the very complex maze-like pattern (see Fig. \ref{fig:recontruct}(c)) is in a relatively fine scale. However, since the solution in the $\lambda$--$\Omega$ RD example is much smoother and periodic (see Fig. \ref{fig:recontruct}(b)), our method is able to discover a major portion of the PDE (i.e., precision 90.9\%, recall 83.3\%) with the spatial resolution as low as $11\times11$ (a very sparse data scenario).

{\color{black}
\section{An example with the Neumann BC type}
\label{sec:neumann_bound}
In Section \ref{sec:encoding_mechanism}, we introduce that various boundary conditions (e.g., Dirichlet or Neumann) can be encoded into the network architecture through the convolutional padding. However, the numerical examples in this paper only consider the periodic BC (Dirichlet type) for simplicity. To demonstrate our approach is general to various BCs, this section presents an example with the Neumann boundaries. To this end, we consider a 2D heat transfer problem within a $5\times5$ square domain as shown by Fig. \ref{heat_diagram}. This problem is governed by the equation $T_t=\Delta T$ where $T(x,y)$ denotes the temperature field. PeRCNN is used to solve the Initial and Boundary Value Problem (IBVP) as formulated in subsection \textcolor{blue}{Methods: Forward Analysis of PDE Systems} of the \textcolor{blue}{Main Text}. An initial temperature disturbance is adopted as the initial condition (see Fig. \ref{heat_snapshots}(a)) while the Neumann BC $\partial T/\partial \mathbf{n}=0$ is imposed on four edges where $\mathbf{n}$ denotes the outer unit normal vector on the boundary. To enforce the Neumann BC, we pad the output snapshot with a layer of ghost nodes so that the central difference approximation of $\partial T/\partial \mathbf{n}$ equals zero (see Table \ref{tb:BC_padding}). For instance, prediction of $[0,1,2,3]$ would be padded as $[1,0,1,2,3,2]$ at the left boundary in the 1D scenario. 

The PeRCNN we employ in this example features two Conv layers and two channels. The spatial and time spacing are set to be $\delta x=0.05$ and $\delta t=0.0005$ respectively. In addition, 1000 high-resolution ($101\times101$) snapshots are predicted from PeRCNN. Figure \ref{heat_snapshots} gives the comparison between the PeRCNN prediction and the reference solution (from FD method) which agree very well. To take a closer look at the prediction near the boundary, we also plot the temperature distribution on $x=5.0$ in Fig. \ref{dist_temp}. It can be seen that PeRCNN gives an excellent prediction on the Neumann boundary. All these comparisons demonstrate the effectiveness of PeRCNN on handling the Neumann BC through the proposed padding mechanism. 


	

	\begin{figure}[h!]
    \centering
	\includegraphics[width=0.23\linewidth]{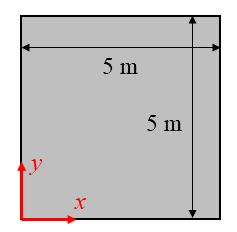}
	\caption{\hsedit{Diagram of the heat transfer problem.  } }
	\label{heat_diagram}
    \end{figure}
	
	\begin{figure}[h!]
    \centering
	\includegraphics[width=0.7\linewidth]{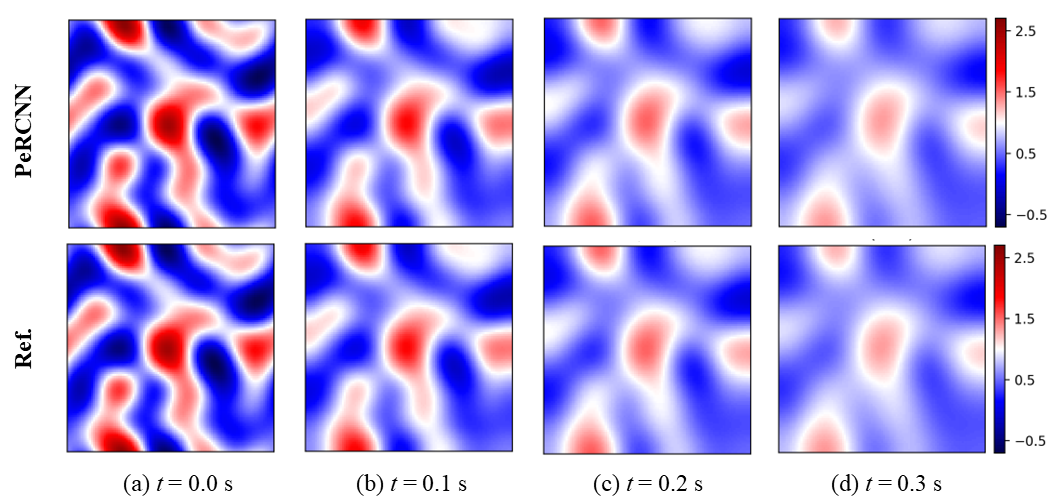}
	\caption{\hsedit{2D Heat equation under Neumann boundary condition.  } }
	\label{heat_eqn}
    \end{figure}
    
\begin{figure}[h!]
\centering
\includegraphics[width=0.65\linewidth]{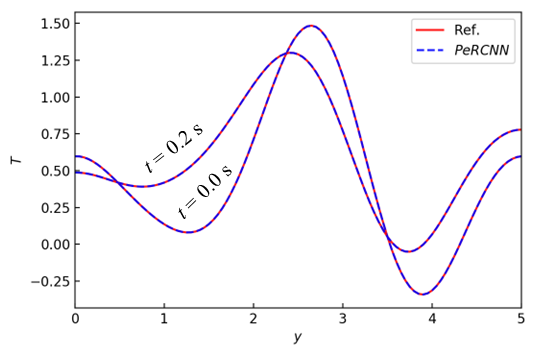}
\caption{\hsedit{Distribution of temperature $T$ at boundary $x=5.0$. } } 
\label{dist_temp}
\end{figure}

}

{\color{black}
\section{Selection of $\delta x$ and $\delta t$}
Given a pre-defined computational domain, a proper selection of $\delta x$ and $\delta t$ is critical to ensure the model's numerical stability and convergence while maintaining desired resolution. In the dataset generation phase (see \textcolor{blue}{Extended Data Table 1} in the \textcolor{blue}{Main Text}) where the governing PDEs are given, we choose $\delta x$ and $\delta t$ based on the stability and convergence requirement (e.g., $\mu \delta t/\delta x^2 \leq 1/2$, where $\mu$ denotes the largest diffusion coefficient depending on the specific problem). We also performed convergence tests to identify suitable $\delta x$ and $\delta t$ so that further refining the resolution would not significantly improve the accuracy. We select $\delta x$ and $\delta t$ in a similar way in the cases of forward and inverse analysis of PDEs by PeRCNN, when the explicit PDE expression is given. For example, the cross validation tests with various spatiotemporal resolutions were performed on our PeRCNN model to find a suitable set of $\delta x$ and $\delta t$, as demonstrated by Fig.~\ref{conv_test}. Since the system response is very smooth, as long as the above-mentioned stability condition is satisfied, the trained models with different spatiotemporal resolutions predict the system response accurately.

For the case of data-driven modeling where the the PDE expression is unknown or incomplete, $\delta x$ and $\delta t$ are selected empirically. However, based on measured snapshots at different time intervals, we can roughly estimate the diffusion coefficient $\tilde{\mu}$ using finite difference and select $\delta x$ and $\delta t$ following $\tilde{\mu} \delta t/\delta x^2 \leq 1/2$ to ensure numerical stability and convergence. Nevertheless, smaller $\delta x$ and $\delta t$ produce finer spatiotemporal solutions with higher resolution depending on our specific needs. The computational resource (e.g., memory) is another constraint. For example, the use of $49^3\times 1501$ grids in the 3D cases reached the computational resource we currently have. Parallel computing is a potential alternative to tackle this issue and scale up the model.

\begin{figure}[t!]
\centering
\includegraphics[width=0.7\linewidth]{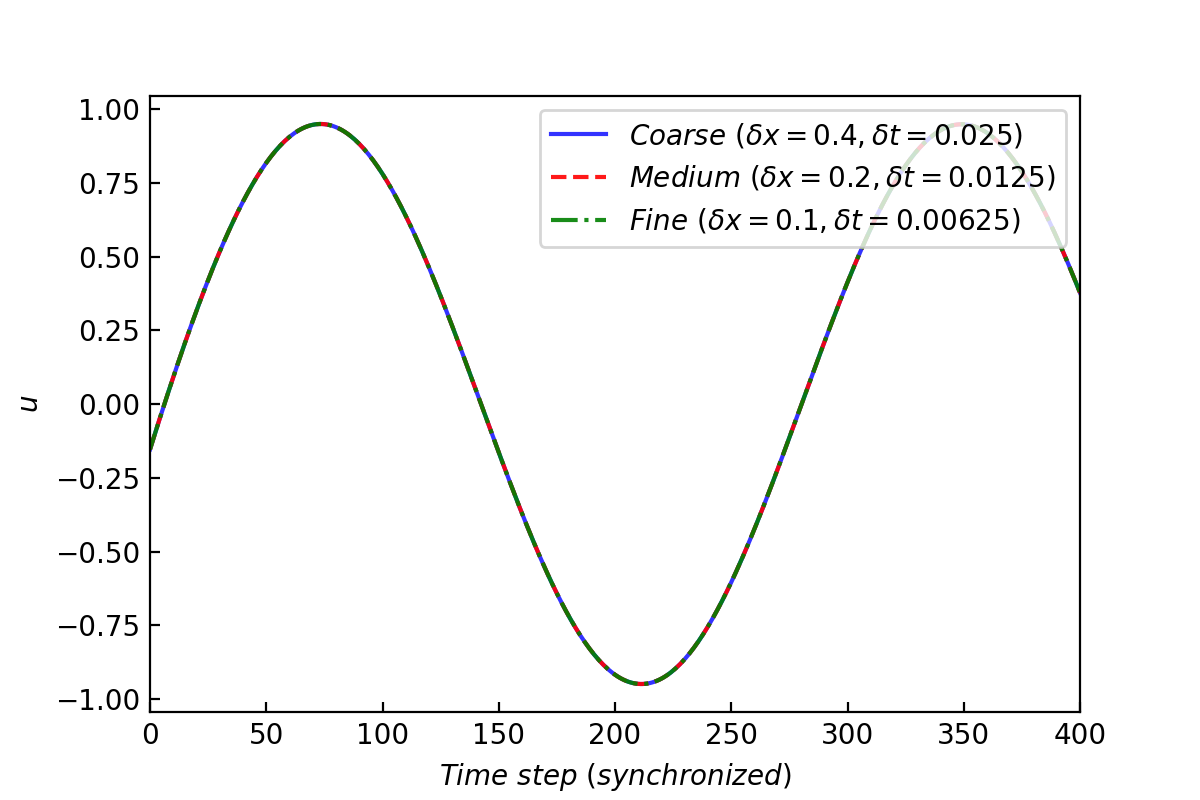}
\caption{\hsedit{Predictions of $\lambda$-$\Omega$ RD system on the point of $x=y=-5$ m under various resolutions which all satisfy the stability and convergence requirement, namely, $\mu \delta t/\delta x^2 \leq 1/2$. } }
\label{conv_test}
\end{figure}

}

{\color{black}

\section{Data-driven modeling of a PDE with non-polynomial term}\label{sec:non_poly_term}

In the previous sections, we show that $\Pi$-block is a universal polynomial approximator for PDE systems where the nonlinear equation is formulated in the polynomial form. This section seeks to test the performance of the proposed approach on PDEs with non-polynomial terms. To this end, we consider the variant 2D GS RD system as follows
\begin{equation} 
\label{eq:2d_gs_sinh} 
\begin{aligned}
u_t&=\mu_u\Delta u -  uv^2 + F(1- \sinh (u)) \\
v_t&=\mu_v\Delta v + uv^2- (F+\kappa)v
\end{aligned}
\end{equation}
where $\displaystyle \sinh (u)=\left (\exp(u)-\exp(-u) \right )/2$.  We adopt the computational framework proposed in Section \ref{sec:architecture} to solve the data-driven modeling problem formulated in Section \ref{sec:data_driven_model}.

\begin{figure}[t!]
\centering
\includegraphics[width=0.99\linewidth]{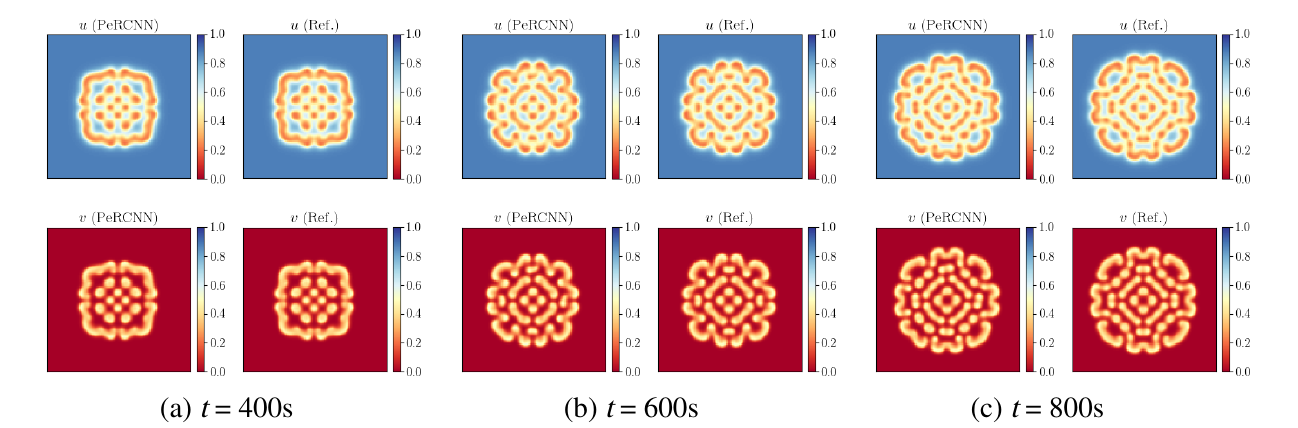}
\caption{\hsedit{\color{black}Comparison of the predicted solution and ground truth:  (a) training, and (b-c) extrapolation.} }
\label{fig:data_driven_model_sinh}
\end{figure}

The parameters listed in Table \ref{tb:data_set} are used to generate the ground truth solution for this system. The same PeRCNN architecture as in Section \ref{sec:data_driven_model} is employed to predict 800 time steps of the system in a $100\times100$ resolution. The measurement data used for model fitting includes 40 snapshots of $50\times50$ resolution. Note that 2\% Gaussian noise is added to the measurement data to test the robustness of the method. Snapshots of the predicted solution are provided in Fig. \ref{fig:data_driven_model_sinh} which shows a good agreement with the ground truth. In particular, inference beyond the time interval of training dataset shows that the dynamics of the system are well captured by the trained model.

}

{\color{black}

\section{Extension of PeRCNN for modeling turbulent flows}\label{sec:nse}

To demonstrate the capability of PeRCNN for modeling more complex systems, we consider the Kolmogorov turbulent flows at a high Reynolds number. In particular, we aim to build a PeRCNN model, which is trained with only a few amount of data, to predict the velocity and vorticity fields under general initial conditions. Let us first recall the vorticity-form Navier-Stokes equations defined in a 2D domain $\Omega$, given by
\begin{align}
    \frac{\partial \omega}{\partial t} &= -\mathbf{u}\cdot\boldsymbol{\nabla}\omega + \frac{1}{Re}\Delta \omega + f \label{NS1} \\
    \omega &= \boldsymbol{\nabla}\times\mathbf{u} \label{NS2} \\
    \boldsymbol{\nabla}\cdot\mathbf{u} &= 0 \label{NS3}
\end{align}
where $\omega(t, \mathbf{x})$ denotes the vorticity field, $\mathbf{u}(t, \mathbf{x})=[u, v]^\texttt{T}$ the velocity field, $f$ the external force (e.g., $f=0$ for decaying turbulence), $\boldsymbol{\nabla}$ the Nabla operator, $\Delta=\boldsymbol{\nabla}^2$ the Laplace operator, and $Re$ the Reynolds number. Here, $t\in[0, T]$ and $\mathbf{x}\in\Omega$ represent the time and spatial coordinates. Note that the combination of Eqs. \eref{NS2} and \eref{NS3} yields two Poisson equations for the velocity field as follows
\begin{equation}
    \Delta u = -\frac{\partial \omega}{\partial y}, ~~~\Delta v = \frac{\partial \omega}{\partial x} \label{poisson} 
\end{equation}
We leverage the Poisson equations shown in Eq. \eref{poisson} as the known \textit{a priori} knowledge to design an extended PeRCNN model for predicting $\{u, v, \omega\}$. In specific, the PeRCNN model shown in Fig. 1 in the {\color{blue} Main text} is used to unroll temporally the vorticity field, e.g., $\omega^{(k)}$, where $k = 0, 1, 2, ..., N$. At each time step, the velocity fields ($u^{(k)}$ and $v^{(k)}$) are updated by solving the Poisson equations using the standard Fast Fourier Transform method. The update of the system states follows the flow given by
\begin{equation*}
    \boxed{ \omega^{(k)} \xrightarrow{\text{Poison Solver}} \big\{u^{(k)}, v^{(k)}\big\} } \xrightarrow{\text{Assembly}} \big\{\omega^{(k)}, u^{(k)}, v^{(k)} \big\} \xrightarrow{\text{PeRCNN Update}} \omega^{(k+1)}
\end{equation*}
Two parallel layers (filter size of $5\times5$, 32 channels), are used to build the PeRCNN model. Our numerical tests demonstrate preliminarily that the above design of the extended PeRCNN model which incorporates additional prior physics (e.g., the continuity equation) improves the convergence and approximation ability. We will further refine the model and test it comprehensively in our future study.

We consider turbulent flows at Reynolds number $Re$ = 1,000, in a computational domain $[0, \pi]\times[0, \pi]$ with periodic boundary conditions. The flow is subjected to a random initial condition. The ground truth (reference) solution of the turbulent flows was generated using the direct numerical simulation (DNS) on the 1,024$\times$1,024 grid with time step $\delta t = 3.6523\times 10^{-4}$ satisfying the Courant–Friedrichs–Lewy condition \cite{kochkov2021machine}. We coarsen the ground truth solution along each spatial dimension with a factor of 8 and time with a factor of 32 to establish the training data. We only use one trajectory of training data (from one random initial condition) with 480 coarse time steps (e.g., $\Delta t = 32\delta t$). 

Fig. \ref{fig:NS_fig} shows the generalization test of the trained PeRCNN model on decaying turbulence with a different initial condition. Although the training dataset only consists 480 steps, the trained model can unroll the prediction for a longer duration (e.g., 1,000 steps) with satisfactory accuracy. The predicted flow patterns match well the ground truth depicted in Fig. \ref{fig:NS_fig}(a, c, e). The error propagation is also present in a reasonable range as shown in Fig. \ref{fig:NS_fig}(b, d, f), where the accumulated correlations are beyond 0.92 for both the vorticity and velocity fields. We conclude that this result shows the good potential of PeRCNN in turbulent flow modeling. Nevertheless, this above extension of PeRCNN in turbulence modeling is just a preliminary attempt. We will systematically investigate its capacity for modeling of more complex turbulent flows (e.g., forced flows in a much larger computational domain at higher Reynolds number) in our future work.

\begin{figure}[t!]
\centering
\includegraphics[width=0.99\linewidth]{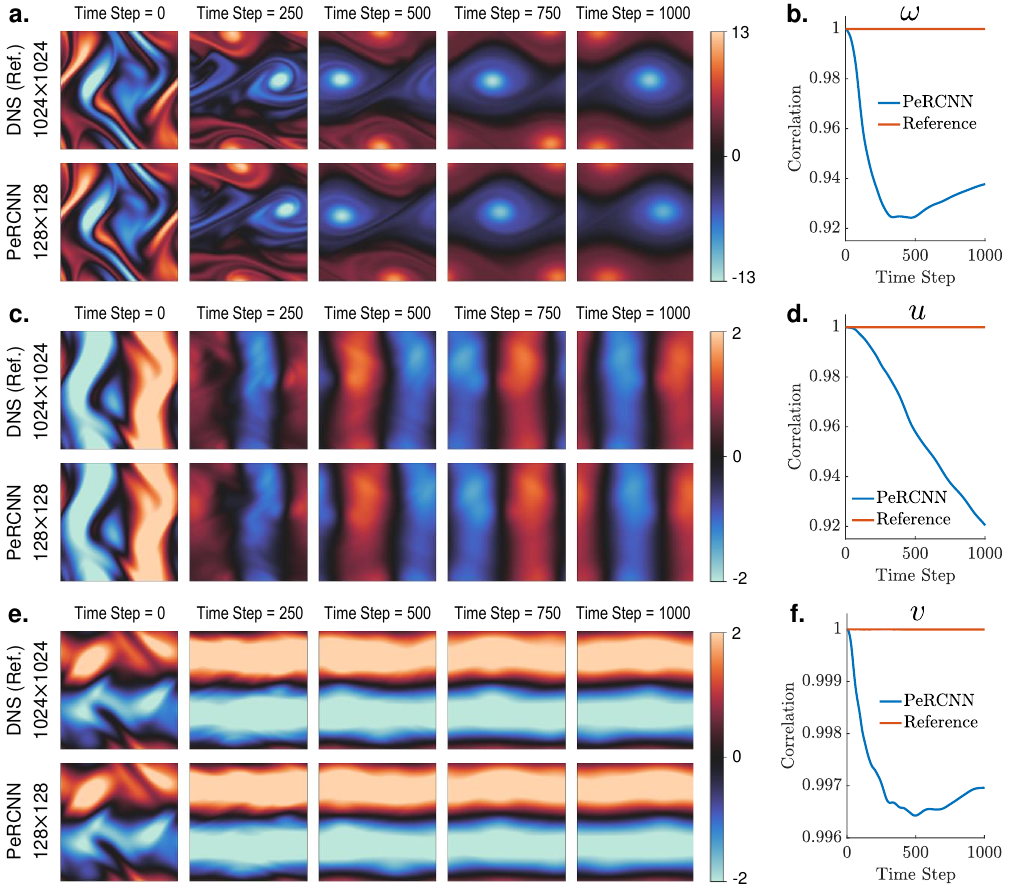}
\caption{{\color{black} The generalization test of the trained PeRCNN model on decaying turbulence with $Re =$ 1,000. (a), (c) and (e) show the evolution of predicted vorticity and velocity fields, namely, $\{\omega, u, v\}$ as a function of time, respectively. (b), (d) and (f) show the corresponding correlation between the predicted flows and the reference solution, respectively. Note that the reference solution was generated using direct numerical simulation (DNS) on the 1,024$\times$1,024 grid and down-sampled to the 128$\times$128 resolution. Each time step herein represents $\Delta t = 32\delta t$, where $\delta t = 3.6523\times 10^{-4}$. }}
\label{fig:NS_fig}
\end{figure}

}

\bibliographystyle{unsrt}
\bibliography{references_supp}